\documentclass{article}

\PassOptionsToPackage{numbers, compress}{natbib}

\usepackage{xcolor}

\usepackage{algorithm}
\usepackage{multicol}
\usepackage{caption}
\usepackage{comment}
\usepackage[export]{adjustbox}
\definecolor{ourspecialtextcolor}{rgb}{0.528, 0.471, 0.701} %
\usepackage{algorithm}
\usepackage{enumitem}   
\usepackage[noend]{algpseudocode}
\DeclareCaptionSubType*{algorithm}

\algrenewcommand{\algorithmiccomment}[1]{\bgroup\hfill//~#1\egroup}
\algrenewcommand{\Return}{\State\textbf{return}\ }
\algnewcommand{\Save}{\State\textbf{save}\ }
\algnewcommand{\Load}{\State\textbf{load}\ }

\usepackage{ragged2e}

\usepackage[final]{neurips_2023}

\usepackage{subfig}
\usepackage[utf8]{inputenc} 
\usepackage[T1]{fontenc}    
\usepackage[hyperfootnotes=false]{hyperref}       
\usepackage{url}            
\usepackage{booktabs}       
\usepackage{amsfonts}       
\usepackage{nicefrac}       
\usepackage{microtype}      
\usepackage{graphicx}
\usepackage{float}
\usepackage{mathtools}
\usepackage{bbm}
\usepackage{amssymb}
\usepackage{amsmath}
\usepackage{wrapfig}
\usepackage{color, colortbl}
\definecolor{LightCyan}{rgb}{0.88,1,1}
\definecolor{Gray}{gray}{0.95}

\usepackage{bm}

\DeclareMathOperator*{\argmin}{arg\,min}

\usepackage{mathtools}

\usepackage{multirow}

\usepackage{xspace}

\newcommand{\bc}{\bm{c}}

\newcommand{\bz}{\bm{z}}
\newcommand{\bx}{\bm{x}}
\newcommand{\by}{\bm{y}}

\newcommand{\bv}{\bm{v}}

\newcommand{\bepsilon}{\bm{\epsilon}}

\newcommand{\btheta}{\bm{\theta}}

\newcommand{\grad}[2]{\nabla_{#1}#2}

\newcommand{\bnoisedist}{\rho(\bepsilon)}

\newcommand\blfootnote[1]{%
  \begingroup
  \renewcommand\thefootnote{}\footnote{#1}%
  \addtocounter{footnote}{-1}%
  \endgroup
}
\newcommand{\code}[1]{\texttt{#1}}
\newcommand{\com}[1]{{\color{purple}\texttt{#1}}}
\usepackage[textsize=tiny]{todonotes}

\definecolor{cyan}{cmyk}{.3,0,0,0}

\title{LVM-Med: Learning Large-Scale Self-Supervised Vision Models for Medical Imaging via Second-order Graph Matching}

\author{Duy M. H. Nguyen$^{*\,1,2,3}$,~Hoang Nguyen$^{3}$,~Nghiem T. Diep$^{3}$,~Tan N. Pham$^{3,4}$,~Tri Cao$^{3}$, \\ \textbf{Binh T. Nguyen$^{4}$,~Paul Swoboda$^{5}$,~Nhat Ho$^{6}$,~Shadi Albarqouni$^{7,8}$,~Pengtao Xie$^{9,10}$},\\ \textbf{Daniel Sonntag$^{\dagger\,3,11}$,~Mathias Niepert$^{*\,\dagger\,1,2}$}
\\\\
$^{1}$University of Stuttgart, $^{2}$IMPRS for Intelligent Systems \\
$^{3}$German Research Center for Artificial Intelligence, $^{4}$University of Science - VNUHCM, \\ $^{5}$Max Planck Institute for Informatics, $^{6}$University of Texas at Austin $^{7}$Helmholtz Munich, \\ $^{8}$University Hospital Bonn, $^{9}$UC San Diego, $^{10}$MBZUAI, $^{11}$Oldenburg University.
}
\begin{document}
\maketitle
\vspace{-0.2in}
\begin{abstract}
Obtaining large pre-trained models that can be fine-tuned to new tasks with limited annotated samples has remained an open challenge for medical imaging data.
While pre-trained deep networks on ImageNet and vision-language foundation models trained on web-scale data are prevailing approaches, 
their effectiveness on medical tasks is limited due to the significant domain shift between natural and medical images. To bridge this gap, we introduce LVM-Med, the first family of deep networks trained on large-scale medical datasets. 
We have collected approximately $1.3$ million in medical images from 55 publicly available datasets, covering a large number of organs and 
modalities such as CT, MRI, X-ray, and Ultrasound. We benchmark several state-of-the-art self-supervised algorithms on this dataset and propose a \emph{novel self-supervised contrastive learning algorithm using a graph matching formulation}. The proposed approach makes three contributions: (i)~it integrates prior pair-wise image similarity metrics based on local and global information; (ii)~it captures the structural constraints of feature embeddings through a loss function constructed via a combinatorial graph-matching objective; and (iii)~it can be trained efficiently end-to-end using modern gradient-estimation techniques for black-box solvers. We thoroughly evaluate the proposed LVM-Med on $15$ downstream medical tasks ranging from segmentation and classification to object detection, and both for the in and out-of-distribution settings. LVM-Med empirically outperforms a number of state-of-the-art supervised, self-supervised, and foundation models. For challenging tasks such as Brain Tumor Classification or Diabetic Retinopathy Grading, LVM-Med improves previous vision-language models trained on 1 billion masks by 6-7$\%$ while using only a ResNet-50. We release pre-trained models at this link \url{https://github.com/duyhominhnguyen/LVM-Med}.\\
\blfootnote{$^{*}$Corresponding authors, $^{\dagger}$Co-Senior authors.}
\end{abstract}

\section{Introduction}
\vspace{-0.1in}
Constructing large-scale  annotated medical image datasets for training deep networks is challenging due to data acquisition complexities, high annotation costs, and privacy concerns \citep{cheplygina2019not,kaissis2020secure}.
Vision-language pretraining has emerged as a promising approach for developing foundational models that support various AI tasks. Methods such as CLIP \cite{radford2021learning}, Align \cite{jia2021scaling}, and Flava \cite{singh2022flava} propose a unified model trained on large-scale image-text data, showing exceptional capabilities and performance across various tasks. However, their effectiveness in the medical domain still remains unclear. A recent work SAM \cite{kirillov2023segment} trains large vision models  on over one billion annotated masks from 11M natural images, enabling interactive segmentation. Nevertheless, SAM's zero-shot learning performance is moderate on other datasets \cite{mazurowski2023segment,he2023accuracy}, highlighting the need for fine-tuning to achieve satisfactory results \cite{ma2023segment}.

\begin{figure}[H]
    \centering
    \small
    \includegraphics[width=.6\textwidth]{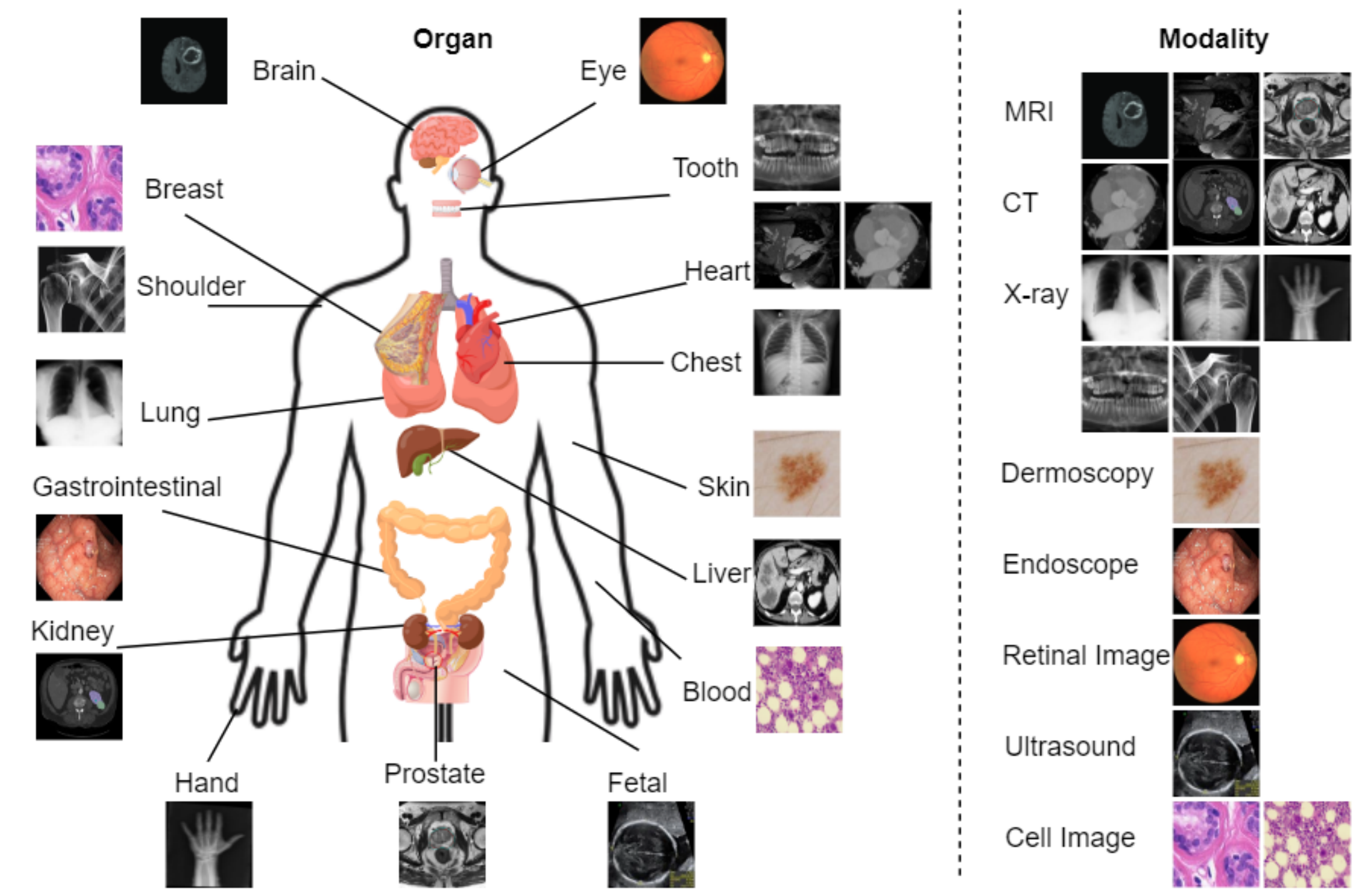}
    \includegraphics[width=.35\textwidth]{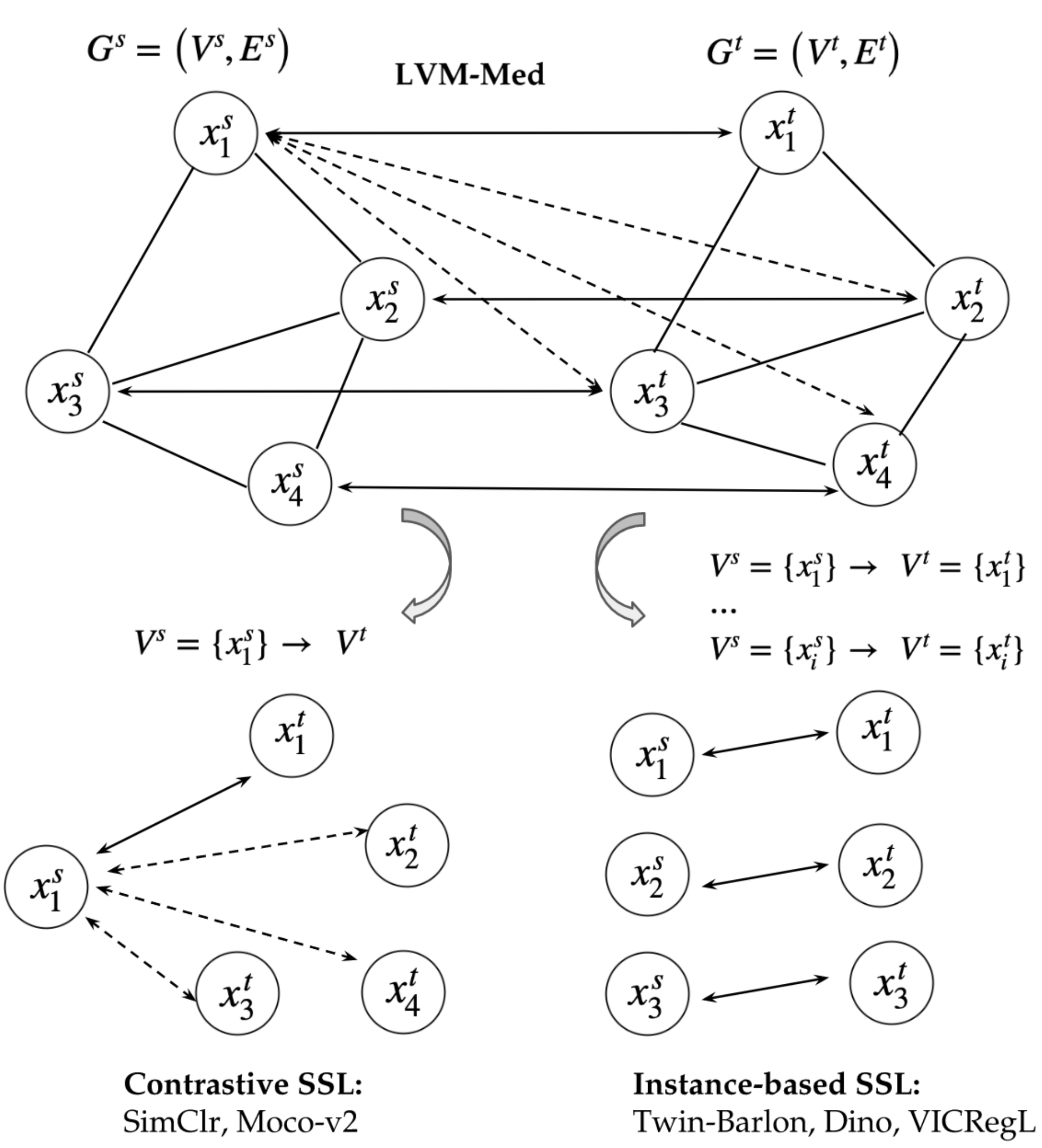}
    \caption{\small{(left) Overview of the body organs and modalities in our collected dataset; (right) LVM-Med unifies and extends contrastive and instance-based self-supervised learning approaches by specifying graph's properties.}}
    \label{fig:my_label}
\end{figure}
\vspace{-0.2in}
To facilitate the development of foundation models in the medical domain, we make two major contributions. First, we have curated a vast collection of 55 publicly available datasets, resulting in approximately 1.3 million medical images covering various body organs and modalities such as CT, MRI, X-ray, ultrasound, and dermoscopy, to name a few. 
Second, we propose LVM-Med, a novel class of contrastive learning methods, utilizes pre-trained ResNet-50 and a ViT network SAM\cite{kirillov2023segany}. We evaluate various instances of LVM-Med relative to popular supervised architectures and vision-language models across $15$ medical tasks. To our best knowledge, this is the first time such a large-scale medical dataset has been constructed and used to investigate the capabilities of SSL algorithms.


LVM-Med incorporates a second-order graph-matching formulation, 
which subsumes and extends a large class of contrastive SSL methods. Given a batch of images, two random transformations are applied to each image, and the resulting transformed images are then fed to an image encoder. The embedding vectors obtained from images in a batch are used to construct two graphs where vertices represent pairs of transformed images generated from the same original one. 
Through solving a graph-matching problem \cite{sun2020survey,haller2022comparative}, we learn feature representation such that their encoding serves as suitable priors for a global solution of the graph-matching objective.  This approach is distinct from prior contrastive learning methods that focus on merely optimizing pair-wise distances \cite{zbontar2021barlow,bardes2022vicregl} between transformed images or learning contrastive distances with positive and negative samples \citep{chen2020simple,he2020momentum,chen2020improved,tomasev2022pushing,tang2022unifying}.
It is worthwhile noting that previous contrastive learning methods are special instances of our general framework (Figure \eqref{fig:my_label}, right). 

LVM-Med has several advantages over existing approaches. First, it integrates advanced pair-wise image similarity taken from prior SSL methods into vertex affinities, resulting in both global and local   information that can be efficiently fused.
Second, it uncovers underlying structures of feature embeddings by utilizing edge constraints, enhancing robustness in the presence of similar entities in medical datasets. Third, though combinatorial problems are typically non-differentiable, LVM-Med can efficiently calculate gradients through the discrete combinatorial loss function using modern implicit maximum likelihood estimation techniques. 
Consequently, LVM-Med can scale successfully on large-scale data. In a wide range of $15$ medical experiments, LVM-Med sets a new state-of-the-art in fully fine-tuning or prompt-based segmentation, linear and fully fine-tuning image classification, and domain generalization, outperforming several vision-language models trained on a hundred million image-text instances.

We summarize major contributions in this work, including:
\begin{enumerate}[label=(\roman*)]
    \item We present a collection of large-scale medical datasets, serving as a resource for exploring and evaluating self-supervised algorithms.
    \item We propose LVM-Med, a novel SSL approach based on second-order graph matching. The proposed method is flexible in terms of integrating advanced pair-wise image distance and being able to capture structural feature embedding through the effective utilization of second-order constraints within a global optimization framework. 
    \item On both ResNet-50 and ViT architectures, LVM-Med consistently outperforms multiple existing self-supervised learning techniques and foundation models across a wide range of downstream tasks. 
\end{enumerate}


\section{Related Work}
\vspace{-0.05in}
\subsection{Self-supervised learning in medical image analysis}
\vspace{-0.1in}
The latest approaches of \textit{global feature} SSL rely on shared embedding architecture representations that remain invariant to different viewpoints. The variation lies in how these methods prevent collapsing solutions. \textit{Clustering methods} \citep{caron2018deep,caron2020unsupervised,zhan2020online} constrain a balanced partition of the samples within a set of cluster assignments. \textit{Contrastive methods} \citep{chen2020simple,he2020momentum,chen2020improved,tomasev2022pushing} uses negative samples to push far away dissimilar samples from each other through contrastive loss, which can be constructed through memory bank \cite{wu2018unsupervised}, momentum encoder \cite{chen2021empirical}, or graph neural network \cite{tang2022unifying}. Unlike contrastive learning, \textit{instant-based learning} depends on maintaining the informational context of the feature representations by either explicit regularization \cite{zbontar2021barlow,garrido2022duality} or architectural design \cite{grill2020bootstrap,lee2021compressive}. Our work relates to contrastive and instance-based learning, where a simplified graph-matching version of 1-N or 1-1 reverts to these approaches. 

In contrast to global methods, \textit{local methods} specifically concentrate on acquiring a collection of local features that depict small portions of an image. A contrastive loss function can be used on those feature patches at different criteria such as image region levels \cite{xiao2021region}, or feature maps \citep{wang2021dense,bardes2022vicregl}. These strategies are also widely applied in the medical context, thereby pre-text tasks based on 3D volume’s properties, such as reconstructing the spatial context \cite{zhuang2019self}, random permutation prediction \cite{chen2019self} and self-restoration \cite{zhou2021models,haghighi2021transferable}, are proposed. Our LVM-Med model on this aspect can flexible unifying both global and local information by adding them to the affinities matrixes representing the proximity of two graphs, enhancing expressive feature representations.
\vspace{-0.1in}
\subsection{Vision-language foundation models}
\vspace{-0.1in}
In order to comprehend the multi-modal world using machines, it is necessary to create foundational models that can operate across diverse modalities and domains \cite{bommasani2021opportunities}. CLIP \cite{radford2021learning} and ALIGN \cite{jia2021scaling}  are recognized as groundbreaking explorations in foundation model development. These models demonstrate exceptional proficiency in tasks such as cross-modal alignment and zero-shot classification by learning contrastive pretraining on extensive image-text pairs from the web, despite the presence of noise. To further support multi-modal generation tasks such as visual question answering or video captioning, recent works such as FLAVA \cite{singh2022flava} and OmniVL \cite{wang2022omnivl} are designed to learn cross-modal alignment as well as image-video language models. Conversely, the SAM model \cite{kirillov2023segment} utilized a supervised learning strategy with over $1$ billion masks on $11$ million user-prompt interactions and achieved impressive zero-shot segmentation performance on unseen images. While many efforts have been proposed for natural image domains, limited research has been conducted on large-scale vision models for medical imaging. This motivated us to develop the LVM-Med model.
\vspace{-0.1in}
\subsection{Graph matching in visual computing}
\vspace{-0.1in}
Graph matching is a fundamental problem in computer vision, which aims to find correspondences between elements of two discrete sets, such as key points in images or vertices of 3D meshes, and used in numerous vision tasks, including 3D vision \cite{ma2021image,cao2023self}, tracking \cite{yilmaz2006object,hyun2023detection}, shape model learning \cite{heimann2009statistical,roetzer2023conjugate}, and many others
\cite{bian2022unsupervised,wu2023unsupervised,peng2022gate,liu2022self}. In this framework, the vertices of the matched graphs correspond to the elements of the discrete sets to be matched. Graph edges define the cost structure of the problem, namely, second order, where pairs of matched vertices are penalized in addition to the vertex-to-vertex matchings. This allows us to integrate the underlying geometrical relationship between vertices into account but also makes the optimization problem NP-hard. Therefore, many approximate approaches have been proposed to seek acceptable suboptimal solutions by relaxing discrete constraints \cite{zass2008probabilistic,zhou2015factorized}. In other directions, gradient estimation techniques for black-box solvers are employed to make the hybrid discrete-continuous matching framework be differentially end-to-end \cite{poganvcic2020differentiation,rolinek2020deep,niepert21imle}. Our LVM-Med follows the latter direction and, for the first time, presents the formulation of contrastive learning as a second-order graph-matching problem.
\section{Methodology}
\subsection{Dataset construction}
We provide detailed information about the collected datasets in the Appendix. The data was collected from publicly available resources, which include a diverse set of modalities and body organs as illustrated in Figure~\ref{fig:my_label} (left). The data format is a combination of 2D images and 3D volumes as well as X-ray, MRI, CT, Ultrasounds, etc. For datasets whose data
dimensions are 3D volumes, we slice them into 2D images.
To avoid potential test data leaking for downstream tasks,
we use the default training partition in each dataset; otherwise, we randomly sample with $20\%$ total images. In total, we obtain approximately $1.3$ million images. More statistics on the dataset are presented in the Appendix.
\subsection{Contrastive learning as graph matching}
\begin{figure}[t!]
\centering
\includegraphics[width=0.85 \textwidth]{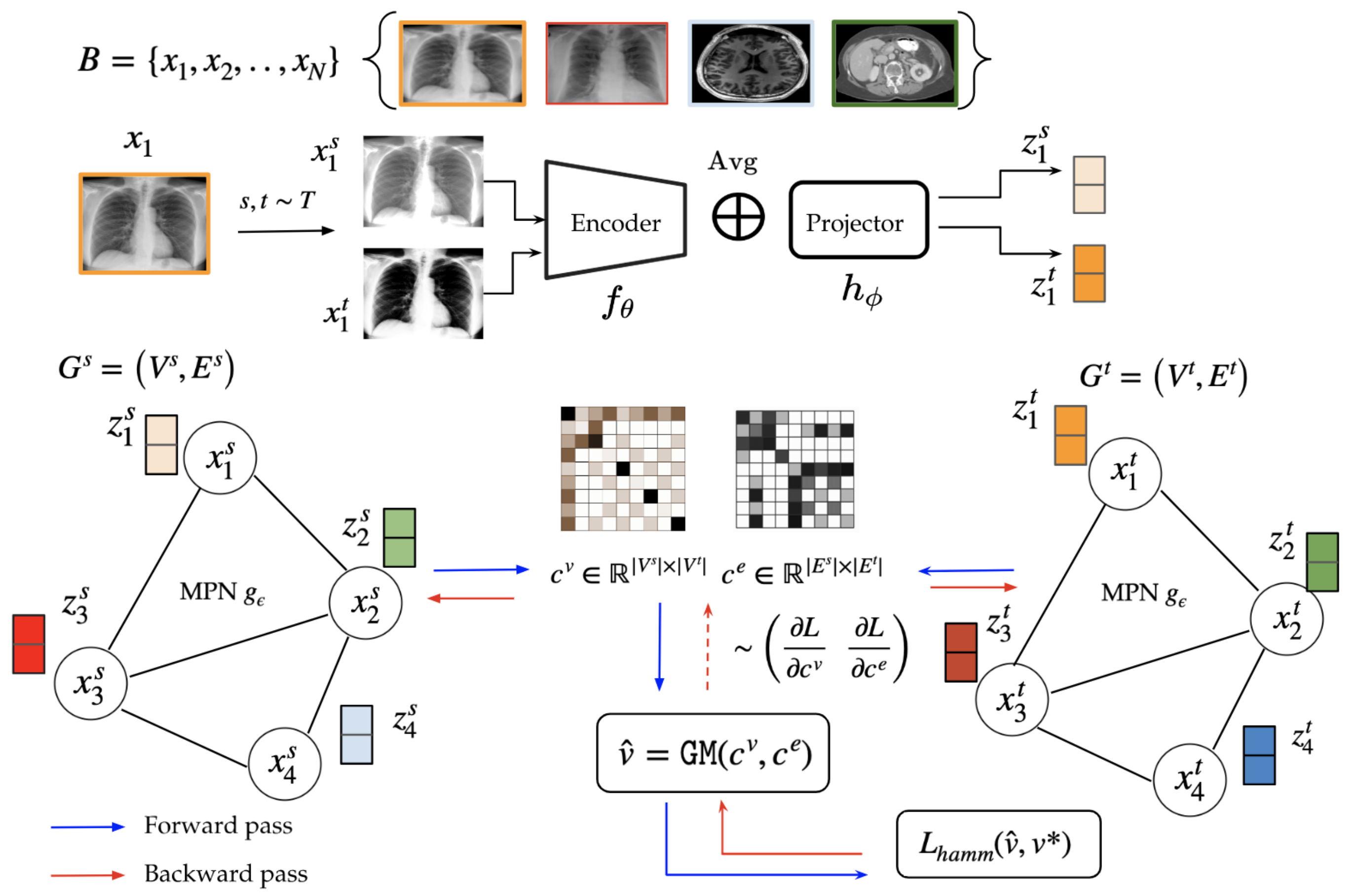}
\caption{\small{LVM-Med Overview. $\mathrm{Avg}$ is the average pooling layer, MPN denotes for message passing network, $\mathtt{GM}$ indicates the combinatorial solver, and $\left(c^v, c^e\right)$ represents vertex and edge affinity matrices. For each image $\bx_{i}$ in batch size, we generated two distorted versions and fed them into the feature representation $f_{\theta}$ and another projector $h_{\theta}$. The obtained embeddings $\bm{z}_i^{\ell},\ \ell \in (s,t)$ are used to build two graphs $G^{s}, G^{t}$. We further design a message passing network $g_{\epsilon}$ that aggregate feature per node by their neighbor information. Then we compute vertex and edge affinities $\bm{c}^v, \bm{c}^e$ and use them to solve the graph matching. The output afterward is compared with pairs of ground truth $\left(\bx_{i}^{s},\bx_{i}^{t}\right), i \in(1,..,N)$ representing distorted images generated from the same sample. In the backward pass, we use modern gradient-estimation techniques to approximate $\frac{\partial L}{\partial \bm{c}^v}$ and $\frac{\partial L}{\partial \bm{c}^e}$ }.}
\label{fig:overview_framework}
\end{figure}

Figure \ref{fig:overview_framework} provides an illustration 
of our LVM-Med method, which learns the feature representation $f_{\theta}$ by matching two distorted views derived from the same input image through a graph-matching formulation. Below we describe in detail each component.
\subsubsection{Graph construction on feature embedding}


Given a batch  of $N$ images $\bm{B} =\{\bm{x}_1,\, \bm{x}_2,\,..,\bm{x}_N\}$ sampled from a dataset, we generate for each image $\bm{x}_i \in \bm{B}$  two transformed images $\bx_{i}^s$ and $\bx_{i}^t$ by using two transformations $s, t \sim T$ sampled from $T$, a set of pre-defined image transformations. After the transformations, each image is of shape $(C \times H \times W)$, where $C$ is the number of channels and $(H, W)$ the original spatial dimensions. These distorted images are fed into an encoder $f_{\theta}: \mathbb{R}^{C \times H \times W}  \rightarrow \mathbb{R}^{D\times R \times S}$ to produce two representations $\by_{i}^{s} = f_{\theta}(\bx_{i}^{s})$ and $\by_{i}^{t} = f_{\theta}(\bx_{i}^{t})$ where $D$ is the number of feature channels and $(R, S)$ are the spatial dimensions of the feature map.
On each such representation, we perform an average pooling operation $\mathrm{Avg}:\mathbb{R}^{D\times R \times S} \rightarrow \mathbb{R}^{D}$ followed by another projection $h_{\phi}: \mathbb{R}^{D} \rightarrow \mathbb{R}^{F}$ to form two feature embeddings $\bz_{i}^{s} = h_{\phi}(\mathrm{Avg}(\by_{i}^{s}))$, and
$\bz_{i}^{t} = h_{\phi}(\mathrm{Avg}(\by_{i}^{t})) \in \mathbb{R}^{F}$ with $F < D$. 

Given a set of embeddings for a batch $\bm{B}$, we construct two graphs $G^s$ and $G^t$ where, for each pair $(\bx_{i}^s, \bx_{i}^t)$ of corresponding distorted images, we add a node representing $\bx_{i}^s$ to $G^s$ and a node representing $\bx_{i}^t$ to $G^t$. Hence, for each $\ell \in \{s, t\}$, we construct a graph $G^{\ell} =(V^{\ell}, E^{\ell})$ with $V^{\ell} = \{\bx_{1}^{\ell},...,\bx_{N}^{\ell}\}$ the set of vertices and $E^{\ell}$ the set of edges $e_{ij}^{\ell} = (\bx_i^{\ell}, \bx_j^{\ell})$. The node-level feature matrix is given by $\bm{X}^{\ell} = \left[\bz_{1}^{\ell}; ...; \bz_{N}^{\ell} \right] \in \mathbb{R}^{N\times F}$ which associates each vertex $\bx_{i}^{\ell}$  with its feature embedding $\bz_{i}^{\ell}$. We create edges for each graph $G^{\ell}$ through a $k$-nearest neighbors algorithm using the feature matrix $\bm{X}^{\ell}$. The adjacency matrix $\bm{A}^{\ell} \in \mathbb{R}^{N\times N}$ is defined as $A_{ij}^{\ell} = 1$ if $e_{ij}^{\ell} \in E^{\ell}$ and $A_{ij} = 0$ otherwise.  
With the two graph structures given, we obtain a node-attributed graph $G^{\ell} = (V^{\ell}, \bm{A}^{\ell}, \bm{X}^{\ell})$ on which a graph neural network $g_{\varepsilon}$ is used to aggregate the nodes' features. In particular, $g_{\varepsilon}$ computes an embedding $\hat{\bm{Z}}^{\ell} = g_{\varepsilon}(\bm{X}^{\ell},\bm{A}^{\ell})$ by performing message passing operations. We set  $g_{\varepsilon}$ to be a  graph convolutional network \citep{gcn,wu2019simplifying} consisting of  $l+1$ layers $g_{\varepsilon} = \{g_{l}, g_{l-1},.., g_{0}\}$ where the output of layer $l$ is computed as
\begin{equation}
H_{l}^{\ell} = \sigma\left(\tilde{D}^{-\frac{1}{2}} (\bm{A}^{\ell} +\bm{I}_{N})\tilde{D}^{-\frac{1}{2}}H_{l-1}^{\ell}g_{l-1}\right),
\label{eq:gnn}
\end{equation}
where $\bm{I}_{N}$ is the identity matrix modeling  self-connections; $\tilde{D}$ is a diagonal matrix with $\tilde{D}_{ii} =\sum_{j} {\bm{A}}_{ij}^{\ell}$; $g^{l-1}$ are the trainable parameters for each layer; $\sigma(\cdot)$ is an activation function; and $H_{0}^{\ell} =\bm{X}^{\ell}$. We use the outputs of the last layer as embeddings for the nodes, that is, $\hat{\bm{Z}}^{\ell} = H_{l}^{\ell} \in \mathbb{R}^{N\times F}$ given the shared graph network $g_{\varepsilon}$.

We now have two graphs $G^{s}, G^{t}$ with node attribute matrices $\hat{\bm{Z}}^s,\, \hat{\bm{Z}}^t$, the outputs of the graph neural networks. Next, a graph-matching problem is constructed and solved where the gold matching is given by the pairs  
$(\bm{x}_i^{s}, \bm{x}_i^{t}) \ \  \forall i \in \{1,..,N\}$. 

\subsubsection{Learning affinities with global and local context}
To represent potential connections for a pair of node $(\bx_{i}^{s}, \bx_{a}^{t})$ where $\bx_{i}^{s} \in G^{s}, \ \bx_{a}^{t} \in G^{t}$, we design a vertex affinity matrix $\bm{c}^{v} \in \mathbb{R}^{|V^{s}||V^{t}|}$ where $c_{ia}^{v}$ is the prior (feature-based) similarity between $\bx_{i}^{s}$ and $\bx_{a}^{t}$. An advantage of our formulation is its ability to integrate advanced pair-wise distance can be smoothly integrated to $c_{ia}^{v}$, resulting in more expressive proximity representation. 
In particular, we leverage both global and local consistency derived from feature embeddings of distorted images. The \textit{global distance} used in several prior works can be computed as  $c_{ia}^{{glo}} (\bx_{i}^{s},\bx_{a}^{t}) = \cos(\hat{\bm{z}}_{i}^{s}, \hat{\bm{z}}_{a}^{t})$ where $\cos(\cdot)$ denotes  cosine similarity; $\hat{\bm{z}}_{m}^{\ell}$ is the embedding of $\bx_{m}^{\ell} \left(\ell \in \{s,t\},\ m \in \{i,a\}\right)$ obtained after message passing  in Eq.\,\eqref{eq:gnn}. 

Compared to global methods that implicitly learn features for the entire image, local methods concentrate on explicitly learning a specific group of features that characterize small regions of the image. As a result, they are more effective for dense prediction tasks such as segmentation  \cite{wang2021dense,bardes2022vicregl,yang2022inscon}. While recent works applied these tactics as a part of pair-wise minimization conditions \cite{xie2021propagate,xiao2021region}
Instead, we integrate them as a part of vertex costs $c_{ia}^v$ and use it to solve the graph matching problem.
Indeed, we adapt both location- and feature-based local affinity computed as: 
\begin{equation}
  c_{ia}^{\,lo} (\bx_{i}^{s},\bx_{a}^{t}) =  \mathbb{E}_{p \in \bm{P}} \cos(\bm{q}_{p}^{s}, \bm{q}_{\mathrm{m}(p)}^{t}) + \mathbb{E}_{p \in \bm{P}} \cos(\bm{q}_{p}^{s}, \bm{q}_{\mathrm{m'}(p)}^{t})
\label{eq:local_cost}
\end{equation}
where $\bm{P} = \{(r, s)| \ (r,s) \in \left[1,...,R\right] \times \left[1,..,S\right]\}$ be the set of coordinates in the feature map $\bm{y}_{i}^s \in \mathbb{R}^{D\times R \times S}$ of $\bm{x}_{i}^{s}$; $\bm{q}_{p}^{\ell}$ $(\ell \in \{s, t\})$ be the feature vector at position $p$; $\mathrm{m}(p)$ denote the spatial closest coordinate to $p$ in coordinates of feature map $\bm{y}_{a}^t$ estimated through transformations on original image $\bm{x}_{i}$; finally $\ \mathrm{m'}(p)$ represents the closest feature vector to $p$ in $\bm{y}_{a}^t$ using $l^2$ distance. Intuitively, the local cost in Eq. \eqref{eq:local_cost} enforces invariance on both spatial location and between embedding space at a local scale. Our final affinity cost is computed as:
\begin{equation}
    c_{ia}^v(\bx_{i}^{s},\bx_{a}^{t}) = \alpha\left(c_{ia}^{\,glo} (\bx_{i}^{s}, \bx_{a}^{t})\right) + (1-\alpha)\left(c_{ia}^{lo}(\bx_{i}^{s}, \bx_{a}^{t}) + c_{ia}^{lo}(\bx_{a}^{t},\bx_{i}^{s})\right)
    \label{eq:total_cost}
\end{equation}

\subsubsection{Self-supervision through second-order graph matching}

While the standard graph matching problem for vertex-to-vertex correspondences can be used in our setting (LAP), it fails to capture the similarity between edges. If there are duplicated entities represented by distinct nodes in the same graph, the LAP will consider them identical and skip their neighboring relations. For instance, during the image sampling, two consecutive image slides were sampled from a 3D volume, resulting in their appearances have s a small difference. In such cases, 
it is complicated to correctly identify those augmented images generated from the same one without using information from the relations among connected nodes in the constructed graph. To address this problem, we introduce additional edge costs $\bm{c}^{e} \in 
\mathbb{R}^{|E^{s}||E^{t}|}$ where $c^{e}_{ia,jb}$ represents the similarity between an edge $v_{ij}^{s} = \left(\bx_{i}^{s},\bx_{j}^{s}\right) \in E^{s}$ and  $v_{ab}^{t} = (\bx_{a}^{t},\bx_{b}^{t}) \in E^{t}$. 
These edge costs (second-order) are computed as $ c_{ia,jb}^{e} = \cos((\mathbf{\hat{z}}_{i}^s - \mathbf{\hat{z}}_{j}^{s}),(\mathbf{\hat{z}}_{a}^t - \mathbf{\hat{z}}_{b}^{t}))$.


We now establish the second-order graph-matching problem. Denoting $\bm{v} = \{0, 1\}^{|V^s||V^t|}$ be indicator vector of matched vertices, i.e., $v_{ia} = 1$ if the vertex $\bx_{i}^{s} \in V^s$ is matched with $\bx_{a}^{t} \in V^t$ and $v_{ia} = 0$ otherwise. The node correspondence between two graphs $G^{s}$ and $G^{t}$ that minimizes the global condition stated as:
\begin{equation}
\label{eq:QAP3}
\begin{array}{rll}
   & \mathtt{GM}(\bm{c}^v, \bm{c}^e) = \argmin \limits_{\substack{\bm{v} \in U(\textbf{1}, \textbf{1})}}  -\sum_{i,a} c^v_{ia} v_{ia} - \sum_{i,j,a,b} c^e_{ia,jb}  v_{ia}v_{jb} \\ 
       \text{where}  
       &  U(\textbf{1}, \textbf{1}) = \{\bm{v} \in \{0, 1\}^{N \times N} | \bm{v}\textbf{1}_N = \textbf{1}, \bm{v}^T\textbf{1}_N = \textbf{1}\} 
    \end{array}
\end{equation}
and $\textbf{1}_{N}$ be a $n$-dimension one-value vector. The constraint $U(\textbf{1},\textbf{1})$ restricts $\bm{v}$ satisfying the one-to-one matching. Essentially, the Eq. \eqref{eq:QAP3} solves the vertex-to-vertex correspondence problem using both node and edges affinities, which can be seen as a form of structural matching (Figure\,\eqref{fig:my_label},right) and generally can be integrated with higher-order graph constraints as triangle connections or circles.  In the experiment, we found out that Eq. \eqref{eq:QAP3} significantly improved downstream task performance compared to the pure linear matching approach (Table \eqref{tab:ablation-study}).
Since the Eq. \ref{eq:QAP3} in general is an NP-Hard problem \cite{burkard1998quadratic} due to its combinatorial nature, we thus use efficient heuristic solvers based on Lagrange decomposition techniques \cite{swoboda2017study}.
\subsubsection{Backpropagating through a graph matching formulation}
With $\hat{\bm{v}} = \mathtt{GM}(\bm{c}^v, \bm{c}^e)$
 a solution obtained from the solver, we use the Hamming distance
and an optimal solution $\bm{v}^{*}$ to define the following loss function
\begin{equation}
    L(\hat{\bm{v}}, \bm{v}^*) = \hat{\bm{v}}.(1-\bm{v}^*) + \bm{v}^*.(1-\hat{\bm{v}}).
    \label{eq:hamming}
\end{equation}
The proposed approach aims to learn the feature representation function $f_{\theta}$ such that its output minimizes Eq.\,\eqref{eq:hamming}. However, this is a difficult problem because the partial derivatives of the loss function w.r.t vector costs $\bm{c}^v, \bm{c}^e$, i.e., $\partial L / \partial \bc^v$ and $\partial L / \bc^e$, are zero almost everywhere \citep{poganvcic2020differentiation,rolinek2020optimizing} due to the objective function in Eq.\,(\ref{eq:QAP3}) being piece-wise constant, preventing direct gradient-based optimization.  

To approximate the gradients required for backpropagation, we adopt IMLE \citep{niepert21imle,minervini2022adaptive}. Let $\btheta = \left(\bm{c}^v, \bm{c}^e\right)$ be the input to the combinatorial graph matching problem in Eq. \eqref{eq:QAP3}. The core idea of IMLE is to define a probability distribution $\rho(\bv; \btheta)$ over solutions of the combinatorial optimization problem, where the probability of a solution is proportional to its negative cost, and to estimate $\partial L / \partial \btheta$ through the gradients of the expectation $\nabla_{\btheta} \mathbb{E}_{\bm{\hat{v}}\sim \rho(\bv; \btheta)}\left[L(\bm{\hat{v}},\bv^{*})\right]$. Since exact sampling from $\rho(\bv; \btheta)$ is typically intractable, IMLE instead chooses a noise distribution $\rho(\bepsilon)$ and approximates the gradient of the expectation over $\rho(\bv; \btheta)$ with the gradient of the expectation over $\rho(\bepsilon)$ 
\begin{equation*}
\nabla_{\btheta} \mathbb{E}_{\bm{\hat{v}}\sim \rho(\bv; \btheta)}\left[L(\bm{\hat{v}},\bv^{*})\right] \approx \nabla_{\btheta}\mathbb{E}_{\bepsilon \sim \bnoisedist} [L(\mathtt{GM}(\btheta + \bepsilon), \bm{v}^{*})].
\label{eq:first_approx}
\end{equation*}
The above approximation invokes the reparameterization trick for a complex discrete distribution.  A typical choice for $\rho(\bepsilon)$ is the Gumbel distribution, that is, $\rho(\bepsilon) \sim \mathrm{Gumbel}(0, 1)$ \cite{papandreou2011perturb}. Now, by using a finite-difference approximation of the derivative in the direction of the gradient of the loss $\grad{\bm{\tilde{v}}}L(\bm{\tilde{v}}, \bm{v}^{*}$), we obtain the following estimation rule:
\begin{align}
\nabla_{\btheta} \mathbb{E}_{\bm{\hat{v}}\sim p(\bv; \btheta)}\left[L(\bm{\hat{v}}, \bv^{*})\right] & \approx  \mathbb{E}_{\bepsilon \sim \bnoisedist} \biggl[ \frac{1}{\lambda} \biggl\{
\bm{\tilde{v}} - \mathtt{GM}\left(\btheta + \bepsilon - \lambda \grad{\bm{\tilde{v}}}L(\bm{\tilde{v}}, \bm{v}^{*}) \right) \biggr\} \biggr]\label{eq:imle}, 
\end{align}
\vspace{-0.12in}
\algrenewcommand\algorithmicindent{1em}
\algrenewcommand{\algorithmiccomment}[1]{\bgroup\hskip1em\textcolor{ourspecialtextcolor}{//~\textsl{#1}}\egroup}
\begin{algorithm}[H]
\begin{multicols}{2}
\begin{algorithmic}
       \Function{ForwardPass}{$\bc^{v},\bc^{e}$}
       \\ \Comment{Gumbel noise distribution sampling}
       \State $\epsilon, \epsilon' \sim  \mathrm{Gumbel}(0,1)$
       \\ \Comment{{Graph-matching with perturbed $\left(\bc^{v},\bc^{e}\right)$}}
       \State $\bm{\Tilde{v}} = \mathtt{GM}\left(\bc^{v} + \epsilon, \bc^{e} + \epsilon' \right)$
       \\ \Comment{Save values for the backward pass}
       \Save $\left(\bc^{v}, \bc^{e} \right)$, $(\epsilon, \epsilon')$ and $\bm{\tilde{v}}$
       \Return $\bm{\tilde{v}}$
       \EndFunction
\end{algorithmic}
\begin{algorithmic}
       \Function{BackwardPass}{$\grad{\bm{\tilde{v}}}{L}(\bm{\tilde{v}}, \bv^{*}), \lambda$}
       \Load $\left(\bc^{v}, \bc^{e} \right)$, $(\epsilon, \epsilon')$ and $\bm{\tilde{v}}$
       
       \\ \Comment{Add gradient-based pertubations}
       \State $\left(\bc^{v}_{\lambda}, \bc^{e}_{\lambda}\right) = \left(\bc^v + \epsilon, \bc^{e} + \epsilon' \right) - \lambda \grad{\bm{\tilde{v}}}{L}(\bm{\tilde{v}}, \bv^{*})$       
       \\ \Comment{Single sample gradient estimate}
       \State
       $\left(\frac{\partial L}{\partial \bc^v}, \frac{\partial L}{\partial \bc^e}\right) = \bm{\tilde{v}} - \mathtt{GM}\left(\bc^{v}_{\lambda}, \bc^{e}_{\lambda}\right)$
       \Return \ $\tfrac{1}{\lambda}\left(\frac{\partial L}{\partial \bc^v}, \frac{\partial L}{\partial \bc^e}\right)$
       \EndFunction
\end{algorithmic}
\end{multicols}
\caption{Forward and Backward Pass for $\bm{c}^v$, $\bm{c}^{e}$}
\label{algo:main}
\vspace{-0.12in}
\end{algorithm}
\vspace{-0.1in}
where $\bm{\tilde{v}} = \mathtt{GM}(\btheta + \bepsilon)$,  $\lambda$ is a step size of finite difference approximation. Using a Monte Carlo approximation of the above expectation, the gradient for $\btheta$ is computed as a difference of two or more pairs of perturbed graph-matching outputs. We summarize in Algorithm~\ref{algo:main} the forward and backward steps for $\bm{c^v}, \bm{c^e}$.
\vspace{-0.13in}
\section{Experiments}
\vspace{-0.1in}
\subsection{Implementation details}
\vspace{-0.1in}
\paragraph{Pre-training}
We utilize Resnet50 \citep{he2016deep} and Vision Transformer (ViT-B/16) \cite{dosovitskiy2020image} to train our LVM-Med. For Resnet50, we load pre-trained from ImageNet-1K \cite{deng2009imagenet}, and SAM Encoder backbone weight \cite{kirillov2023segany} 
for ViT. The raw image is augmented to two different views by using multi-crop techniques as \cite{bardes2022vicregl} and followed by flip (probability 50 $\%$), color jitter, random Gaussian blur, and normalization.
We trained the LVM-Med with 100 epochs on the collected dataset. The batch size of $3200$ is used for ResNet50 and we reduced it to $2800$ for ViT due to memory limitation. 
The model is optimized with Adam \cite{kingma2014adam} with an initial learning rate $2\times10^{-3}$ and reduced halved four times. We use $16$ A100-GPUs per with $80$GB and complete the training process for LVM-Med with ResNet-50 in five days and LVM-Med with ViT encoder in seven days. Other competitor SSL methods as VicRegl, Twin-Barlon, Dino, etc, are initialized from ResNet-50 pre-trained ImageNet-1K and trained with $100$ epochs with default settings as LVM-Med.

To balance samples among different modalities, we combine over-sampling and data augmentation to increase the total samples. Specifically, new samples from minority classes are generated by duplicating images and applying random crop operations covering $85-95\%$ of image regions and then rescaling them to the original resolutions. Note that these augmentations are not used in the self-supervised algorithm (operations $s, t \sim T$) to avoid generating identical distorted versions in this sampling procedure.

\begin{minipage}{\textwidth}
  \begin{minipage}[b]{0.65\textwidth}
  \vspace{-0.2in}
    \begin{table}[H]
    \caption{{Summary of datasets and downstream tasks}}
    \label{tab:data-summary}
    \begin{center}
    \vspace{0.01in}
    \setlength{\tabcolsep}{1pt}
    \resizebox{0.95\columnwidth}{!}{
    \begin{tabular}{c|c|c|c|c}
    \toprule
    \textbf{Evaluation}  & \textbf{Downstream Task Data} & \textbf{Modality}         & \textbf{Nums} &       \textbf{Task}                          \\ \midrule
    Fine-Tuning & BraTS2018 \citep{bakas2018identifying}            & 3D MRI           & 285  &       Tumor Segmentation            \\
    Fine-Tuning & MMWHS-CT \citep{zhuang2016multi}            & 3D CT            & 20   &       Heart Structures Segmentation \\
    Fine-Tuning & MMWHS-MRI \citep{zhuang2016multi}           & 3D MRI           & 30   &       Heart Structures Segmentation \\
    Fine-Tuning & ISIC-2018 \citep{codella2019skin}           & 2D Dermoscopy    & 2596 &       Skin Leision Segmentation     \\
    Fine-Tuning & JSRT  \citep{shiraishi2000development}              & 2D X-ray         & 247  &       Multi-Organ Segmentation      \\
    Fine-Tuning & KvaSir \citep{pogorelov2017kvasir}              & 2D Endoscope     & 1000 &      \begin{tabular}[c]{@{}c@{}}Polyp Segmentation \& \\ Detection\end{tabular}            \\
    Fine-Tuning & Drive \citep{staal2004ridge}            & Fundus & 40   &       Vessel Segmentation           \\
    Fine-Tuning & BUID \citep{al2020dataset}                & 2D Ultrasound    & 647  &       Breast Cancer Segmentation    \\
    \begin{tabular}[c]{@{}c@{}}Linear Evaluation \& \\ Fine-Tuning\end{tabular} &
      FGADR \citep{zhou2020benchmark} &
      Fundus &
      1841 &
      DR Grading \\
    \begin{tabular}[c]{@{}c@{}}Linear Evaluation \&\\ Fine-Tuning\end{tabular} &
      Brain Tumor Classification  &
      2D MRI &
      3264 &
      Brain Tumor Classification \\
    Fine-Tuning &
      \begin{tabular}[c]{@{}c@{}}Multi-site Prostate \\ MRI Segmentation \citep{liu2020ms} \end{tabular} &
      3D MRI &
      116 &
      Prostate Segmentation \\ 
      Fine-Tuning &
      VinDr \citep{nguyen2022vindr}  &
      2D X-ray &
      18000 &
      Lung Diseases Detection \\ \bottomrule
    \end{tabular}}
    \end{center}
    \end{table}
    \vspace{0.02in}
  \end{minipage}
  \hfill
  \begin{minipage}[b]{0.3\textwidth}
    \begin{figure}[H]
      \centering
      \includegraphics[width=\textwidth]{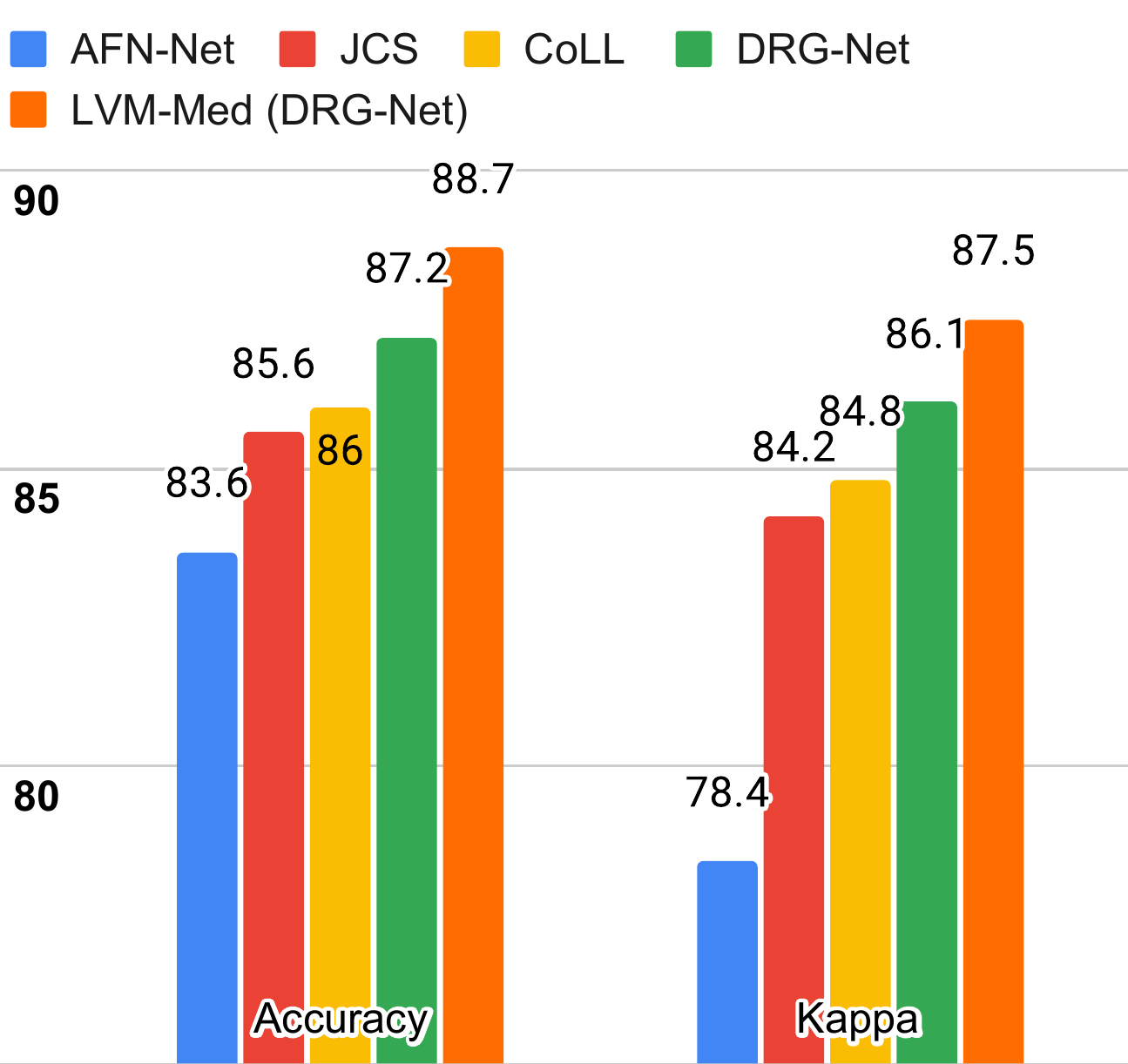}
      \caption{\small{FGADR performance with top architectures.}}
      \label{fig:fgadr_result}
    \end{figure}
  \end{minipage}
\end{minipage}
\vspace{-\abovedisplayskip}
\vspace{-0.1in}
\paragraph{Downstream Tasks}
Table \ref{tab:data-summary} lists the datasets and downstream tasks  used in our experiments. We cover 
segmentation, object detection, and image classification problems. It is important to note that in most settings, we utilize simple configurations for all datasets, skipping extra pre-processing for data augmentation. For instance, overlapping image patches with stride operations in the original samples \cite{kamran2021rv} to increase training data in the Drive dataset or combining different 3D MRI modalities to fuse information \cite{myronenko20193d} in the BRATS-2018 are excluded in our downstream setups. 

To validate LVM-Med algorithms, we compare with 2D-SSL methods trained in our dataset and foundation models like Clip \citep{radford2021learning}, Align \citep{jia2021scaling}, Flava \citep{singh2022flava}, and SAM \cite{kirillov2023segment} with pre-trained ViT (Bert for Align) taken from each method, respectively.
During the downstream task, trained SSL weights are then extracted and attached in U-Net for ResNet50 backbone, TransUnet \cite{chen2021transunet} for ViT, and then fine-tuned with training splits of each dataset. Depending on the downstream task's properties, we apply different image resolutions and other parameters like the number of epochs and learning rate for different data domains. Details for these configurations are presented in Appendix.   
\subsection{2D- and 3D-based segmentation}
\vspace{-2mm}
We evaluate LVM-Med on \textit{eight} medical segmentation tasks, including five 2D-based and three 3D-based segmentation. In 2D settings, we also compare with 2D supervised architectures, such as U-Net, U-Net++, Attention U-Net, etc. These networks are initialized with ResNet-50 pre-trained ImageNet. Additionally, we investigate the prompt-based segmentation settings inspired by the current success of SAM's zero-shot learning. We utilized the ground truths and added random noise to simulate box-based user prompts as \cite{ma2023segment}. 
 We next compare three variations of SAM: (i) freezing image and prompt encoders, only fine-tuning mask decoder; (ii) without any training and inference using box prompts; (iii) similar to (i) but replacing the original image encoder by LVM-Med's ViT architecture taken from SAM trained in our dataset.
 \vspace{-0.1in}
\begin{table}[H]
\begin{center}
\caption{{Performance comparison on five 2D segmentation tasks with fully fine-tuning. Results are reported with an average 2D Dice score on three trial times. The best results in each group are in
bold, the overall best value, excluding prompt-based segmentation, is underlined.}}
\vspace{0.01in}
\label{tab:2Dsegmentation}
\resizebox{0.95\columnwidth}{!}{
\begin{tabular}{l|l|ccccc}
\toprule
\textbf{} &
  {\textbf{Method}} &
  \textbf{ISIC-2018 (Skin Lesion)} &
  \textbf{JSRT  (Lung X-ray)} &
  \textbf{KvaSir (Polyp)} &
  \textbf{Drive (Vessel)} &
  \textbf{BUID (Breast Cancer)} \\ \midrule
 &
  Randomly (R50) &
  86.16 $\pm$ 0.14 &
  93.10 $\pm$ 0.12 &
  62.85 $\pm$ 1.32 &
  59.82 $\pm$ 2.00 &
  65.54 $\pm$ 0.21 \\
 &
  Pre-trained ImageNet \citep{he2016deep} &
  \textbf{86.87 $\pm$ 0.47} &
  \textbf{94.52 $\pm$ 2.66} &
  \textbf{83.85 $\pm$ 1.32} &
  65.12 $\pm$ 1.55 &
  72.64 $\pm$ 1.14 \\
 &
  Attention-Unet \citep{oktay2018attention} &
  86.81 $\pm$ 0.51 &
  94.47 $\pm$ 2.71 &
  82.23 $\pm$ 1.41 &
  65.02 $\pm$ 1.44 &
  72.19 $\pm$ 1.16 \\
 &
  U-Net ++ \citep{zhou1807nested} &
  86.71 $\pm$ 0.49 &
  94.32 $\pm$ 2.81 &
  82.23 $\pm$ 1.41 &
  \textbf{65.38 $\pm$ 0.78} &
  \textbf{73.76 $\pm$ 2.83} \\
\multirow{-5}{*}{\textbf{2D Supervised   Method}} &
  Trans U-Net \citep{chen2021transunet} &
  86.60 $\pm$ 0.82 &
  89.80 $\pm$ 0.35 &
   67.11 $\pm$ 0.24 &
  62.63 $\pm$ 0.24 &
  67.90 $\pm$ 0.40
   \\ \midrule
 &
  Twin-Barlon \cite{zbontar2021barlow}&
  86.01 $\pm$ 0.07 &
  94.56 $\pm$ 3.09 &
  83.00 $\pm$ 0.23 &
  65.73 $\pm$ 1.46 &
  74.46 $\pm$ 1.19 \\
 &
  Dino  \cite{caron2021emerging}&
  86.79 $\pm$ 0.09 &
  94.84 $\pm$ 2.79 &
  79.84 $\pm$ 1.62 &
  65.39 $\pm$ 0.81 &
  76.21 $\pm$ 0.57 \\
 &
  SimCLR \cite{chen2020simple} &
  87.28 $\pm$ 0.21 &
  94.79 $\pm$ 2.93 &
  82.20 $\pm$ 0.51 &
  65.22 $\pm$ 2.18 &
  76.52 $\pm$ 0.22 \\
 &
  Moco-v2 \citep{chen2020improved}&
  87.24 $\pm$ 0.14 &
  94.05 $\pm$ 3.52 &
  78.24 $\pm$ 1.35 &
  64.92 $\pm$ 2.21 &
  75.93 $\pm$ 1.96 \\
 &
  Deepcluster-v2 \citep{caron2018deep}&
  86.73 $\pm$ 0.42 &
  94.79 $\pm$ 2.89 &
  82.69 $\pm$ 0.75 &
  64.14 $\pm$ 0.92 &
  76.33 $\pm$ 0.99 \\
 &
  VicRegl \citep{bardes2022vicregl}&
  86.27 $\pm$ 0.33 &
  94.39 $\pm$ 3.25 &
  81.93 $\pm$ 0.48 &
  66.17 $\pm$ 0.27 &
  75.29 $\pm$ 0.64 \\
\multirow{-7}{*}{\textbf{2D-SSL on medical}} &
  \textbf{LVM-Med (R50)} &
  \textbf{87.76 $\pm$ 0.30} &
  \textbf{\underline{95.13} $\pm$ 2.64} &
  \textbf{\underline{86.76} $\pm$ 0.94} &
  \textbf{\underline{66.97} $\pm$ 0.27} &
  \textbf{\underline{78.65} $\pm$ 0.72} \\ \midrule
 &
  Clip \citep{radford2021learning}&
  85.98 $\pm$ 0.19 &
  89.00 $\pm$ 1.08 &
  72.63 $\pm$ 0.37 &
  63.01 $\pm$ 0.36 &
  70.43 $\pm$ 0.24 \\
 &
  Flava \citep{singh2022flava}&
  86.42 $\pm$ 0.10 &
  90.08 $\pm$ 0.20 &
  69.47 $\pm$ 0.05 &
  61.09 $\pm$ 0.45 &
  67.54 $\pm$ 1.17 \\
 &
  SAM \citep{kirillov2023segment}&
  88.17 $\pm$ 0.30 &
  90.68 $\pm$ 0.40 &
  70.75 $\pm$ 0.60 &
  64.04 $\pm$ 0.41 &
  73.07 $\pm$ 0.66 \\
\multirow{-4}{*}{\textbf{Foundation Model}} &
  \textbf{LVM-Med (SAM's ViT)} &
  \textbf{\underline{88.41} $\pm$ 0.28} &
  \textbf{90.74 $\pm$ 0.47} &
  \textbf{73.10 $\pm$ 0.08} &
  \textbf{65.49 $\pm$ 0.12} &
  \textbf{77.20 $\pm$ 0.42} \\ \midrule
{ } &
  SAM (fixed encoder) \citep{ma2023segment} &
  92.42 $\pm$ 0.12 &
  92.89 $\pm$ 5.24 &
  89.37 $\pm$ 0.57 &
  59.74 $\pm$ 0.63 &
  87.63 $\pm$ 0.67 \\
{ } &
  SAM with Prompt (no-train) \citep{kirillov2023segment} &
   55.78 $\pm$ 0.66 &
   61.97 $\pm$ 4.48 &
   80.77 $\pm$ 0.19 &
   15.12 $\pm$ 0.24 &
   78.44 $\pm$ 1.01 \\
\multirow{-3}{*}{{\textbf{Prompt-based Seg.}}} &
  \textbf{LVM-Med (SAM's ViT)} &
  \textbf{92.48 $\pm$ 0.07} &
  \textbf{93.74 $\pm$ 4.06} &
  \textbf{90.09 $\pm$ 0.14} &
  \textbf{63.01 $\pm$ 0.02} &
  \textbf{89.69 $\pm$ 0.61} \\ \bottomrule
\end{tabular}}
\end{center}
\end{table}
\vspace{-0.35in}
\begin{minipage}{\textwidth}
  \begin{minipage}[t]{0.43\textwidth}
    \begin{table}[H]
\begin{center}
\caption{{3D segmentation task
performance with fine-tuning on three datasets. Results are reported with an average 3D IoU on five trial times. The best results in each group and overall are in bold and underlined.}}
\vspace{0.05in}
\label{tab:3D_segmentation}
\resizebox{\columnwidth}{!}{
\begin{tabular}{l|ccc}
\toprule
  \textbf{Method} &
  \textbf{BraTS} &
  \textbf{MMWHS-CT} &
  \textbf{MMWHS-MRI} \\ \midrule
  3D-Transformer \citep{hatamizadeh2022unetr}                  & 66.54 $\pm$ 0.40          & 67.30 $\pm$ 2.29          & 67.64 $\pm$ 2.21          \\
  I3D \citep{carreira2017quo}                        & 67.83 $\pm$ 0.75          & 76.63 $\pm$ 2.32           & 66.71 $\pm$ 1.27          \\
  NiftyNet \citep{gibson2018niftynet}                       & 60.78 $\pm$ 1.60          & 74.91 $\pm$ 2.78          & 64.60 $\pm$ 1.96          \\
  Med3D \citep{chen2019med3d}                           & 66.09 $\pm$ 1.35          & 75.01 $\pm$ 0.74          & 63.43 $\pm$ 0.61          \\
  Model Genesis \citep{zhou2021models}                    & 67.96 $\pm$ 1.29          & 76.48 $\pm$ 2.89          & 74.53 $\pm$ 1.69          \\
  Universal Model  \citep{Zhang2020UniversalMF}                & 72.10 $\pm$ 0.67          & 78.14 $\pm$ 0.77          & 77.52 $\pm$ 0.50          \\
  TransVW \citep{haghighi2021transferable}                          & 68.82 $\pm$ 0.38          & 79.74 $\pm$ 2.78          & 75.08 $\pm$ 2.04          \\
  SwinViT3D \citep{tang2022self}                      & 70.58 $\pm$ 1.27          & 70.19 $\pm$ 1.23          & \textbf{78.25 $\pm$ 1.66} \\
  Joint-2D-3D (Deepc)  \citep{nguyen2022joint}   & \textbf{72.81 $\pm$ 0.15}          & \textbf{83.58 $\pm$ 1.54}          & 78.14 $\pm$ 1.32          \\ \midrule
  Twin-Barlon \cite{zbontar2021barlow}&
  73.30 $\pm$ 0.18 &
  84.74 $\pm$ 1.01 &
  76.39 $\pm$ 2.23 \\
  Dino  \cite{caron2021emerging}                           & 71.72 $\pm$ 0.55          & 81.08 $\pm$ 1.62          & 70.42 $\pm$ 78.74          \\
  SimCLR \cite{chen2020simple}                          & 73.15 $\pm$ 0.27          & 84.60 $\pm$ 1.11          & 76.54 $\pm$ 2.22          \\
  Moco-v2  \citep{chen2020improved}                        & 71.97 $\pm$ 0.63          & 75.82 $\pm$ 4.20          & 68.29 $\pm$ 0.15          \\
  Deepcluster  \citep{caron2018deep}                   & 72.96 $\pm$ 0.51          & 84.03 $\pm$ 0.50          & \textbf{\underline{79.05} $\pm$ 1.63} \\
  VicRegl \citep{bardes2022vicregl}                       & 73.23 $\pm$ 0.33          & 84.72 $\pm$ 0.86          & 76.32 $\pm$ 0.78           \\
 
  \textbf{LVM-Med (R50)} &
  \textbf{\underline{73.58} $\pm$ 0.14} &
  \textbf{\underline{84.91} $\pm$ 0.77} &
  78.59 $\pm$ 0.84 \\ \midrule
  Clip \citep{radford2021learning}&
  70.24 $\pm$ 1.23 &
  78.5 $\pm$ 2.70 &
  65.9 $\pm$ 3.98 \\
  Flava \citep{singh2022flava}           & 71.19 $\pm$ 0.48          & 78.91 $\pm$ 2.24        & 67.14 $\pm$ 1.20          \\
  SAM (Encoder)  \citep{kirillov2023segment}                 & 70.11 $\pm$ 1.45          & 77.8 $\pm$ 1.60            & 68.09 $\pm$ 5.49          \\
  \textbf{LVM-Med (SAM's ViT)} & \textbf{71.42 $\pm$ 0.70}                    & \textbf{80.78 $\pm$ 1.77}                     & \textbf{69.36 $\pm$ 0.18}                    \\ \bottomrule
\end{tabular}}
\end{center}
\end{table}
    \vspace{0.01in}
  \end{minipage}
  \hfill
  \begin{minipage}[t]{0.55\textwidth}
   \vspace{0.3in}
\begin{table}[H]
\begin{center}
\caption{{In-out-distribution evaluation for the segmentation task on the Prostate dataset. Results are reported with an average 2D Dice score on three training times.}}
\vspace{0.05in}
\label{tab:in-out-distribution}
\resizebox{\columnwidth}{!}{
\begin{tabular}{l|ccccc}
\toprule
  \textbf{Method} &
  \multicolumn{5}{c}{\textbf{Multi-site Prostate Segmentation}} \\ \midrule
  &
  \textbf{BMC (Based)} &
  \textbf{RUNMC} &
  \textbf{BIDMC} &
  \textbf{HK} &
  \textbf{Average} \\ \midrule
  \textbf{\textit{2D Supervised}} &   &   &  &  & \\
  Random &
  65.04 $\pm$ 2.07 &
  51.44 $\pm$ 4.13 &
  9.95 $\pm$ 13.56 &
  12.38 $\pm$ 7.68 &
  34.7  \\
 Pretrained ImageNet \citep{he2016deep}  &
  \textbf{76.47 $\pm$ 1.26} &
  \textbf{62.11 $\pm$ 0.85} &
  \textbf{43.74 $\pm$ 4.38} &
  \textbf{\underline{53.90} $\pm$ 2.01} &
  \textbf{59.1}  \\ \midrule
  \textbf{\textit{2D SSL on medical data}} &   &   &  &  & \\
  Twin-Barlon \cite{zbontar2021barlow}&
  76.28 $\pm$ 1.76 &
  60.09 $\pm$ 1.98 &
  32.63 $\pm$ 12.32 &
  34.82 $\pm$ 15.09 &
  51.0  \\

  Dino \cite{caron2021emerging}&
  77.90 $\pm$ 1.15 &
  56.90 $\pm$ 1.97 &
  21.53 $\pm$ 5.54 &
  30.92 $\pm$ 5.41 &
  46.8  \\
 
  SimCLR \cite{chen2020simple}&
  76.51 $\pm$ 2.07 &
  64.10 $\pm$ 4.53 &
  32.88 $\pm$ 5.43 &
  42.29 $\pm$ 5.98 &
  53.9  \\
 
  Moco-v2 \citep{chen2020improved}&
  74.40 $\pm$ 0.89 &
  55.49 $\pm$ 5.45 &
  27.53 $\pm$ 10.18 &
  13.65 $\pm$ 14.33 &
  42.8  \\
 
  Deepcluster \citep{caron2018deep}&
  77.45 $\pm$ 0.35 &
  \textbf{\underline{64.35} $\pm$ 3.15} &
  37.73 $\pm$ 8.08 &
  44.95 $\pm$ 8.57 &
  56.1  \\
 
  Swav \citep{caron2020unsupervised}&
  77.59 $\pm$ 0.61 &
  57.61 $\pm$ 2.16 &
  38.43 $\pm$ 12.55 &
  44.90 $\pm$ 4.78 &
  54.6  \\
 
  VicRegl \citep{bardes2022vicregl}&
  74.85 $\pm$ 1.13 &
  54.09 $\pm$ 4.35 &
  25.56 $\pm$ 5.44 &
  35.45 $\pm$ 13.03 &
  47.5  \\
 
  \textbf{LVM-Med (R50)} &
  \textbf{\underline{80.17} $\pm$ 0.55} &
  62.48 $\pm$ 2.03 &
  \textbf{\underline{56.76} $\pm$ 6.50} &
  \textbf{52.78 $\pm$ 3.04} &
  \textbf{\underline{63.0}}  \\ \midrule
  \textbf{\textit{Prompt-based Seg.}} & & & & & \\
  SAM (Fixed encoder) \citep{ma2023segment} &
  95.50 $\pm$ 0.29 &
  90.39 $\pm$ 0.39 &
  91.41 $\pm$ 0.14 &
  91.82 $\pm$ 0.26 &
  92.28  \\
 
  SAM with Prompt (no-train) \citep{kirillov2023segment} &
  59.11 $\pm$ 1.55 &
  66.95 $\pm$ 2.49 &
  59.68 $\pm$ 0.49 &
  57.41 $\pm$ 2.83 &
  60.79  \\
 
  \textbf{LVM-Med (SAM's ViT)} &
  \textbf{95.75 +- 0.06} &
  \textbf{90.40 +- 0.36} &
  \textbf{92.03 +- 0.20} &
  \textbf{92.75 +- 0.48} &
  \textbf{92.73} \\ \bottomrule
\end{tabular}}
\end{center}
\end{table}
  \end{minipage}
\end{minipage}

In 3D settings, we segment 2D slices and merge results for a 3D volume. We also benchmarked with 3D self-supervised methods from \cite{nguyen2022joint}. Tables \eqref{tab:2Dsegmentation} and \eqref{tab:3D_segmentation} show that 
our two versions with ResNet-50 and Sam's ViT hold the best records in each category. For instance, we outperform 2D SSL methods trained on the same dataset, surpassing foundation models such as SAM, Flava, and Clip. In the prompt-based settings, LVM-Med also delivers better performance compared with SAM. Second, LVM-Med achieves the best overall results on \textit{seven of eight segmentation tasks}, mostly held by LVM-Med with ResNet-50. The improvement gaps vary on each dataset, 
for e.g., from $3-5\%$ on Kvasir and BUID compared with 2D supervised methods.
\vspace*{-1mm}
\subsection{Linear and finetuning image classification}
\vspace{-0.1in}
We analyze LVM-Med on image classification tasks using linear probing (frozen encoders) and fully fine-tuning settings, two popular evaluations used in self-supervised learning. The experiments are conducted on the FGADR Grading and Brain tumor classification tasks. Table \eqref{tab:classification} presents the average accuracy metric on three training times. 
LVM-Med (ResNet-50) consistently outperforms other approaches on two datasets. For example, it is better than Clip by $10.46\%$ and $8.46\%$ on FGADR and Brain Tumor datasets with linear evaluation. In the foundation model setting, LVM-Med (ViT) also improves SAM's results by $7.32\%$ and $4.69\%$ on FGADR with linear and fully-finetuning. Another point we observe is that the overall 2D-SSL methods based on ResNet-50 and trained on the collected medical dataset achieve higher accuracy than foundation models using ViT. We also compare LVM-Med with the top methods on the FGADR dataset, including AFN-Net \cite{lin2018framework}, JCS \cite{wu2021jcs}, CoLL \cite{zhou2019collaborative}, and DRG-Net \cite{tusfiqur2022drg}. We choose the DRG-Net as the backbone and replace the employed encoder with our weights (R50). Figure \eqref{fig:fgadr_result} shows that LVM-Med hold the first rank overall.
\vspace{-0.25in}
\begin{minipage}{\textwidth}
  \begin{minipage}[t]{0.6\textwidth}
    \begin{table}[H]
        \caption{{Performance comparison on linear evaluation and fine-tuning classification. The results are reported with average Accuracy on three training times. }}
        \label{tab:classification}
        \begin{center}
        \resizebox{1.0\columnwidth}{!}{
        \begin{tabular}{l|cccc}
            \toprule
    
              \textbf{Method} &
              \multicolumn{2}{c}{\textbf{Linear Evaluation (Frozen)}} &
              \multicolumn{2}{c}{\textbf{Fine-tuning}} \\ \midrule
            \textit{} &
              \textbf{FGADR} &
              \textbf{Brain Tumor Cls.} &
              \textbf{FGADR} &
              \textbf{Brain Tumor Cls.} \\ \midrule
            Twin-Barlon \cite{zbontar2021barlow}&
  66.86 $\pm$ 0.41 &
  63.03 $\pm$ 0.32 &
  66.37 $\pm$ 0.77 &
  74.20 $\pm$ 1.38 \\
 
  Dino \cite{caron2021emerging}&
  65.98 $\pm$ 1.91 &
  62.27 $\pm$ 0.32 &
  67.35 $\pm$ 1.36 &
  71.91 $\pm$ 1.55 \\
 
  SimCLR \cite{chen2020simple}&
  65.30 $\pm$ 1.70 &
  62.52 $\pm$ 1.67 &
  67.55 $\pm$ 0.28 &
  73.52 $\pm$ 3.56 \\
 
  Moco-v2 \citep{chen2020improved}&
  65.98 $\pm$ 1.04 &
  62.35 $\pm$ 1.92 &
  67.55 $\pm$ 1.79 &
  74.53 $\pm$ 0.43 \\
 
  Deepcluster \citep{caron2018deep}&
  65.34 $\pm$ 1.93 &
  64.47 $\pm$ 0.55 &
  67.94 $\pm$ 1.78 &
  73.10 $\pm$ 0.55 \\
 
  VicRegl \citep{bardes2022vicregl}&
  64.71 $\pm$ 0.60 &
  59.64 $\pm$ 1.36 &
  65.69 $\pm$ 1.46 &
  73.18 $\pm$ 2.03 \\
 
  \textbf{LVM-Med (R50)} &
  \textbf{\underline{68.33} $\pm$ 0.48}  &
  \textbf{\underline{66.33} $\pm$ 0.31}  &
  \textbf{\underline{68.32} $\pm$ 0.48}  &
  \textbf{\underline{76.82} $\pm$ 2.23} \\ \midrule

  Clip \citep{radford2021learning}&
  57.87 $\pm$ 0.50 &
  57.87 $\pm$ 0.71 &
  57.48 $\pm$ 0.86 &
  34.86 $\pm$ 2.27  \\
 
  Flava \citep{singh2022flava}&
  31.87 $\pm$ 0.69 &
  35.19 $\pm$ 0.43  &
  57.18 $\pm$ 0.96  &
  34.01 $\pm$ 5.97  \\
 
  Algin \citep{jia2021scaling}&
  36.95 $\pm$ 1.04  &
  30.71 $\pm$ 2.35  &
  57.28 $\pm$ 0.97  &
  63.96 $\pm$ 0.04\\
 
  SAM \citep{kirillov2023segment}&
  55.13 $\pm$ 0.41  &
  31.81 $\pm$ 4.26  &
  58.75 $\pm$ 1.32  &
  60.66 $\pm$ 1.36  \\
 
  \textbf{LVM-Med (SAM's ViT)} &
  \textbf{62.46 $\pm$ 0.86}  &
  \textbf{59.31 $\pm$ 0.48}  &
  \textbf{63.44 $\pm$ 0.73}  &
  \textbf{67.34 $\pm$ 2.08}  \\ \bottomrule
\end{tabular}}
\end{center}
\end{table}
  \end{minipage} \ 
  \begin{minipage}[t]{0.37\textwidth}
    \vspace{0.3in}
    \begin{table}[H]
        \centering
\caption{{LVM-Med ablation study. Results are reported on an average of five 2D segmentation and two linear classification tasks. The two most important factors are highlighted.}}
\label{tab:ablation-study}
\vspace{2mm}
\resizebox{\columnwidth}{!}{
\begin{tabular}{ @{\hspace{-0pt}}l@{\hspace{8pt}}c@{\hspace{8pt}}c@{\hspace{10pt}}c}
\toprule
Method         & Cls.(Acc)  & Seg. (Dice)\\
\midrule
LVM-Med (Full)                    & \textbf{67.47}     & \textbf{83.05} \\
\rowcolor{cyan!50}
LVM-Med w/o second-order                   &  62.17  & 80.21  \\
LVM-Med w/o message passing              &  65.08   & 81.19  \\
\rowcolor{cyan!50}
LVM-Med w/o Gumbel noise             &  64.32   & 81.37 \\
LVM-Med w/o local similarity             &  65.67   & 81.54 \\
\bottomrule
\end{tabular}}
    \end{table}
  \end{minipage}
\end{minipage}


\subsection{Object detection \& In-out-distribution evaluation}
\vspace*{-3mm}
\begin{wrapfigure}{r}{0.48\textwidth}
    \centering
    \vspace*{-6mm}
    \hspace{6mm}
    \includegraphics[width=0.4\textwidth]{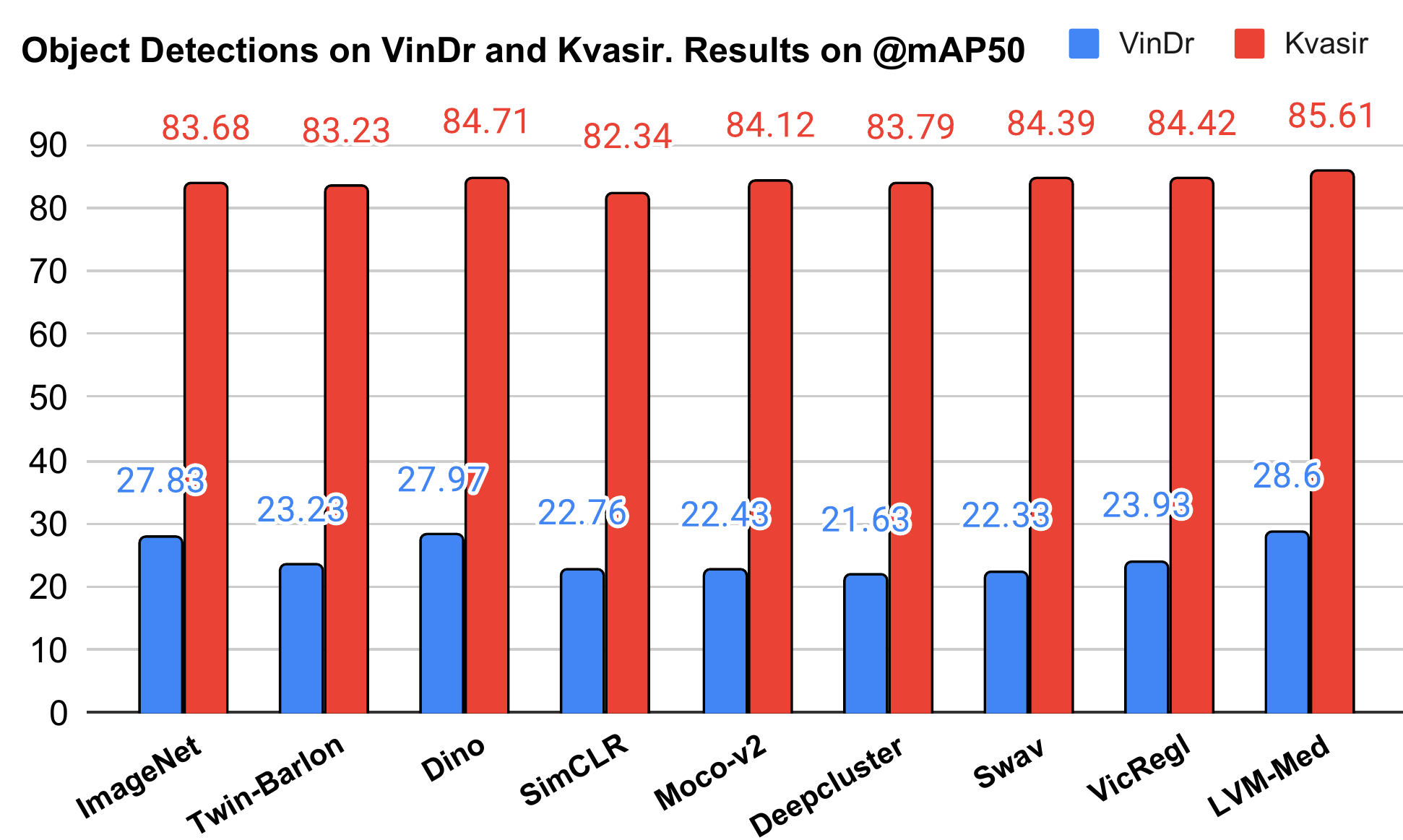}   \captionsetup{width=0.38\textwidth}
    \vspace*{-3mm}
    \caption{\small{LVM-Med on object detection.}}
    \vspace*{-5mm}
    \label{fig:object_det_results}
\end{wrapfigure}
Figure \ref{fig:object_det_results} indicates our performance on the object detection task using VinDr and Kvasir datasets. We use Faster R-CNN and load ResNet-50 from 2D SSL pre-trained weights. Results are presented by Average Precision with IoU=0.5 over three training times. Compared to pre-trained Imagenet, LVM-Med still outperforms by $1$-$2\%$ though overall, our improvements are smaller than image classification and segmentation tasks. 

We also validate LVM-Med performance on the in-out-distribution setting in Table \eqref{tab:in-out-distribution} using the segmentation task on  the Multi-Prostate dataset. We train LVM-Med and other competitors in BMC data and use the trained models to predict the remaining datasets. Both two versions of LVM-Med with ResNet-50 and ViT, on average, surpass all baselines, which validates the potential abilities of LVM-Med for the in-out-distribution problem.
\vspace{-0.05in}
\subsection{Ablation study}
\vspace{-0.05in}
We do the following settings to evaluate the performance of components used in LVM-Med: (i) LVM-Med without using second-order graph matching conditions, i.e., only solving vertex-to-vertex correspondence problem; (ii) LVM-Med without using message passing network $g_{\epsilon}$ in Eq. \eqref{eq:gnn} to aggregate information from local connections; (iii) LVM-Med w/o using approximate gradients from Gumbel noise in Eq. \eqref{eq:imle}. For this, we add a constant value to $\bm{c}^{v}, \bm{c}^e$ as prior works \citep{rolinek2020optimizing,rolinek2020deep}, and finally (iv) LMV-Med without using local similarity $c_{ia}^{lo}$ in Eq. \eqref{eq:local_cost}. Other ablation studies are presented in Appendix.
Table \eqref{tab:ablation-study} indicates that all factors contribute to the final performance, wherein the second-order and Gumbel noise are the two most two important parts.

\vspace{-0.1in}
\section{Conclusion}
We have demonstrated that a self-supervised learning technique based on  second-order graph-matching, trained on a large-scale medical imaging dataset, significantly enhances performance in various downstream medical imaging tasks compared to other supervised learning methods and foundation models trained on hundreds of millions of image-text instances. Our findings are supported by the benefits shown in two different architectures: ResNet-50 and ViT backbones, which can be used for either end-to-end or prompt-based segmentation.

\textbf{Limitations and Future Work.} We propose to investigate the following points to improve LVM-Med performance. Firstly, extending LVM-Med to a hybrid 2D-3D architecture to allow direct application for 3D medical tasks instead of 2D slices. Secondly, although LVM-Med with ViT backbone utilizes more total parameters, in some cases, it is less effective than LVM-Med ResNet-50. This raises the question of whether a novel approach could improve the performance of ViT architectures. Finally, integrating multi-modal information such as knowledge graphs, bio-text, or electronic health records for LVM-Med is also important to make the model more useful in real-world applications. 

\section*{Acknowledgements}
This research has been supported by the pAItient project (BMG, 2520DAT0P2), Ophthalmo-AI project (BMBF, 16SV8639) and the Endowed Chair of Applied Artificial Intelligence, Oldenburg University. Binh T. Nguyen wants to thank the University of Science, Vietnam National University in Ho Chi Minh City for their support. Tan Ngoc Pham would like to thank the Vingroup Innovation Foundation (VINIF) for the Master's training scholarship program. The authors thank the International Max Planck Research School for Intelligent Systems (IMPRS-IS) for supporting Duy M. H. Nguyen. Mathias Niepert acknowledges funding by Deutsche Forschungsgemeinschaft (DFG, German Research Foundation) under Germany’s Excellence Strategy - EXC and support by the Stuttgart Center for Simulation Science (SimTech).
\bibliography{references}

\begin{thebibliography}{149}
\providecommand{\natexlab}[1]{#1}
\providecommand{\url}[1]{\texttt{#1}}
\expandafter\ifx\csname urlstyle\endcsname\relax
  \providecommand{\doi}[1]{doi: #1}\else
  \providecommand{\doi}{doi: \begingroup \urlstyle{rm}\Url}\fi

\bibitem[Cheplygina et~al.(2019)Cheplygina, de~Bruijne, and
  Pluim]{cheplygina2019not}
Veronika Cheplygina, Marleen de~Bruijne, and Josien~PW Pluim.
\newblock Not-so-supervised: a survey of semi-supervised, multi-instance, and
  transfer learning in medical image analysis.
\newblock \emph{Medical image analysis}, 54:\penalty0 280--296, 2019.

\bibitem[Kaissis et~al.(2020)Kaissis, Makowski, R{\"u}ckert, and
  Braren]{kaissis2020secure}
Georgios~A Kaissis, Marcus~R Makowski, Daniel R{\"u}ckert, and Rickmer~F
  Braren.
\newblock Secure, privacy-preserving and federated machine learning in medical
  imaging.
\newblock \emph{Nature Machine Intelligence}, 2\penalty0 (6):\penalty0
  305--311, 2020.

\bibitem[Radford et~al.(2021)Radford, Kim, Hallacy, Ramesh, Goh, Agarwal,
  Sastry, Askell, Mishkin, Clark, et~al.]{radford2021learning}
Alec Radford, Jong~Wook Kim, Chris Hallacy, Aditya Ramesh, Gabriel Goh,
  Sandhini Agarwal, Girish Sastry, Amanda Askell, Pamela Mishkin, Jack Clark,
  et~al.
\newblock Learning transferable visual models from natural language
  supervision.
\newblock In \emph{International conference on machine learning}, pages
  8748--8763. PMLR, 2021.

\bibitem[Jia et~al.(2021)Jia, Yang, Xia, Chen, Parekh, Pham, Le, Sung, Li, and
  Duerig]{jia2021scaling}
Chao Jia, Yinfei Yang, Ye~Xia, Yi-Ting Chen, Zarana Parekh, Hieu Pham, Quoc Le,
  Yun-Hsuan Sung, Zhen Li, and Tom Duerig.
\newblock Scaling up visual and vision-language representation learning with
  noisy text supervision.
\newblock In \emph{International Conference on Machine Learning}, pages
  4904--4916. PMLR, 2021.

\bibitem[Singh et~al.(2022)Singh, Hu, Goswami, Couairon, Galuba, Rohrbach, and
  Kiela]{singh2022flava}
Amanpreet Singh, Ronghang Hu, Vedanuj Goswami, Guillaume Couairon, Wojciech
  Galuba, Marcus Rohrbach, and Douwe Kiela.
\newblock Flava: A foundational language and vision alignment model.
\newblock In \emph{Proceedings of the IEEE/CVF Conference on Computer Vision
  and Pattern Recognition}, pages 15638--15650, 2022.

\bibitem[Kirillov et~al.(2023{\natexlab{a}})Kirillov, Mintun, Ravi, Mao,
  Rolland, Gustafson, Xiao, Whitehead, Berg, Lo, et~al.]{kirillov2023segment}
Alexander Kirillov, Eric Mintun, Nikhila Ravi, Hanzi Mao, Chloe Rolland, Laura
  Gustafson, Tete Xiao, Spencer Whitehead, Alexander~C Berg, Wan-Yen Lo, et~al.
\newblock Segment anything.
\newblock \emph{arXiv preprint arXiv:2304.02643}, 2023{\natexlab{a}}.

\bibitem[Mazurowski et~al.(2023)Mazurowski, Dong, Gu, Yang, Konz, and
  Zhang]{mazurowski2023segment}
Maciej~A Mazurowski, Haoyu Dong, Hanxue Gu, Jichen Yang, Nicholas Konz, and
  Yixin Zhang.
\newblock Segment anything model for medical image analysis: an experimental
  study.
\newblock \emph{arXiv preprint arXiv:2304.10517}, 2023.

\bibitem[He et~al.(2023)He, Bao, Li, Grant, and Ou]{he2023accuracy}
Sheng He, Rina Bao, Jingpeng Li, P~Ellen Grant, and Yangming Ou.
\newblock Accuracy of segment-anything model (sam) in medical image
  segmentation tasks.
\newblock \emph{arXiv preprint arXiv:2304.09324}, 2023.

\bibitem[Ma and Wang(2023)]{ma2023segment}
Jun Ma and Bo~Wang.
\newblock Segment anything in medical images.
\newblock \emph{arXiv preprint arXiv:2304.12306}, 2023.

\bibitem[Kirillov et~al.(2023{\natexlab{b}})Kirillov, Mintun, Ravi, Mao,
  Rolland, Gustafson, Xiao, Whitehead, Berg, Lo, Doll{\'a}r, and
  Girshick]{kirillov2023segany}
Alexander Kirillov, Eric Mintun, Nikhila Ravi, Hanzi Mao, Chloe Rolland, Laura
  Gustafson, Tete Xiao, Spencer Whitehead, Alexander~C. Berg, Wan-Yen Lo, Piotr
  Doll{\'a}r, and Ross Girshick.
\newblock Segment anything.
\newblock \emph{arXiv:2304.02643}, 2023{\natexlab{b}}.

\bibitem[Sun et~al.(2020)Sun, Zhou, and Fei]{sun2020survey}
Hui Sun, Wenju Zhou, and Minrui Fei.
\newblock A survey on graph matching in computer vision.
\newblock In \emph{2020 13th International Congress on Image and Signal
  Processing, BioMedical Engineering and Informatics (CISP-BMEI)}, pages
  225--230. IEEE, 2020.

\bibitem[Haller et~al.(2022)Haller, Feineis, Hutschenreiter, Bernard, Rother,
  Kainm{\"u}ller, Swoboda, and Savchynskyy]{haller2022comparative}
Stefan Haller, Lorenz Feineis, Lisa Hutschenreiter, Florian Bernard, Carsten
  Rother, Dagmar Kainm{\"u}ller, Paul Swoboda, and Bogdan Savchynskyy.
\newblock A comparative study of graph matching algorithms in computer vision.
\newblock In \emph{Computer Vision--ECCV 2022: 17th European Conference, Tel
  Aviv, Israel, October 23--27, 2022, Proceedings, Part XXIII}, pages 636--653.
  Springer, 2022.

\bibitem[Zbontar et~al.(2021)Zbontar, Jing, Misra, LeCun, and
  Deny]{zbontar2021barlow}
Jure Zbontar, Li~Jing, Ishan Misra, Yann LeCun, and St{\'e}phane Deny.
\newblock Barlow twins: Self-supervised learning via redundancy reduction.
\newblock In \emph{International Conference on Machine Learning}, pages
  12310--12320. PMLR, 2021.

\bibitem[Bardes et~al.(2022)Bardes, Ponce, and LeCun]{bardes2022vicregl}
Adrien Bardes, Jean Ponce, and Yann LeCun.
\newblock Vicregl: Self-supervised learning of local visual features.
\newblock In \emph{NeurIPS}, 2022.

\bibitem[Chen et~al.(2020{\natexlab{a}})Chen, Kornblith, Norouzi, and
  Hinton]{chen2020simple}
Ting Chen, Simon Kornblith, Mohammad Norouzi, and Geoffrey Hinton.
\newblock A simple framework for contrastive learning of visual
  representations.
\newblock In \emph{International conference on machine learning}, pages
  1597--1607. PMLR, 2020{\natexlab{a}}.

\bibitem[He et~al.(2020)He, Fan, Wu, Xie, and Girshick]{he2020momentum}
Kaiming He, Haoqi Fan, Yuxin Wu, Saining Xie, and Ross Girshick.
\newblock Momentum contrast for unsupervised visual representation learning.
\newblock In \emph{Proceedings of the IEEE/CVF conference on computer vision
  and pattern recognition}, pages 9729--9738, 2020.

\bibitem[Chen et~al.(2020{\natexlab{b}})Chen, Fan, Girshick, and
  He]{chen2020improved}
Xinlei Chen, Haoqi Fan, Ross Girshick, and Kaiming He.
\newblock Improved baselines with momentum contrastive learning.
\newblock \emph{arXiv preprint arXiv:2003.04297}, 2020{\natexlab{b}}.

\bibitem[Tomasev et~al.(2022)Tomasev, Bica, McWilliams, Buesing, Pascanu,
  Blundell, and Mitrovic]{tomasev2022pushing}
Nenad Tomasev, Ioana Bica, Brian McWilliams, Lars Buesing, Razvan Pascanu,
  Charles Blundell, and Jovana Mitrovic.
\newblock Pushing the limits of self-supervised resnets: Can we outperform
  supervised learning without labels on imagenet?
\newblock \emph{arXiv preprint arXiv:2201.05119}, 2022.

\bibitem[Tang et~al.(2022{\natexlab{a}})Tang, Zhu, Bai, Zhao, Wang, and
  Ouyang]{tang2022unifying}
Shixiang Tang, Feng Zhu, Lei Bai, Rui Zhao, Chenyu Wang, and Wanli Ouyang.
\newblock Unifying visual contrastive learning for object recognition from a
  graph perspective.
\newblock In \emph{Computer Vision--ECCV 2022: 17th European Conference, Tel
  Aviv, Israel, October 23--27, 2022, Proceedings, Part XXVI}, pages 649--667.
  Springer, 2022{\natexlab{a}}.

\bibitem[Caron et~al.(2018)Caron, Bojanowski, Joulin, and Douze]{caron2018deep}
Mathilde Caron, Piotr Bojanowski, Armand Joulin, and Matthijs Douze.
\newblock Deep clustering for unsupervised learning of visual features.
\newblock In \emph{Proceedings of the European conference on computer vision
  (ECCV)}, pages 132--149, 2018.

\bibitem[Caron et~al.(2020)Caron, Misra, Mairal, Goyal, Bojanowski, and
  Joulin]{caron2020unsupervised}
Mathilde Caron, Ishan Misra, Julien Mairal, Priya Goyal, Piotr Bojanowski, and
  Armand Joulin.
\newblock Unsupervised learning of visual features by contrasting cluster
  assignments.
\newblock \emph{Advances in neural information processing systems},
  33:\penalty0 9912--9924, 2020.

\bibitem[Zhan et~al.(2020)Zhan, Xie, Liu, Ong, and Loy]{zhan2020online}
Xiaohang Zhan, Jiahao Xie, Ziwei Liu, Yew-Soon Ong, and Chen~Change Loy.
\newblock Online deep clustering for unsupervised representation learning.
\newblock In \emph{Proceedings of the IEEE/CVF conference on computer vision
  and pattern recognition}, pages 6688--6697, 2020.

\bibitem[Wu et~al.(2018)Wu, Xiong, Yu, and Lin]{wu2018unsupervised}
Zhirong Wu, Yuanjun Xiong, Stella~X Yu, and Dahua Lin.
\newblock Unsupervised feature learning via non-parametric instance
  discrimination.
\newblock In \emph{Proceedings of the IEEE conference on computer vision and
  pattern recognition}, pages 3733--3742, 2018.

\bibitem[Chen et~al.(2021{\natexlab{a}})Chen, Xie, and He]{chen2021empirical}
Xinlei Chen, Saining Xie, and Kaiming He.
\newblock An empirical study of training self-supervised vision transformers.
\newblock In \emph{Proceedings of the IEEE/CVF International Conference on
  Computer Vision}, pages 9640--9649, 2021{\natexlab{a}}.

\bibitem[Garrido et~al.(2022)Garrido, Chen, Bardes, Najman, and
  Lecun]{garrido2022duality}
Quentin Garrido, Yubei Chen, Adrien Bardes, Laurent Najman, and Yann Lecun.
\newblock On the duality between contrastive and non-contrastive
  self-supervised learning.
\newblock \emph{arXiv preprint arXiv:2206.02574}, 2022.

\bibitem[Grill et~al.(2020)Grill, Strub, Altch{\'e}, Tallec, Richemond,
  Buchatskaya, Doersch, Avila~Pires, Guo, Gheshlaghi~Azar,
  et~al.]{grill2020bootstrap}
Jean-Bastien Grill, Florian Strub, Florent Altch{\'e}, Corentin Tallec, Pierre
  Richemond, Elena Buchatskaya, Carl Doersch, Bernardo Avila~Pires, Zhaohan
  Guo, Mohammad Gheshlaghi~Azar, et~al.
\newblock Bootstrap your own latent-a new approach to self-supervised learning.
\newblock \emph{Advances in neural information processing systems},
  33:\penalty0 21271--21284, 2020.

\bibitem[Lee et~al.(2021)Lee, Arnab, Guadarrama, Canny, and
  Fischer]{lee2021compressive}
Kuang-Huei Lee, Anurag Arnab, Sergio Guadarrama, John Canny, and Ian Fischer.
\newblock Compressive visual representations.
\newblock \emph{Advances in Neural Information Processing Systems},
  34:\penalty0 19538--19552, 2021.

\bibitem[Xiao et~al.(2021)Xiao, Reed, Wang, Keutzer, and
  Darrell]{xiao2021region}
Tete Xiao, Colorado~J Reed, Xiaolong Wang, Kurt Keutzer, and Trevor Darrell.
\newblock Region similarity representation learning.
\newblock In \emph{Proceedings of the IEEE/CVF International Conference on
  Computer Vision}, pages 10539--10548, 2021.

\bibitem[Wang et~al.(2021)Wang, Zhang, Shen, Kong, and Li]{wang2021dense}
Xinlong Wang, Rufeng Zhang, Chunhua Shen, Tao Kong, and Lei Li.
\newblock Dense contrastive learning for self-supervised visual pre-training.
\newblock In \emph{Proceedings of the IEEE/CVF Conference on Computer Vision
  and Pattern Recognition}, pages 3024--3033, 2021.

\bibitem[Zhuang et~al.(2019)Zhuang, Li, Hu, Ma, Yang, and
  Zheng]{zhuang2019self}
Xinrui Zhuang, Yuexiang Li, Yifan Hu, Kai Ma, Yujiu Yang, and Yefeng Zheng.
\newblock Self-supervised feature learning for 3d medical images by playing a
  rubik’s cube.
\newblock In \emph{International Conference on Medical Image Computing and
  Computer-Assisted Intervention}, pages 420--428. Springer, 2019.

\bibitem[Chen et~al.(2019{\natexlab{a}})Chen, Bentley, Mori, Misawa, Fujiwara,
  and Rueckert]{chen2019self}
Liang Chen, Paul Bentley, Kensaku Mori, Kazunari Misawa, Michitaka Fujiwara,
  and Daniel Rueckert.
\newblock Self-supervised learning for medical image analysis using image
  context restoration.
\newblock \emph{Medical image analysis}, 58:\penalty0 101539,
  2019{\natexlab{a}}.

\bibitem[Zhou et~al.(2021)Zhou, Sodha, Pang, Gotway, and Liang]{zhou2021models}
Zongwei Zhou, Vatsal Sodha, Jiaxuan Pang, Michael~B Gotway, and Jianming Liang.
\newblock Models genesis.
\newblock \emph{Medical image analysis}, 67:\penalty0 101840, 2021.

\bibitem[Haghighi et~al.(2021)Haghighi, Taher, Zhou, Gotway, and
  Liang]{haghighi2021transferable}
Fatemeh Haghighi, Mohammad Reza~Hosseinzadeh Taher, Zongwei Zhou, Michael~B
  Gotway, and Jianming Liang.
\newblock Transferable visual words: Exploiting the semantics of anatomical
  patterns for self-supervised learning.
\newblock \emph{IEEE transactions on medical imaging}, 40\penalty0
  (10):\penalty0 2857--2868, 2021.

\bibitem[Bommasani et~al.(2021)Bommasani, Hudson, Adeli, Altman, Arora, von
  Arx, Bernstein, Bohg, Bosselut, Brunskill,
  et~al.]{bommasani2021opportunities}
Rishi Bommasani, Drew~A Hudson, Ehsan Adeli, Russ Altman, Simran Arora, Sydney
  von Arx, Michael~S Bernstein, Jeannette Bohg, Antoine Bosselut, Emma
  Brunskill, et~al.
\newblock On the opportunities and risks of foundation models.
\newblock \emph{arXiv preprint arXiv:2108.07258}, 2021.

\bibitem[Wang et~al.(2022)Wang, Chen, Wu, Luo, Zhou, Zhao, Xie, Liu, Jiang, and
  Yuan]{wang2022omnivl}
Junke Wang, Dongdong Chen, Zuxuan Wu, Chong Luo, Luowei Zhou, Yucheng Zhao,
  Yujia Xie, Ce~Liu, Yu-Gang Jiang, and Lu~Yuan.
\newblock Omnivl: One foundation model for image-language and video-language
  tasks.
\newblock \emph{arXiv preprint arXiv:2209.07526}, 2022.

\bibitem[Ma et~al.(2021)Ma, Jiang, Fan, Jiang, and Yan]{ma2021image}
Jiayi Ma, Xingyu Jiang, Aoxiang Fan, Junjun Jiang, and Junchi Yan.
\newblock Image matching from handcrafted to deep features: A survey.
\newblock \emph{International Journal of Computer Vision}, 129:\penalty0
  23--79, 2021.

\bibitem[Cao and Bernard(2023)]{cao2023self}
Dongliang Cao and Florian Bernard.
\newblock Self-supervised learning for multimodal non-rigid 3d shape matching.
\newblock In \emph{Proceedings of the IEEE/CVF Conference on Computer Vision
  and Pattern Recognition}, pages 17735--17744, 2023.

\bibitem[Yilmaz et~al.(2006)Yilmaz, Javed, and Shah]{yilmaz2006object}
Alper Yilmaz, Omar Javed, and Mubarak Shah.
\newblock Object tracking: A survey.
\newblock \emph{Acm computing surveys (CSUR)}, 38\penalty0 (4):\penalty0
  13--es, 2006.

\bibitem[Hyun et~al.(2023)Hyun, Kang, Wee, and Yeung]{hyun2023detection}
Jeongseok Hyun, Myunggu Kang, Dongyoon Wee, and Dit-Yan Yeung.
\newblock Detection recovery in online multi-object tracking with sparse graph
  tracker.
\newblock In \emph{Proceedings of the IEEE/CVF Winter Conference on
  Applications of Computer Vision}, pages 4850--4859, 2023.

\bibitem[Heimann and Meinzer(2009)]{heimann2009statistical}
Tobias Heimann and Hans-Peter Meinzer.
\newblock Statistical shape models for 3d medical image segmentation: a review.
\newblock \emph{Medical image analysis}, 13\penalty0 (4):\penalty0 543--563,
  2009.

\bibitem[Roetzer et~al.(2023)Roetzer, L{\"a}hner, and
  Bernard]{roetzer2023conjugate}
Paul Roetzer, Zorah L{\"a}hner, and Florian Bernard.
\newblock Conjugate product graphs for globally optimal 2d-3d shape matching.
\newblock In \emph{Proceedings of the IEEE/CVF Conference on Computer Vision
  and Pattern Recognition}, pages 21866--21875, 2023.

\bibitem[Bian et~al.(2022)Bian, Hui, Qian, and Xie]{bian2022unsupervised}
Yikai Bian, Le~Hui, Jianjun Qian, and Jin Xie.
\newblock Unsupervised domain adaptation for point cloud semantic segmentation
  via graph matching.
\newblock In \emph{2022 IEEE/RSJ International Conference on Intelligent Robots
  and Systems (IROS)}, pages 9899--9904. IEEE, 2022.

\bibitem[Wu and Ye(2023)]{wu2023unsupervised}
Zesen Wu and Mang Ye.
\newblock Unsupervised visible-infrared person re-identification via
  progressive graph matching and alternate learning.
\newblock In \emph{Proceedings of the IEEE/CVF Conference on Computer Vision
  and Pattern Recognition}, pages 9548--9558, 2023.

\bibitem[Peng et~al.(2022)Peng, Wang, Xu, Zhu, and Li]{peng2022gate}
Liang Peng, Nan Wang, Jie Xu, Xiaofeng Zhu, and Xiaoxiao Li.
\newblock Gate: graph cca for temporal self-supervised learning for
  label-efficient fmri analysis.
\newblock \emph{IEEE Transactions on Medical Imaging}, 42\penalty0
  (2):\penalty0 391--402, 2022.

\bibitem[Liu et~al.(2022)Liu, Zhang, Yang, and Yan]{liu2022self}
Chang Liu, Shaofeng Zhang, Xiaokang Yang, and Junchi Yan.
\newblock Self-supervised learning of visual graph matching.
\newblock In \emph{European Conference on Computer Vision}, pages 370--388.
  Springer, 2022.

\bibitem[Zass and Shashua(2008)]{zass2008probabilistic}
Ron Zass and Amnon Shashua.
\newblock Probabilistic graph and hypergraph matching.
\newblock In \emph{2008 IEEE Conference on Computer Vision and Pattern
  Recognition}, pages 1--8. IEEE, 2008.

\bibitem[Zhou and De~la Torre(2015)]{zhou2015factorized}
Feng Zhou and Fernando De~la Torre.
\newblock Factorized graph matching.
\newblock \emph{IEEE transactions on pattern analysis and machine
  intelligence}, 38\penalty0 (9):\penalty0 1774--1789, 2015.

\bibitem[Pogan{\v{c}}i{\'c} et~al.(2020)Pogan{\v{c}}i{\'c}, Paulus, Musil,
  Martius, and Rolinek]{poganvcic2020differentiation}
Marin~Vlastelica Pogan{\v{c}}i{\'c}, Anselm Paulus, Vit Musil, Georg Martius,
  and Michal Rolinek.
\newblock Differentiation of blackbox combinatorial solvers.
\newblock In \emph{International Conference on Learning Representations}, 2020.

\bibitem[Rol{\'\i}nek et~al.(2020{\natexlab{a}})Rol{\'\i}nek, Swoboda, Zietlow,
  Paulus, Musil, and Martius]{rolinek2020deep}
Michal Rol{\'\i}nek, Paul Swoboda, Dominik Zietlow, Anselm Paulus, V{\'\i}t
  Musil, and Georg Martius.
\newblock Deep graph matching via blackbox differentiation of combinatorial
  solvers.
\newblock In \emph{Computer Vision--ECCV 2020: 16th European Conference,
  Glasgow, UK, August 23--28, 2020, Proceedings, Part XXVIII 16}, pages
  407--424. Springer, 2020{\natexlab{a}}.

\bibitem[Niepert et~al.(2021)Niepert, Minervini, and Franceschi]{niepert21imle}
Mathias Niepert, Pasquale Minervini, and Luca Franceschi.
\newblock Implicit {MLE:} backpropagating through discrete exponential family
  distributions.
\newblock In \emph{NeurIPS}, Proceedings of Machine Learning Research. {PMLR},
  2021.

\bibitem[Kipf and Welling(2017)]{gcn}
Thomas~N. Kipf and Max Welling.
\newblock Semi-supervised classification with graph convolutional networks.
\newblock In \emph{5th International Conference on Learning Representations,
  {ICLR} 2017, Toulon, France, April 24-26, 2017, Conference Track
  Proceedings}. OpenReview.net, 2017.
\newblock URL \url{https://openreview.net/forum?id=SJU4ayYgl}.

\bibitem[Wu et~al.(2019)Wu, Souza, Zhang, Fifty, Yu, and
  Weinberger]{wu2019simplifying}
Felix Wu, Amauri Souza, Tianyi Zhang, Christopher Fifty, Tao Yu, and Kilian
  Weinberger.
\newblock Simplifying graph convolutional networks.
\newblock In \emph{International conference on machine learning}, pages
  6861--6871. PMLR, 2019.

\bibitem[Yang et~al.(2022)Yang, Zhang, Cui, Su, Luo, and Wei]{yang2022inscon}
Junwei Yang, Ke~Zhang, Zhaolin Cui, Jinming Su, Junfeng Luo, and Xiaolin Wei.
\newblock Inscon: instance consistency feature representation via
  self-supervised learning.
\newblock \emph{arXiv preprint arXiv:2203.07688}, 2022.

\bibitem[Xie et~al.(2021)Xie, Lin, Zhang, Cao, Lin, and Hu]{xie2021propagate}
Zhenda Xie, Yutong Lin, Zheng Zhang, Yue Cao, Stephen Lin, and Han Hu.
\newblock Propagate yourself: Exploring pixel-level consistency for
  unsupervised visual representation learning.
\newblock In \emph{Proceedings of the IEEE/CVF Conference on Computer Vision
  and Pattern Recognition}, pages 16684--16693, 2021.

\bibitem[Burkard et~al.(1998)Burkard, Cela, Pardalos, and
  Pitsoulis]{burkard1998quadratic}
Rainer~E Burkard, Eranda Cela, Panos~M Pardalos, and Leonidas~S Pitsoulis.
\newblock \emph{The quadratic assignment problem}.
\newblock Springer, 1998.

\bibitem[Swoboda et~al.(2017)Swoboda, Rother, Abu~Alhaija, Kainmuller, and
  Savchynskyy]{swoboda2017study}
Paul Swoboda, Carsten Rother, Hassan Abu~Alhaija, Dagmar Kainmuller, and Bogdan
  Savchynskyy.
\newblock A study of lagrangean decompositions and dual ascent solvers for
  graph matching.
\newblock In \emph{Proceedings of the IEEE conference on computer vision and
  pattern recognition}, pages 1607--1616, 2017.

\bibitem[Rol{\'\i}nek et~al.(2020{\natexlab{b}})Rol{\'\i}nek, Musil, Paulus,
  Vlastelica, Michaelis, and Martius]{rolinek2020optimizing}
Michal Rol{\'\i}nek, V{\'\i}t Musil, Anselm Paulus, Marin Vlastelica, Claudio
  Michaelis, and Georg Martius.
\newblock Optimizing rank-based metrics with blackbox differentiation.
\newblock In \emph{Proceedings of the IEEE/CVF Conference on Computer Vision
  and Pattern Recognition}, pages 7620--7630, 2020{\natexlab{b}}.

\bibitem[Minervini et~al.(2022)Minervini, Franceschi, and
  Niepert]{minervini2022adaptive}
Pasquale Minervini, Luca Franceschi, and Mathias Niepert.
\newblock Adaptive perturbation-based gradient estimation for discrete latent
  variable models.
\newblock \emph{arXiv preprint arXiv:2209.04862}, 2022.

\bibitem[Papandreou and Yuille(2011)]{papandreou2011perturb}
George Papandreou and Alan~L Yuille.
\newblock Perturb-and-map random fields: Using discrete optimization to learn
  and sample from energy models.
\newblock In \emph{2011 International Conference on Computer Vision}, pages
  193--200. IEEE, 2011.

\bibitem[He et~al.(2016)He, Zhang, Ren, and Sun]{he2016deep}
Kaiming He, Xiangyu Zhang, Shaoqing Ren, and Jian Sun.
\newblock Deep residual learning for image recognition.
\newblock In \emph{Proceedings of the IEEE conference on computer vision and
  pattern recognition}, pages 770--778, 2016.

\bibitem[Dosovitskiy et~al.(2020)Dosovitskiy, Beyer, Kolesnikov, Weissenborn,
  Zhai, Unterthiner, Dehghani, Minderer, Heigold, Gelly,
  et~al.]{dosovitskiy2020image}
Alexey Dosovitskiy, Lucas Beyer, Alexander Kolesnikov, Dirk Weissenborn,
  Xiaohua Zhai, Thomas Unterthiner, Mostafa Dehghani, Matthias Minderer, Georg
  Heigold, Sylvain Gelly, et~al.
\newblock An image is worth 16x16 words: Transformers for image recognition at
  scale.
\newblock \emph{arXiv preprint arXiv:2010.11929}, 2020.

\bibitem[Deng et~al.(2009)Deng, Dong, Socher, Li, Li, and
  Fei-Fei]{deng2009imagenet}
Jia Deng, Wei Dong, Richard Socher, Li-Jia Li, Kai Li, and Li~Fei-Fei.
\newblock Imagenet: A large-scale hierarchical image database.
\newblock In \emph{2009 IEEE conference on computer vision and pattern
  recognition}, pages 248--255. Ieee, 2009.

\bibitem[Kingma and Ba(2014)]{kingma2014adam}
Diederik~P Kingma and Jimmy Ba.
\newblock Adam: A method for stochastic optimization.
\newblock \emph{arXiv preprint arXiv:1412.6980}, 2014.

\bibitem[Bakas et~al.(2018)Bakas, Reyes, Jakab, Bauer, Rempfler, Crimi,
  Shinohara, Berger, Ha, Rozycki, et~al.]{bakas2018identifying}
Spyridon Bakas, Mauricio Reyes, Andras Jakab, Stefan Bauer, Markus Rempfler,
  Alessandro Crimi, Russell~Takeshi Shinohara, Christoph Berger, Sung~Min Ha,
  Martin Rozycki, et~al.
\newblock Identifying the best machine learning algorithms for brain tumor
  segmentation, progression assessment, and overall survival prediction in the
  brats challenge.
\newblock \emph{arXiv preprint arXiv:1811.02629}, 2018.

\bibitem[Zhuang and Shen(2016)]{zhuang2016multi}
Xiahai Zhuang and Juan Shen.
\newblock Multi-scale patch and multi-modality atlases for whole heart
  segmentation of mri.
\newblock \emph{Medical image analysis}, 31:\penalty0 77--87, 2016.

\bibitem[Codella et~al.(2019)Codella, Rotemberg, Tschandl, Celebi, Dusza,
  Gutman, Helba, Kalloo, Liopyris, Marchetti, et~al.]{codella2019skin}
Noel Codella, Veronica Rotemberg, Philipp Tschandl, M~Emre Celebi, Stephen
  Dusza, David Gutman, Brian Helba, Aadi Kalloo, Konstantinos Liopyris, Michael
  Marchetti, et~al.
\newblock Skin lesion analysis toward melanoma detection 2018: A challenge
  hosted by the international skin imaging collaboration (isic).
\newblock \emph{arXiv preprint arXiv:1902.03368}, 2019.

\bibitem[Shiraishi et~al.(2000)Shiraishi, Katsuragawa, Ikezoe, Matsumoto,
  Kobayashi, Komatsu, Matsui, Fujita, Kodera, and
  Doi]{shiraishi2000development}
Junji Shiraishi, Shigehiko Katsuragawa, Junpei Ikezoe, Tsuneo Matsumoto,
  Takeshi Kobayashi, Ken-ichi Komatsu, Mitate Matsui, Hiroshi Fujita, Yoshie
  Kodera, and Kunio Doi.
\newblock Development of a digital image database for chest radiographs with
  and without a lung nodule: receiver operating characteristic analysis of
  radiologists' detection of pulmonary nodules.
\newblock \emph{American Journal of Roentgenology}, 174\penalty0 (1):\penalty0
  71--74, 2000.

\bibitem[Pogorelov et~al.(2017)Pogorelov, Randel, Griwodz, Eskeland, de~Lange,
  Johansen, Spampinato, Dang-Nguyen, Lux, Schmidt, et~al.]{pogorelov2017kvasir}
Konstantin Pogorelov, Kristin~Ranheim Randel, Carsten Griwodz, Sigrun~Losada
  Eskeland, Thomas de~Lange, Dag Johansen, Concetto Spampinato, Duc-Tien
  Dang-Nguyen, Mathias Lux, Peter~Thelin Schmidt, et~al.
\newblock Kvasir: A multi-class image dataset for computer aided
  gastrointestinal disease detection.
\newblock In \emph{Proceedings of the 8th ACM on Multimedia Systems
  Conference}, pages 164--169, 2017.

\bibitem[Staal et~al.(2004)Staal, Abr{\`a}moff, Niemeijer, Viergever, and
  Van~Ginneken]{staal2004ridge}
Joes Staal, Michael~D Abr{\`a}moff, Meindert Niemeijer, Max~A Viergever, and
  Bram Van~Ginneken.
\newblock Ridge-based vessel segmentation in color images of the retina.
\newblock \emph{IEEE transactions on medical imaging}, 23\penalty0
  (4):\penalty0 501--509, 2004.

\bibitem[Al-Dhabyani et~al.(2020)Al-Dhabyani, Gomaa, Khaled, and
  Fahmy]{al2020dataset}
Walid Al-Dhabyani, Mohammed Gomaa, Hussien Khaled, and Aly Fahmy.
\newblock Dataset of breast ultrasound images.
\newblock \emph{Data in brief}, 28:\penalty0 104863, 2020.

\bibitem[Zhou et~al.(2020)Zhou, Wang, Huang, Cui, and Shao]{zhou2020benchmark}
Yi~Zhou, Boyang Wang, Lei Huang, Shanshan Cui, and Ling Shao.
\newblock A benchmark for studying diabetic retinopathy: segmentation, grading,
  and transferability.
\newblock \emph{IEEE Transactions on Medical Imaging}, 40\penalty0
  (3):\penalty0 818--828, 2020.

\bibitem[Liu et~al.(2020)Liu, Dou, Yu, and Heng]{liu2020ms}
Quande Liu, Qi~Dou, Lequan Yu, and Pheng~Ann Heng.
\newblock Ms-net: Multi-site network for improving prostate segmentation with
  heterogeneous mri data.
\newblock \emph{IEEE Transactions on Medical Imaging}, 2020.

\bibitem[Nguyen et~al.(2022{\natexlab{a}})Nguyen, Lam, Le, Pham, Tran, Nguyen,
  Le, Pham, Tong, Dinh, et~al.]{nguyen2022vindr}
Ha~Q Nguyen, Khanh Lam, Linh~T Le, Hieu~H Pham, Dat~Q Tran, Dung~B Nguyen,
  Dung~D Le, Chi~M Pham, Hang~TT Tong, Diep~H Dinh, et~al.
\newblock Vindr-cxr: An open dataset of chest x-rays with radiologist’s
  annotations.
\newblock \emph{Scientific Data}, 9\penalty0 (1):\penalty0 429,
  2022{\natexlab{a}}.

\bibitem[Kamran et~al.(2021)Kamran, Hossain, Tavakkoli, Zuckerbrod, Sanders,
  and Baker]{kamran2021rv}
Sharif~Amit Kamran, Khondker~Fariha Hossain, Alireza Tavakkoli, Stewart~Lee
  Zuckerbrod, Kenton~M Sanders, and Salah~A Baker.
\newblock Rv-gan: Segmenting retinal vascular structure in fundus photographs
  using a novel multi-scale generative adversarial network.
\newblock In \emph{Medical Image Computing and Computer Assisted
  Intervention--MICCAI 2021: 24th International Conference, Strasbourg, France,
  September 27--October 1, 2021, Proceedings, Part VIII 24}, pages 34--44.
  Springer, 2021.

\bibitem[Myronenko(2019)]{myronenko20193d}
Andriy Myronenko.
\newblock 3d mri brain tumor segmentation using autoencoder regularization.
\newblock In \emph{Brainlesion: Glioma, Multiple Sclerosis, Stroke and
  Traumatic Brain Injuries: 4th International Workshop, BrainLes 2018, Held in
  Conjunction with MICCAI 2018, Granada, Spain, September 16, 2018, Revised
  Selected Papers, Part II 4}, pages 311--320. Springer, 2019.

\bibitem[Chen et~al.(2021{\natexlab{b}})Chen, Lu, Yu, Luo, Adeli, Wang, Lu,
  Yuille, and Zhou]{chen2021transunet}
Jieneng Chen, Yongyi Lu, Qihang Yu, Xiangde Luo, Ehsan Adeli, Yan Wang, Le~Lu,
  Alan~L Yuille, and Yuyin Zhou.
\newblock Transunet: Transformers make strong encoders for medical image
  segmentation.
\newblock \emph{arXiv preprint arXiv:2102.04306}, 2021{\natexlab{b}}.

\bibitem[Oktay et~al.(2018)Oktay, Schlemper, Folgoc, Lee, Heinrich, Misawa,
  Mori, McDonagh, Hammerla, Kainz, et~al.]{oktay2018attention}
Ozan Oktay, Jo~Schlemper, Loic~Le Folgoc, Matthew Lee, Mattias Heinrich,
  Kazunari Misawa, Kensaku Mori, Steven McDonagh, Nils~Y Hammerla, Bernhard
  Kainz, et~al.
\newblock Attention u-net: Learning where to look for the pancreas.
\newblock \emph{arXiv preprint arXiv:1804.03999}, 2018.

\bibitem[Zhou et~al.(2018)Zhou, Siddiquee, Tajbakhsh, and
  Liang]{zhou1807nested}
Z~Zhou, MMR Siddiquee, N~Tajbakhsh, and J~UNet+ Liang.
\newblock A nested u-net architecture for medical image segmentation (2018).
\newblock \emph{arXiv preprint arXiv:1807.10165}, 2018.

\bibitem[Caron et~al.(2021)Caron, Touvron, Misra, J{\'e}gou, Mairal,
  Bojanowski, and Joulin]{caron2021emerging}
Mathilde Caron, Hugo Touvron, Ishan Misra, Herv{\'e} J{\'e}gou, Julien Mairal,
  Piotr Bojanowski, and Armand Joulin.
\newblock Emerging properties in self-supervised vision transformers.
\newblock In \emph{Proceedings of the IEEE/CVF international conference on
  computer vision}, pages 9650--9660, 2021.

\bibitem[Hatamizadeh et~al.(2022)Hatamizadeh, Tang, Nath, Yang, Myronenko,
  Landman, Roth, and Xu]{hatamizadeh2022unetr}
Ali Hatamizadeh, Yucheng Tang, Vishwesh Nath, Dong Yang, Andriy Myronenko,
  Bennett Landman, Holger~R Roth, and Daguang Xu.
\newblock Unetr: Transformers for 3d medical image segmentation.
\newblock In \emph{Proceedings of the IEEE/CVF Winter Conference on
  Applications of Computer Vision}, pages 574--584, 2022.

\bibitem[Carreira and Zisserman(2017)]{carreira2017quo}
Joao Carreira and Andrew Zisserman.
\newblock Quo vadis, action recognition? a new model and the kinetics dataset.
\newblock In \emph{proceedings of the IEEE Conference on Computer Vision and
  Pattern Recognition}, pages 6299--6308, 2017.

\bibitem[Gibson et~al.(2018)Gibson, Li, Sudre, Fidon, Shakir, Wang,
  Eaton-Rosen, Gray, Doel, Hu, et~al.]{gibson2018niftynet}
Eli Gibson, Wenqi Li, Carole Sudre, Lucas Fidon, Dzhoshkun~I Shakir, Guotai
  Wang, Zach Eaton-Rosen, Robert Gray, Tom Doel, Yipeng Hu, et~al.
\newblock {NiftyNet}: a deep-learning platform for medical imaging.
\newblock \emph{Computer methods and programs in biomedicine}, 158:\penalty0
  113--122, 2018.

\bibitem[Chen et~al.(2019{\natexlab{b}})Chen, Ma, and Zheng]{chen2019med3d}
Sihong Chen, Kai Ma, and Yefeng Zheng.
\newblock Med3d: Transfer learning for 3d medical image analysis.
\newblock \emph{arXiv:1904.00625}, 2019{\natexlab{b}}.

\bibitem[Zhang et~al.(2020)Zhang, Zhang, Zhang, and Wang]{Zhang2020UniversalMF}
Xiaoman Zhang, Ya~Zhang, Xiaoyun Zhang, and Yanfeng Wang.
\newblock Universal model for 3d medical image analysis.
\newblock \emph{ArXiv}, abs/2010.06107, 2020.

\bibitem[Tang et~al.(2022{\natexlab{b}})Tang, Yang, Li, Roth, Landman, Xu,
  Nath, and Hatamizadeh]{tang2022self}
Yucheng Tang, Dong Yang, Wenqi Li, Holger~R Roth, Bennett Landman, Daguang Xu,
  Vishwesh Nath, and Ali Hatamizadeh.
\newblock Self-supervised pre-training of swin transformers for 3d medical
  image analysis.
\newblock In \emph{Proceedings of the IEEE/CVF Conference on Computer Vision
  and Pattern Recognition}, pages 20730--20740, 2022{\natexlab{b}}.

\bibitem[Nguyen et~al.(2022{\natexlab{b}})Nguyen, Nguyen, Truong, Cao, Nguyen,
  Ho, Swoboda, Albarqouni, Xie, and Sonntag]{nguyen2022joint}
Duy~MH Nguyen, Hoang Nguyen, Mai~TN Truong, Tri Cao, Binh~T Nguyen, Nhat Ho,
  Paul Swoboda, Shadi Albarqouni, Pengtao Xie, and Daniel Sonntag.
\newblock Joint self-supervised image-volume representation learning with
  intra-inter contrastive clustering.
\newblock \emph{arXiv preprint arXiv:2212.01893}, 2022{\natexlab{b}}.

\bibitem[Lin et~al.(2018)Lin, Guo, Wang, Wu, Chen, Wang, Chen, and
  Wu]{lin2018framework}
Zhiwen Lin, Ruoqian Guo, Yanjie Wang, Bian Wu, Tingting Chen, Wenzhe Wang,
  Danny~Z Chen, and Jian Wu.
\newblock A framework for identifying diabetic retinopathy based on anti-noise
  detection and attention-based fusion.
\newblock In \emph{Medical Image Computing and Computer Assisted
  Intervention--MICCAI 2018: 21st International Conference, Granada, Spain,
  September 16-20, 2018, Proceedings, Part II 11}, pages 74--82. Springer,
  2018.

\bibitem[Wu et~al.(2021)Wu, Gao, Mei, Xu, Fan, Zhang, and Cheng]{wu2021jcs}
Yu-Huan Wu, Shang-Hua Gao, Jie Mei, Jun Xu, Deng-Ping Fan, Rong-Guo Zhang, and
  Ming-Ming Cheng.
\newblock Jcs: An explainable covid-19 diagnosis system by joint classification
  and segmentation.
\newblock \emph{IEEE Transactions on Image Processing}, 30:\penalty0
  3113--3126, 2021.

\bibitem[Zhou et~al.(2019)Zhou, He, Huang, Liu, Zhu, Cui, and
  Shao]{zhou2019collaborative}
Yi~Zhou, Xiaodong He, Lei Huang, Li~Liu, Fan Zhu, Shanshan Cui, and Ling Shao.
\newblock Collaborative learning of semi-supervised segmentation and
  classification for medical images.
\newblock In \emph{Proceedings of the IEEE/CVF conference on computer vision
  and pattern recognition}, pages 2079--2088, 2019.

\bibitem[Tusfiqur et~al.(2022)Tusfiqur, Nguyen, Truong, Nguyen, Nguyen, Barz,
  Profitlich, Than, Le, Xie, et~al.]{tusfiqur2022drg}
Hasan~Md Tusfiqur, Duy~MH Nguyen, Mai~TN Truong, Triet~A Nguyen, Binh~T Nguyen,
  Michael Barz, Hans-Juergen Profitlich, Ngoc~TT Than, Ngan Le, Pengtao Xie,
  et~al.
\newblock Drg-net: Interactive joint learning of multi-lesion segmentation and
  classification for diabetic retinopathy grading.
\newblock \emph{arXiv preprint arXiv:2212.14615}, 2022.

\bibitem[Ronneberger et~al.(2015)Ronneberger, Fischer, and
  Brox]{ronneberger2015u}
Olaf Ronneberger, Philipp Fischer, and Thomas Brox.
\newblock U-net: Convolutional networks for biomedical image segmentation.
\newblock In \emph{Medical Image Computing and Computer-Assisted
  Intervention--MICCAI 2015: 18th International Conference, Munich, Germany,
  October 5-9, 2015, Proceedings, Part III 18}, pages 234--241. Springer, 2015.

\bibitem[Jadon(2020)]{jadon2020survey}
Shruti Jadon.
\newblock A survey of loss functions for semantic segmentation.
\newblock In \emph{2020 IEEE conference on computational intelligence in
  bioinformatics and computational biology (CIBCB)}, pages 1--7. IEEE, 2020.

\bibitem[Girshick(2015)]{girshick2015fast}
Ross Girshick.
\newblock Fast r-cnn.
\newblock In \emph{Proceedings of the IEEE international conference on computer
  vision}, pages 1440--1448, 2015.

\bibitem[Setio et~al.(2015)Setio, Jacobs, Gelderblom, and van
  Ginneken]{setio2015automatic}
Arnaud~AA Setio, Colin Jacobs, Jaap Gelderblom, and Bram van Ginneken.
\newblock Automatic detection of large pulmonary solid nodules in thoracic ct
  images.
\newblock \emph{Medical physics}, 42\penalty0 (10):\penalty0 5642--5653, 2015.

\bibitem[Bilic et~al.(2019)Bilic, Christ, Vorontsov, Chlebus, Chen, Dou, Fu,
  Han, Heng, Hesser, et~al.]{bilic2019liver}
Patrick Bilic, Patrick~Ferdinand Christ, Eugene Vorontsov, Grzegorz Chlebus,
  Hao Chen, Qi~Dou, Chi-Wing Fu, Xiao Han, Pheng-Ann Heng, J{\"u}rgen Hesser,
  et~al.
\newblock The liver tumor segmentation benchmark {(LiTS)}.
\newblock \emph{arXiv:1901.04056}, 2019.

\bibitem[Simpson et~al.(2019)Simpson, Antonelli, Bakas, Bilello, Farahani,
  Van~Ginneken, Kopp-Schneider, Landman, Litjens, Menze,
  et~al.]{simpson2019large}
Amber~L Simpson, Michela Antonelli, Spyridon Bakas, Michel Bilello, Keyvan
  Farahani, Bram Van~Ginneken, Annette Kopp-Schneider, Bennett~A Landman, Geert
  Litjens, Bjoern Menze, et~al.
\newblock A large annotated medical image dataset for the development and
  evaluation of segmentation algorithms.
\newblock \emph{arXiv preprint arXiv:1902.09063}, 2019.

\bibitem[Borgli et~al.(2020)Borgli, Thambawita, Smedsrud, Hicks, Jha, Eskeland,
  Randel, Pogorelov, Lux, Nguyen, et~al.]{borgli2020hyperkvasir}
Hanna Borgli, Vajira Thambawita, Pia~H Smedsrud, Steven Hicks, Debesh Jha,
  Sigrun~L Eskeland, Kristin~Ranheim Randel, Konstantin Pogorelov, Mathias Lux,
  Duc Tien~Dang Nguyen, et~al.
\newblock Hyperkvasir, a comprehensive multi-class image and video dataset for
  gastrointestinal endoscopy.
\newblock \emph{Scientific data}, 7\penalty0 (1):\penalty0 283, 2020.

\bibitem[Veeling et~al.(2018)Veeling, Linmans, Winkens, Cohen, and
  Welling]{Veeling2018-qh}
Bastiaan~S Veeling, Jasper Linmans, Jim Winkens, Taco Cohen, and Max Welling.
\newblock Rotation equivariant {CNNs} for digital pathology.
\newblock June 2018.

\bibitem[Bejnordi et~al.(2017)Bejnordi, Veta, Van~Diest, Van~Ginneken,
  Karssemeijer, Litjens, Van Der~Laak, Hermsen, Manson, Balkenhol,
  et~al.]{bejnordi2017diagnostic}
Babak~Ehteshami Bejnordi, Mitko Veta, Paul~Johannes Van~Diest, Bram
  Van~Ginneken, Nico Karssemeijer, Geert Litjens, Jeroen~AWM Van Der~Laak,
  Meyke Hermsen, Quirine~F Manson, Maschenka Balkenhol, et~al.
\newblock Diagnostic assessment of deep learning algorithms for detection of
  lymph node metastases in women with breast cancer.
\newblock \emph{Jama}, 318\penalty0 (22):\penalty0 2199--2210, 2017.

\bibitem[Menze et~al.(2015)Menze, Jakab, Bauer, Kalpathy-Cramer, Farahani,
  Kirby, Burren, Porz, Slotboom, Wiest, Lanczi, Gerstner, Weber, Arbel, Avants,
  Ayache, Buendia, Collins, Cordier, Corso, Criminisi, Das, Delingette,
  Demiralp, Durst, Dojat, Doyle, Festa, Forbes, Geremia, Glocker, Golland, Guo,
  Hamamci, Iftekharuddin, Jena, John, Konukoglu, Lashkari, Mariz, Meier,
  Pereira, Precup, Price, Raviv, Reza, Ryan, Sarikaya, Schwartz, Shin, Shotton,
  Silva, Sousa, Subbanna, Szekely, Taylor, Thomas, Tustison, Unal, Vasseur,
  Wintermark, Ye, Zhao, Zhao, Zikic, Prastawa, Reyes, and Van~Leemput]{6975210}
Bjoern~H. Menze, Andras Jakab, Stefan Bauer, Jayashree Kalpathy-Cramer, Keyvan
  Farahani, Justin Kirby, Yuliya Burren, Nicole Porz, Johannes Slotboom, Roland
  Wiest, Levente Lanczi, Elizabeth Gerstner, Marc-André Weber, Tal Arbel,
  Brian~B. Avants, Nicholas Ayache, Patricia Buendia, D.~Louis Collins, Nicolas
  Cordier, Jason~J. Corso, Antonio Criminisi, Tilak Das, Hervé Delingette,
  Çağatay Demiralp, Christopher~R. Durst, Michel Dojat, Senan Doyle, Joana
  Festa, Florence Forbes, Ezequiel Geremia, Ben Glocker, Polina Golland,
  Xiaotao Guo, Andac Hamamci, Khan~M. Iftekharuddin, Raj Jena, Nigel~M. John,
  Ender Konukoglu, Danial Lashkari, José~António Mariz, Raphael Meier,
  Sérgio Pereira, Doina Precup, Stephen~J. Price, Tammy~Riklin Raviv, Syed
  M.~S. Reza, Michael Ryan, Duygu Sarikaya, Lawrence Schwartz, Hoo-Chang Shin,
  Jamie Shotton, Carlos~A. Silva, Nuno Sousa, Nagesh~K. Subbanna, Gabor
  Szekely, Thomas~J. Taylor, Owen~M. Thomas, Nicholas~J. Tustison, Gozde Unal,
  Flor Vasseur, Max Wintermark, Dong~Hye Ye, Liang Zhao, Binsheng Zhao, Darko
  Zikic, Marcel Prastawa, Mauricio Reyes, and Koen Van~Leemput.
\newblock The multimodal brain tumor image segmentation benchmark (brats).
\newblock \emph{IEEE Transactions on Medical Imaging}, 34\penalty0
  (10):\penalty0 1993--2024, 2015.
\newblock \doi{10.1109/TMI.2014.2377694}.

\bibitem[Lloyd et~al.(2017)Lloyd, Sorichetta, and Tatem]{lloyd2017high}
Christopher~T Lloyd, Alessandro Sorichetta, and Andrew~J Tatem.
\newblock High resolution global gridded data for use in population studies.
\newblock \emph{Scientific data}, 4\penalty0 (1):\penalty0 1--17, 2017.

\bibitem[Grossberg et~al.(2018)Grossberg, Mohamed, Elhalawani, Bennett, Smith,
  Nolan, Williams, Chamchod, Heukelom, Kantor, et~al.]{grossberg2018imaging}
Aaron~J Grossberg, Abdallah~SR Mohamed, Hesham Elhalawani, William~C Bennett,
  Kirk~E Smith, Tracy~S Nolan, Bowman Williams, Sasikarn Chamchod, Jolien
  Heukelom, Michael~E Kantor, et~al.
\newblock Imaging and clinical data archive for head and neck squamous cell
  carcinoma patients treated with radiotherapy.
\newblock \emph{Scientific data}, 5\penalty0 (1):\penalty0 1--10, 2018.

\bibitem[Bilic et~al.(2023)Bilic, Christ, Li, Vorontsov, Ben-Cohen, Kaissis,
  Szeskin, Jacobs, Mamani, Chartrand, et~al.]{bilic2023liver}
Patrick Bilic, Patrick Christ, Hongwei~Bran Li, Eugene Vorontsov, Avi
  Ben-Cohen, Georgios Kaissis, Adi Szeskin, Colin Jacobs, Gabriel
  Efrain~Humpire Mamani, Gabriel Chartrand, et~al.
\newblock The liver tumor segmentation benchmark (lits).
\newblock \emph{Medical Image Analysis}, 84:\penalty0 102680, 2023.

\bibitem[Kwan et~al.(2018)Kwan, Su, Huang, Ghoraie, Xu, Chan, Yip, Giuliani,
  Bayley, Kim, et~al.]{kwan2018radiomic}
Jennifer Yin~Yee Kwan, Jie Su, Shao~Hui Huang, Laleh~S Ghoraie, Wei Xu, Biu
  Chan, Kenneth~W Yip, Meredith Giuliani, Andrew Bayley, John Kim, et~al.
\newblock Radiomic biomarkers to refine risk models for distant metastasis in
  hpv-related oropharyngeal carcinoma.
\newblock \emph{International Journal of Radiation Oncology* Biology* Physics},
  102\penalty0 (4):\penalty0 1107--1116, 2018.

\bibitem[Clark et~al.(2013)Clark, Vendt, Smith, Freymann, Kirby, Koppel, Moore,
  Phillips, Maffitt, Pringle, et~al.]{clark2013cancer}
Kenneth Clark, Bruce Vendt, Kirk Smith, John Freymann, Justin Kirby, Paul
  Koppel, Stephen Moore, Stanley Phillips, David Maffitt, Michael Pringle,
  et~al.
\newblock The cancer imaging archive (tcia): maintaining and operating a public
  information repository.
\newblock \emph{Journal of digital imaging}, 26:\penalty0 1045--1057, 2013.

\bibitem[Kwan et~al.()]{kwandata}
JYY Kwan et~al.
\newblock Data from radiomic biomarkers to refine risk models for distant
  metastasis in oropharyngeal carcinoma, the cancer imaging archive, 2019.

\bibitem[Leavey et~al.(2019)Leavey, Sengupta, Rakheja, Daescu, Arunachalam, and
  Mishra]{leavey2019osteosarcoma}
P~Leavey, A~Sengupta, D~Rakheja, O~Daescu, HB~Arunachalam, and R~Mishra.
\newblock Osteosarcoma data from ut southwestern/ut dallas for viable and
  necrotic tumor assessment [data set].
\newblock \emph{The Cancer Imaging Archive}, 14, 2019.

\bibitem[Yorke et~al.(2019)Yorke, McDonald, Solis, and
  Guerrero]{yorke2019pelvic}
AA~Yorke, GC~McDonald, D~Solis, and T~Guerrero.
\newblock Pelvic reference data.
\newblock \emph{Cancer Imaging Arch}, 2019.

\bibitem[Roth et~al.(2016)Roth, Farag, Turkbey, Lu, Liu, and
  Summers]{roth2016data}
Holger~R Roth, Amal Farag, E~Turkbey, Le~Lu, Jiamin Liu, and Ronald~M Summers.
\newblock Data from pancreas-ct. the cancer imaging archive.
\newblock \emph{IEEE Transactions on Image Processing}, 2016.

\bibitem[Roth et~al.(2015)Roth, Lu, Farag, Shin, Liu, Turkbey, and
  Summers]{roth2015deeporgan}
Holger~R Roth, Le~Lu, Amal Farag, Hoo-Chang Shin, Jiamin Liu, Evrim~B Turkbey,
  and Ronald~M Summers.
\newblock Deeporgan: Multi-level deep convolutional networks for automated
  pancreas segmentation.
\newblock In \emph{Medical Image Computing and Computer-Assisted
  Intervention--MICCAI 2015: 18th International Conference, Munich, Germany,
  October 5-9, 2015, Proceedings, Part I 18}, pages 556--564. Springer, 2015.

\bibitem[Litjens et~al.(2014)Litjens, Debats, Barentsz, Karssemeijer, and
  Huisman]{litjens2014computer}
Geert Litjens, Oscar Debats, Jelle Barentsz, Nico Karssemeijer, and Henkjan
  Huisman.
\newblock Computer-aided detection of prostate cancer in mri.
\newblock \emph{IEEE transactions on medical imaging}, 33\penalty0
  (5):\penalty0 1083--1092, 2014.

\bibitem[Litjens et~al.(2017)Litjens, Debats, Barentsz, Karssemeijer, and
  Huisman]{Litjens2017}
Geert Litjens, Oscar Debats, Jelle Barentsz, Nico Karssemeijer, and Henkjan
  Huisman.
\newblock Prostatex challenge data.
\newblock The Cancer Imaging Archive, 2017.
\newblock \url{https://doi.org/10.7937/K9/TCIA.2017.SMPYMRXQ}.

\bibitem[Lucchesi and Aredes(2016{\natexlab{a}})]{lucchesi2016radiology}
FR~Lucchesi and ND~Aredes.
\newblock Radiology data from the cancer genome atlas cervical squamous cell
  carcinoma and endocervical adenocarcinoma (tcga-cesc) collection. the cancer
  imaging archive.
\newblock \emph{Cancer Imaging Arch}, 2016{\natexlab{a}}.

\bibitem[Kirk et~al.(2016)Kirk, Lee, Sadow, Levine, Roche, Bonaccio, and
  Filiippini]{kirk2016radiology}
S~Kirk, Y~Lee, CA~Sadow, S~Levine, C~Roche, E~Bonaccio, and J~Filiippini.
\newblock Radiology data from the cancer genome atlas colon adenocarcinoma
  [tcga-coad] collection.
\newblock \emph{The Cancer Imaging Archive}, 2016.

\bibitem[Lucchesi and Aredes(2016{\natexlab{b}})]{Lucchesi20164}
F.~R. Lucchesi and N.~D. Aredes.
\newblock The cancer genome atlas esophageal carcinoma collection (tcga-esca)
  (version 3) [data set].
\newblock The Cancer Imaging Archive, 2016{\natexlab{b}}.

\bibitem[Linehan et~al.(2016)Linehan, Gautam, Sadow, and
  Levine]{linehan2016radiology}
MW~Linehan, R~Gautam, CA~Sadow, and S~Levine.
\newblock Radiology data from the cancer genome atlas kidney chromophobe
  [tcga-kich] collection.
\newblock \emph{The Cancer Imaging Archive}, 2016.

\bibitem[Akin et~al.(2016)Akin, Elnajjar, Heller, Jarosz, Erickson, Kirk, and
  Filippini]{akin2016radiology}
O~Akin, P~Elnajjar, M~Heller, R~Jarosz, B~Erickson, S~Kirk, and J~Filippini.
\newblock Radiology data from the cancer genome atlas kidney renal clear cell
  carcinoma [tcga-kirc] collection.
\newblock \emph{The Cancer Imaging Archive}, 1310, 2016.

\bibitem[Roche et~al.(2016)Roche, Bonaccio, and Filippini]{roche2016radiology}
C~Roche, E~Bonaccio, and J~Filippini.
\newblock Radiology data from the cancer genome atlas sarcoma [tcga-sarc]
  collection.
\newblock \emph{Cancer Imaging Arch. https://doi. org/10.7937 K}, 9, 2016.

\bibitem[Shenggan(2017)]{BCCD_Dataset}
Shenggan.
\newblock Bccd dataset, 2017.
\newblock URL \url{https://github.com/Shenggan/BCCD_Dataset}.

\bibitem[Gehlot et~al.(2020)Gehlot, Gupta, and Gupta]{gehlot2020sdct}
Shiv Gehlot, Anubha Gupta, and Ritu Gupta.
\newblock Sdct-auxnet$\theta$: Dct augmented stain deconvolutional cnn with
  auxiliary classifier for cancer diagnosis.
\newblock \emph{Medical image analysis}, 61:\penalty0 101661, 2020.

\bibitem[Gupta and Gupta(2019)]{gupta2019all}
A~Gupta and R~Gupta.
\newblock All challenge dataset of isbi 2019 [data set].
\newblock \emph{URL https://doi.org/10.7937/tcia.2019.dc64i46r}, 2019.

\bibitem[Lee et~al.(2017)Lee, Gimenez, Hoogi, Miyake, Gorovoy, and
  Rubin]{lee2017curated}
Rebecca~Sawyer Lee, Francisco Gimenez, Assaf Hoogi, Kanae~Kawai Miyake, Mia
  Gorovoy, and Daniel~L Rubin.
\newblock A curated mammography data set for use in computer-aided detection
  and diagnosis research.
\newblock \emph{Scientific data}, 4\penalty0 (1):\penalty0 1--9, 2017.

\bibitem[Sawyer~Lee et~al.(2016)Sawyer~Lee, Gimenez, Hoogi, and
  Rubin]{sawyer2016curated}
R~Sawyer~Lee, F~Gimenez, A~Hoogi, and D~Rubin.
\newblock Curated breast imaging subset of ddsm.
\newblock \emph{The Cancer Imaging Archive. DOI: https://doi. org/10.7937 K},
  9, 2016.

\bibitem[Wang et~al.(2020)Wang, Lin, and Wong]{Wang2020}
Linda Wang, Zhong~Qiu Lin, and Alexander Wong.
\newblock Covid-net: a tailored deep convolutional neural network design for
  detection of covid-19 cases from chest x-ray images.
\newblock \emph{Scientific Reports}, 10\penalty0 (1):\penalty0 19549, Nov 2020.
\newblock ISSN 2045-2322.
\newblock \doi{10.1038/s41598-020-76550-z}.
\newblock URL \url{https://doi.org/10.1038/s41598-020-76550-z}.

\bibitem[Kermany et~al.(2018)Kermany, Zhang, Goldbaum,
  et~al.]{kermany2018labeled}
Daniel Kermany, Kang Zhang, Michael Goldbaum, et~al.
\newblock Labeled optical coherence tomography (oct) and chest x-ray images for
  classification.
\newblock \emph{Mendeley data}, 2\penalty0 (2):\penalty0 651, 2018.

\bibitem[Maqbool et~al.(2020)Maqbool, Riaz, Sajid, and
  Hasan]{maqbool2020m2caiseg}
Salman Maqbool, Aqsa Riaz, Hasan Sajid, and Osman Hasan.
\newblock m2caiseg: Semantic segmentation of laparoscopic images using
  convolutional neural networks.
\newblock \emph{arXiv preprint arXiv:2008.10134}, 2020.

\bibitem[Amgad et~al.(2019{\natexlab{a}})Amgad, Elfandy, Hussein, Atteya,
  Elsebaie, Abo~Elnasr, Sakr, Salem, Ismail, Saad, Ahmed, Elsebaie, Rahman,
  Ruhban, Elgazar, Alagha, Osman, Alhusseiny, Khalaf, Younes, Abdulkarim,
  Younes, Gadallah, Elkashash, Fala, Zaki, Beezley, Chittajallu, Manthey,
  Gutman, and Cooper]{amgad2102nucls}
Mohamed Amgad, Habiba Elfandy, Hagar Hussein, Lamees~A Atteya, Mai A~T
  Elsebaie, Lamia~S Abo~Elnasr, Rokia~A Sakr, Hazem S~E Salem, Ahmed~F Ismail,
  Anas~M Saad, Joumana Ahmed, Maha A~T Elsebaie, Mustafijur Rahman, Inas~A
  Ruhban, Nada~M Elgazar, Yahya Alagha, Mohamed~H Osman, Ahmed~M Alhusseiny,
  Mariam~M Khalaf, Abo-Alela~F Younes, Ali Abdulkarim, Duaa~M Younes, Ahmed~M
  Gadallah, Ahmad~M Elkashash, Salma~Y Fala, Basma~M Zaki, Jonathan Beezley,
  Deepak~R Chittajallu, David Manthey, David~A Gutman, and Lee A~D Cooper.
\newblock {Structured crowdsourcing enables convolutional segmentation of
  histology images}.
\newblock \emph{Bioinformatics}, 35\penalty0 (18):\penalty0 3461--3467, 02
  2019{\natexlab{a}}.
\newblock ISSN 1367-4803.
\newblock \doi{10.1093/bioinformatics/btz083}.
\newblock URL \url{https://doi.org/10.1093/bioinformatics/btz083}.

\bibitem[Cuzzolin et~al.(2021)Cuzzolin, Bawa, Skarga-Bandurova, Mohamed,
  Charles, Oleari, Leporini, Landolfo, Stabile, Setti,
  et~al.]{cuzzolin2021saras}
Fabio Cuzzolin, Vivek~Singh Bawa, Inna Skarga-Bandurova, Mohamed Mohamed,
  Jackson~Ravindran Charles, Elettra Oleari, Alice Leporini, Carmela Landolfo,
  Armando Stabile, Francesco Setti, et~al.
\newblock Saras challenge on multi-domain endoscopic surgeon action detection,
  2021.

\bibitem[Bawa et~al.(2021)Bawa, Singh, KapingA, Skarga-Bandurova, Oleari,
  Leporini, Landolfo, Zhao, Xiang, Luo, Wang, Li, Wang, Zhao, Li, Stabile,
  Setti, Muradore, and Cuzzolin]{saras-mesad2}
Vivek~Singh Bawa, Gurkirt Singh, Francis KapingA, Inna Skarga-Bandurova,
  Elettra Oleari, Alice Leporini, Carmela Landolfo, Pengfei Zhao, Xi~Xiang,
  Gongning Luo, Kuanquan Wang, Liangzhi Li, Bowen Wang, Shang Zhao, Li~Li,
  Armando Stabile, Francesco Setti, Riccardo Muradore, and Fabio Cuzzolin.
\newblock The saras endoscopic surgeon action detection (esad) dataset:
  Challenges and methods, 2021.

\bibitem[Bawa et~al.(2020)Bawa, Singh, KapingA, Skarga-Bandurova, Leporini,
  Landolfo, Stabile, Setti, Muradore, Oleari, and Cuzzolin]{saras-mesad3}
Vivek~Singh Bawa, Gurkirt Singh, Francis KapingA, Inna Skarga-Bandurova, Alice
  Leporini, Carmela Landolfo, Armando Stabile, Francesco Setti, Riccardo
  Muradore, Elettra Oleari, and Fabio Cuzzolin.
\newblock Esad: Endoscopic surgeon action detection dataset, 2020.

\bibitem[sho()]{shoulder}
Shoulder x-ray classification.
\newblock \emph{Kaggle dataset}.
\newblock URL
  \url{https://www.kaggle.com/datasets/dryari5/shoulder-xray-classification}.

\bibitem[She()]{ShenzhenHospitalX-raySet}
Lung masks for shenzhen hospital chest x-ray set.
\newblock \emph{Kaggle dataset}.
\newblock URL \url{https://www.kaggle.com/datasets/yoctoman/shcxr-lung-mask}.

\bibitem[Mueller et~al.(2005)Mueller, Weiner, Thal, Petersen, Jack, Jagust,
  Trojanowski, Toga, and Beckett]{mueller2005alzheimer}
Susanne~G Mueller, Michael~W Weiner, Leon~J Thal, Ronald~C Petersen, Clifford
  Jack, William Jagust, John~Q Trojanowski, Arthur~W Toga, and Laurel Beckett.
\newblock The alzheimer's disease neuroimaging initiative.
\newblock \emph{Neuroimaging Clinics}, 15\penalty0 (4):\penalty0 869--877,
  2005.

\bibitem[Petersen et~al.(2010)Petersen, Aisen, Beckett, Donohue, Gamst, Harvey,
  Jack, Jagust, Shaw, Toga, et~al.]{petersen2010alzheimer}
Ronald~Carl Petersen, Paul~S Aisen, Laurel~A Beckett, Michael~C Donohue,
  Anthony~Collins Gamst, Danielle~J Harvey, Clifford~R Jack, William~J Jagust,
  Leslie~M Shaw, Arthur~W Toga, et~al.
\newblock Alzheimer's disease neuroimaging initiative (adni): clinical
  characterization.
\newblock \emph{Neurology}, 74\penalty0 (3):\penalty0 201--209, 2010.

\bibitem[Matek et~al.(2019{\natexlab{a}})Matek, Schwarz, Marr, and
  Spiekermann]{matek2019single}
Christian Matek, Simone Schwarz, Carsten Marr, and Karsten Spiekermann.
\newblock A single-cell morphological dataset of leukocytes from aml patients
  and non-malignant controls (aml-cytomorphology\_lmu).
\newblock \emph{The Cancer Imaging Archive (TCIA)[Internet]},
  2019{\natexlab{a}}.

\bibitem[Matek et~al.(2019{\natexlab{b}})Matek, Schwarz, Spiekermann, and
  Marr]{matek2019human}
Christian Matek, Simone Schwarz, Karsten Spiekermann, and Carsten Marr.
\newblock Human-level recognition of blast cells in acute myeloid leukaemia
  with convolutional neural networks.
\newblock \emph{Nature Machine Intelligence}, 1\penalty0 (11):\penalty0
  538--544, 2019{\natexlab{b}}.

\bibitem[kar()]{karthick2019aptos}
Aptos 2019 blindness detection.
\newblock \emph{Kaggle dataset}.
\newblock URL
  \url{https://www.kaggle.com/c/aptos2019-blindness-detection/data}.

\bibitem[Amgad et~al.(2019{\natexlab{b}})Amgad, Elfandy, Hussein, Atteya,
  Elsebaie, Abo~Elnasr, Sakr, Salem, Ismail, Saad, et~al.]{amgad2019structured}
Mohamed Amgad, Habiba Elfandy, Hagar Hussein, Lamees~A Atteya, Mai~AT Elsebaie,
  Lamia~S Abo~Elnasr, Rokia~A Sakr, Hazem~SE Salem, Ahmed~F Ismail, Anas~M
  Saad, et~al.
\newblock Structured crowdsourcing enables convolutional segmentation of
  histology images.
\newblock \emph{Bioinformatics}, 35\penalty0 (18):\penalty0 3461--3467,
  2019{\natexlab{b}}.

\bibitem[Abdi and Kasaei(2020)]{abdi2020panoramic}
Amir Abdi and S~Kasaei.
\newblock Panoramic dental x-rays with segmented mandibles.
\newblock \emph{Mendeley Data, v2}, 2020.

\bibitem[van~den Heuvel et~al.(2018)van~den Heuvel, de~Bruijn, de~Korte, and
  Ginneken]{van2018automated}
Thomas~LA van~den Heuvel, Dagmar de~Bruijn, Chris~L de~Korte, and Bram~van
  Ginneken.
\newblock Automated measurement of fetal head circumference using 2d ultrasound
  images.
\newblock \emph{PloS one}, 13\penalty0 (8):\penalty0 e0200412, 2018.

\bibitem[Hip()]{Hippseg2011}
Hippocampus segmentation in mri images.
\newblock \emph{Kaggle dataset}.
\newblock URL
  \url{https://www.kaggle.com/datasets/andrewmvd/hippocampus-segmentation-in-mri-images?select=Test}.

\bibitem[isi()]{isic}
Skin lesion images for melanoma classification.
\newblock \emph{Kaggle dataset}.
\newblock URL \url{https://www.kaggle.com/datasets/andrewmvd/isic-2019}.

\bibitem[Taha et~al.(2018)Taha, Lo, Li, and Zhao]{taha2018kid}
Ahmed Taha, Pechin Lo, Junning Li, and Tao Zhao.
\newblock Kid-net: convolution networks for kidney vessels segmentation from
  ct-volumes.
\newblock In \emph{Medical Image Computing and Computer Assisted
  Intervention--MICCAI 2018: 21st International Conference, Granada, Spain,
  September 16-20, 2018, Proceedings, Part IV 11}, pages 463--471. Springer,
  2018.

\bibitem[kva()]{kvarsir}
Kvasir v2.
\newblock \emph{Kaggle dataset}.
\newblock URL
  \url{https://www.kaggle.com/datasets/plhalvorsen/kvasir-v2-a-gastrointestinal-tract-dataset}.

\bibitem[Mal()]{Malaria}
Lhncbc malaria.
\newblock
  \url{https://lhncbc.nlm.nih.gov/LHC-downloads/downloads.html#malaria-datasets}.

\bibitem[Wei et~al.(2020)Wei, Lin, Barranco, Wendt, Liu, Yin, Huang, Gupta,
  Jang, Wang, Arganda-Carreras, Lichtman, and Pfister]{wei2020mitoem}
D.~Wei, Z.~Lin, D.~Barranco, N.~Wendt, X.~Liu, W.~Yin, X.~Huang, A.~Gupta,
  W.~Jang, X.~Wang, I.~Arganda-Carreras, J.~Lichtman, and H.~Pfister.
\newblock Mitoem dataset: Large-scale 3d mitochondria instance segmentation
  from em images.
\newblock In \emph{International Conference on Medical Image Computing and
  Computer Assisted Intervention}, 2020.

\bibitem[Matek et~al.(2021)Matek, Krappe, M{\"u}nzenmayer, Haferlach, and
  Marr]{matek2021highly}
Christian Matek, Sebastian Krappe, Christian M{\"u}nzenmayer, Torsten
  Haferlach, and Carsten Marr.
\newblock Highly accurate differentiation of bone marrow cell morphologies
  using deep neural networks on a large image data set.
\newblock \emph{Blood, The Journal of the American Society of Hematology},
  138\penalty0 (20):\penalty0 1917--1927, 2021.

\bibitem[Halabi et~al.(2019)Halabi, Prevedello, Kalpathy-Cramer, Mamonov,
  Bilbily, Cicero, Pan, Pereira, Sousa, Abdala, et~al.]{halabi2019rsna}
Safwan~S Halabi, Luciano~M Prevedello, Jayashree Kalpathy-Cramer, Artem~B
  Mamonov, Alexander Bilbily, Mark Cicero, Ian Pan, Lucas~Ara{\'u}jo Pereira,
  Rafael~Teixeira Sousa, Nitamar Abdala, et~al.
\newblock The rsna pediatric bone age machine learning challenge.
\newblock \emph{Radiology}, 290\penalty0 (2):\penalty0 498--503, 2019.

\bibitem[eye()]{eyePACS}
Kaggle dr dataset (eyepacs).
\newblock \emph{Kaggle dataset}.
\newblock URL
  \url{https://www.kaggle.com/datasets/mariaherrerot/eyepacspreprocess}.

\end{thebibliography}

\newpage
\newcommand\tab[1][5mm]{\hspace*{#1}}
\appendix
\section*{Supplementary Material}
We present below LVM-Med pseudo-code (Section \ref{sec:pseudo-code}), implementations used in downstream tasks (Section \ref{sec:pre-downs-settings}), additional ablation studies of LVM-Med (Section \ref{sec:ablation-lvm-med}), further  prompt-based segmentation results on 3D datasets, image classification benchmark (Section \ref{sec:linear-seg}), predicted masks using the user-based prompt (Section \ref{sec:visualize_images}), and finally the dataset overview (Section \ref{sec:dataset_overview}).   
\section{LVM-Med Pseudo-code}
\label{sec:pseudo-code}
First, we provide a pseudo-code for training LVM-Med in Pytorch style:
\\ \rule{\textwidth}{0.4pt}
\com{\# $\mathtt{f_{\theta}}$:\,encoder network, $\mathtt{h_{\phi}}$:\,projector network, $\mathtt{g_{\epsilon}}$:\,message passing network,}
\\
\com{\# k\_nodes:\,number of nearest neighbors, Avg:\,average pooling, } 
\\ \com{\# pos:\,position of image after transform, cos:\,cosine similarity,}
\\ 
\com{\# $\mathtt{\alpha}$:\,coefficient trades off between global and local costs, $\mathtt{L_2}$:\,L2-distance,  }
\\
\com{\# $\mathtt{\gamma}$:\,maximum pairs are kept, select\_top:\,select to keep the $\mathtt{\gamma}$ best matches.}
\\
\\ \code{for X in loader:}  \com{\# load a batch X = $\mathtt{\left[x_1, x_2,...,x_N\right]}$ with N samples}
\\
\tab \com{\# apply two transformations s and t} 
\\
\tab \code{$\mathtt{X^{s},\ Pos^{s} = s(X)}$} \ \com{\# $\mathtt{X^{k} = \left[x_1^{k},x_2^{k},...,x_{N}^{k}\right]}$,\ $\mathtt{Pos^{k} = \left[pos_{1}^{k}, pos_{2}^{k},...,pos_{N}^{k}\right],\ k \in \{s, t\}}$}\\
\tab \code{$\mathtt{X^{t},\ Pos^{t} = t(X)}$} 
\\
\\
\tab \com{\# compute feature representations}
\\
\tab \code{$\mathtt{Y^{s} = f_{\theta}(X^{s});\ Y^{t} = f_{\theta}(X^{t})}$}  \com{\# feature dimensions:NxDxRxS}
\\
\\
\tab \com{\# applying projection} 
\\
\tab \code{$\mathtt{Z^{s} = h_{\phi}(Avg(Y^{s}));\ Z^{t} = h_{\phi}(Avg(Y^{t}))}$}  \com{\# dimensions:NxF}
\\
\\
\tab \com{\# build graph structures and message passing}
\\
\tab \code{$\mathtt{G^{s}}$ = k-nearest-neighbor($\mathtt{Z^{s}}$, k\_connects)}
\\
\tab \code{$\mathtt{G^{t}}$ = k-nearest-neighbor($\mathtt{Z^{t}}$,k\_connects)}
\\
\tab \code{$\mathtt{\hat{Z}^{s} = g_{\epsilon}(G^{s}, Z^{s})}$;\ $\mathtt{\hat{Z}^{t} = g_{\epsilon}(G^{t}, Z^{t})}$}
\\
\\
\tab \com{\# compute vertex and edge affinity matrices}
\\
\tab \code{$\mathtt{c_{ia}^{v} = \alpha * \,cos(\hat{z}_{i}^{s}, \hat{z}_{a}^{t}) + (1- \alpha)*\,local\_cost(y_{i}^{s},\,y_{a}^{t},\,pos_{i}^{s},\,pos_{a}^{t})}$}\ \com{\# affinity $\mathtt{x_{i}^{s}}$ \& $\mathtt{x_{a}^{t}}$}
\\
\tab \code{$\mathtt{c^{e}_{ia, jb} = cos((\hat{z}_{i}^{s} - \hat{z}_{j}^{s}),(\hat{z}_{a}^{t} - \hat{z}_{b}^{t}))}$}\ \com{\# affinity between edges $\mathtt{v_{ij}^{s}, v_{ab}^{t}}$ }
\\
\tab \code{$\mathtt{c^v = \{c_{ij}^{v}\} \in R^{N\times N};\ c^{e} = \{c^{e}_{ia, jb}\} \in R^{|E^{s}||E^{t}|}}$} \ \com{\# $\mathtt{E^{k}}$ be a set of edges in $\mathtt{G^{k},\,k\in \{s,t\}}$}
\\
\\
\tab \com{\# perturbed costs with Gumbel noise}
\\
\tab \code{$\mathtt{\epsilon, \epsilon' \sim Gumbel(0, 1)}$}
\\
\tab \code{$\mathtt{c^v = c^v + \epsilon;\ c^e = c^e + \epsilon'}$}
\\
\\
\tab \com{\# solving graph matching and compute loss}
\\
\tab \code{$\mathtt{\hat{v} = GM(c^v, c^e)}$}
\\
\tab \code{$\mathtt{L(\hat{\bm{v}}, \bm{v}^*) = \hat{\bm{v}}.(1-\bm{v}^*) + \bm{v}^*.(1-\hat{\bm{v}})}$}
\ \com{\# compute hamming loss}
\\
\\ 
\tab \com{\# update network}
\\ \tab \code{L.backward()}\ \com{\# approximate ($\mathtt{\partial L/\partial c^v, \partial L/\partial c^e}$) by Algorithm 1.}
\\ \tab \code{Update($\mathtt{g_{\epsilon}.params}$),\ Update($\mathtt{h_{\phi}.params}$),\,Update($\mathtt{f_{\theta}.params}$)}
\\
\\
\com{\# define local\_cost}
\\
\code{def local\_cost($\mathtt{y_{i}^{s},\,y_{a}^{t},\,pos_{i}^{s},\,pos_{a}^{t}}$):}
\\ 
\\ \tab \com{\# location-based local cost}
\\ \tab \code{$\mathtt{y_{i,nn}^{s} = torch.zeros\_like(y_{i}^{s})}$}
\\
\tab \code{for r, s in R, S:}
\\
\tab \tab \code{r', s' = argmin($\mathtt{\left(L_2(pos_{i}^{s}[r, s],\ pos_{a}^{t}[r',s']\right)}$)}
\\
\tab \tab \code{$\mathtt{y_{i,nn}^{s}[r, s] = y_{a}^{t}[r', s']}$}
\\
\\ \tab \code{$\mathtt{y_{i\_fil}^{s}, y_{i,nn\_fil}^{s} = select\_top\,\left(y_{i}^{s},\,y_{i,nn}^{s}, \gamma\right)}$}
\\ 
\\ \tab \code{location\_cost = cos($\mathtt{y_{i\_fil}^{s}, y_{i,nn\_fil}^{s}}$)}
\\
\\ \tab \com{\# featured-based local cost}
\\ \tab \code{$\mathtt{y_{i,nn}^{s} = torch.zeros\_like(y_{i}^{s})}$}
\\
\tab \code{for r, s in R, S:}
\\
\tab \tab \code{r', s' = argmin($\mathtt{\left(L_2(y_{i}^{s}[r, s],\ y_{a}^{t}[r',s']\right)}$)}
\\
\tab \tab \code{$\mathtt{y_{i,nn}^{s}[r, s] = y_{a}^{t}[r', s']}$}
\\
\\ \tab \code{$\mathtt{y_{i\_fil}^{s}, y_{i,nn\_fil}^{s} = select\_top\,\left(y_{i}^{s},\,y_{i,nn}^{s}, \gamma\right)}$}
\\
\\ \tab \code{feature\_cost = cos($\mathtt{y_{i\_fil}^{s}, y_{i,nn\_fil}^{s}}$)}
\\
\\ \tab \code{\textbf{return} 0.5*(location\_cost + feature\_cost)}
\\ \rule{\textwidth}{0.5pt}
\\ \linebreak
We trained LVM-Med with graph size of $16$ nodes, each node connected to the top 5 nearest neighbors after using kNN, $\lambda$ value in Algorithm 1 is $80$, and $\alpha = 0.8$ for associating global- and local-based similarities when computing $c^{v}_{ij}$. The size of projector $h_{\phi}$ is $2048 \times 128$ for ResNet-50, and $768 \times 128$ for ViT. We configure the message passing network $g_{\theta}$ with two convolutional layers of size $128$. For the user-based prompt version, because the SAM model \cite{ma2023segment} requires an input of shape $256 \times 14 \times 14$ for the mask decoder part, we add two additional convolutional layers with a kernel size of $1$ and $3$ at the end of ViT backbone to convert from shape $768 \times 14 \times 14$ to the target shape. 

\section{Downstream task setups}
\label{sec:pre-downs-settings}

\subsection{Downstream tasks}
\paragraph{Segmentation tasks}
On 2D-based segmentation tasks, we employ U-Net architecture \cite{ronneberger2015u} and load ResNet-50 \cite{he2016deep} trained by self-supervised learning algorithms as network backbones. With foundation models, we use TransUnet \cite{chen2021transunet} and take pre-trained ViT models as the backbones. For the prompt-based segmentation, we follow the architecture of SAM \cite{kirillov2023segment} consisting of encoder, prompt, and mask decoder layers. We also fine-tune SAM where encoder and prompt networks are frozen, only learning decoder layers \cite{ma2023segment}. Our LVM-Med for prompt-based setting is similar to \cite{ma2023segment} except that we substitute SAM's encoders with our weights. We utilize Adam optimizer for all experiments and train architectures with Dice and Cross-Entropy loss \cite{jadon2020survey}. We also normalize the norm-2 of gradient values to stabilize the training step to maximize 1.
Table \ref{tab:setting_2DSegmentation} summarizes each dataset's learning rate, number of epochs, and image resolution.   

On 3D-based segmentations, we reformulate these tasks as 2D segmentation problems and make predictions on 2D slices taken from 3D volumes. Furthermore, we apply balance sampling to select equally 2D slices covering target regions and other 2D slices, not including the ground truth. Table \ref{tab:setting_3DSegmentation}
presents configurations used for 3D datasets; other settings are identical to 2D cases.
\begin{table}[h!]
\centering
\caption{Configurations for training 2D segmentation tasks}
\vspace{0.1in}
\label{tab:setting_2DSegmentation}
\resizebox{1.0\columnwidth}{!}{
\begin{tabular}{l|l|l|l|l|l}
\toprule
\textbf{} & \textbf{ISIC-2018 (Skin Lesion)} & \textbf{JSRT (Lung X-ray)} & \textbf{KvaSir (Polyp)} & \textbf{Drive (Vessel)} & \textbf{BUID (Breast Cancer)} \\ \midrule
\multirow{3}{*}{\textbf{ResNet-50.}} 
& lr = $10^{-4}$,  epochs $35$  & lr = $10^{-3}$, epochs $50$ &  lr = $10^{-3}$, epochs $35$  & lr = $10^{-3}$, epochs $50$ & lr = $10^{-4}$,  epochs $50$ \\
&  shape $512\times 512$ &  shape $224\times 224$&  shape $224\times 224$ &  shape $224\times 224$ &   shape $256\times 256$\\
& batch size $16$ & batch size $32$ & batch size $64$ & batch size $16$ &  batch size $8$\\ \midrule
\multirow{3}{*}{\textbf{Foundation Model}} 
& lr = $10^{-4}$,  epochs $100$  & lr = $10^{-3}$, epochs $200$ &  lr = $10^{-3}$,  epochs $200$ & lr = $10^{-3}$, epochs $200$ & lr = $10^{-4}$, epochs $200$  \\
&  shape $512\times 512$ &  shape $224\times 224$&  shape $224\times 224$ &  shape $224\times 224$ &  shape $256\times 256$\\
& batch size $16$ & batch size $32$ & batch size $64$ & batch size $16$ &  batch size $8$ \\ \midrule
\multirow{3}{*}{\textbf{Prompt-based Seg.}} 
& lr = $10^{-4}$,  epochs $50$  & lr = $3\times 10^{-4}$, epochs $50$ &  lr = $3\times 10^{-4}$,  epochs $20$ & lr = $3\times 10^{-4}$, epochs $100$ & lr = $10^{-4}$,  epochs $20$ \\
&   shape $1024\times 1024$ &  shape $1024\times 1024$ &  shape $1024\times 1024$ &  shape $1024\times 1024$ &  shape $1024\times 1024$\\
& batch size $16$ & batch size $16$ & batch size $16$ & batch size $16$ &  batch size $16$ \\ 
\bottomrule
\end{tabular}}
\end{table}

\begin{table}[H]
\centering
\caption{Configurations for 3D-based-segmentation tasks}
\vspace{0.1in}
\label{tab:setting_3DSegmentation}
\resizebox{0.95\columnwidth}{!}{
\begin{tabular}{l|l|l|l|l}
\toprule
\textbf{} & \textbf{BraTS} & \textbf{MMWHS-CT} & \textbf{MMWHS-MRI} & \textbf{BMC} \\ \midrule
\multirow{3}{*}{\textbf{\small{ResNet50}}} 
 &  lr = $15\times10^{-4}$, epochs $20$ & lr = $10^{-3}$, epochs $20$ &  lr = $15\times 10^{-4}$, epochs $30$  &  lr = $10^{-3}$, epochs $30$  \\
 &  shape $224 \times 224$ &  shape $224 \times 224$ &  shape $224 \times 224$ & shape $224 \times 224$  \\
 & batch size $128$ & batch size $64$ & batch size $64$ & batch size $64$ \\ 
\midrule
\multirow{3}{*}{\textbf{\small{Foundation Model}}}  
&  lr = $10^{-4}$, epochs $100$ & lr = $10^{-4}$, epochs $100$ &  lr = $10^{-4}$, epochs $100$  &  lr = $10^{-4}$, epochs $100$  \\
 & shape $224 \times 224$ & shape $224 \times 224$ &  shape $224 \times 224$ & shape $224 \times 224$  \\
 & batch size $16$ & batch size $16$ & batch size $16$ & batch size $16$  \\ \midrule
\multirow{3}{*}{\textbf{\small{Prompt-based Seg.}}}  
&  lr = $3 \times 10^{-5}$, epochs $30$ &  lr = $5\times 10^{-5}$, epochs $30$ &  lr = $3\times 10^{-5}$, epochs $30$  & lr = $3\times 10^{-4}$, epochs $50$  \\
 & shape $1024 \times 1024$ & shape $1024 \times 1024$ &  shape $1024 \times 1024$ & shape $1024 \times 1024$  \\
 & batch size $16$ & batch size $16$ & batch size $16$ & batch size $16$ \\ \bottomrule
\end{tabular}}
\end{table}
\paragraph{Image classification tasks}
We take the feature embedding outputs of each architecture and build one fully connected layer to produce desired classes for image classification tasks. We freeze the encoder layers for the linear evaluation and only train the fully connected layer. For the fully-finetuning, the whole network is trained. The Adam optimizer \cite{kingma2014adam} with cross-entropy loss function and learning rates $\{5\times 10^{-4}, 10^{-3}\}$ are used for Brain Tumor and FGADR, respectively. To benchmark LVM-Med with other state-of-the-art methods on FGADR (Figure 3 in paper), we follow the settings of DRG-Net \cite{tusfiqur2022drg} and change their encoder layers by our networks.  

\paragraph{Object detection}
We use Faster-RCNN \cite{girshick2015fast} for object detection tasks. The ResNet-50 of Faster-RCNN is replaced by pre-trained weights. In the Vin-Dr dataset, there is a total of $14$ objects for, e.g., Aortic enlargement, Atelectasis, Calcification, etc. We use  image resolutions of $512 \times 512$, Adam solver, and learning rate $10^{-4}$ in $40$ epochs. In the Kvasir dataset for polyp detection, we also resize images to a fixed size of $512 \times 512$, employ 
the Adam optimizer with learning rate $2.5\times\mathrm{10}^{-4}$ and batch size $8$. 

\section{LVM-Med ablation studies }
\label{sec:ablation-lvm-med}
\subsection{Graph sizes and $\lambda$ in backpropagation}
We provide in Figure \ref{fig:graph-numbers} and Figure \ref{fig:lambda_solver} LVM-Med performance when changing the number of nodes in graph construction steps  $G^{s}, G^{t}$ and $\lambda = 80$ used in Algorithm 1 in the backpropagation step. The results are reported on the average Dice score of five 2D segmentation tasks and the average accuracy of two linear classifications on FGADR and Brain Tumor Classification. Figure \ref{fig:graph-numbers} indicates that $16$ is the best value for both classification and segmentation. Increasing the graph's nodes tends to decrease classification performance. 

Figure \ref{fig:lambda_solver} compared different values for $\lambda \in \{70, 80, 90, 100\}$. We observe that $\lambda = \{80, 90\}$ achieve good results for linear classification tasks though $\lambda = \{90, 100\}$ decreases segmentation performance. 

\subsection{Performance on large- and small-scale}
We investigate LVM-Med performance when reducing the number of datasets in the pre-training step. Especially, we trained LVM-Med on a \textit{small-scale} with four datasets: LUNA2016~\cite{setio2015automatic}, LiTS2017~\cite{bilic2019liver}, BraTS2018~\cite{bakas2018identifying}, and MSD (Heart) \cite{simpson2019large}.  We compare this version with our default settings trained on $55$ datasets (Section \ref{sec:dataset_overview}). Two models are evaluated on dice scores of five 2D segmentation tasks, the accuracy metric of two linear image classifications, and mAP50 of two object detection tasks on VinDr and Kvasir detection. Table \ref{tab:ablation-study-2} shows that LMV-Med full leads to better performance overall, especially with the classification settings; the improvement gap is around $3.6\%$. In summary, we conclude that LVM-Med is beneficial when training in large-scale medical settings.

\begin{minipage}{\textwidth}
  \begin{minipage}[t]{0.48\textwidth}
    \begin{figure}[H]
\centering
\includegraphics[width=\textwidth]{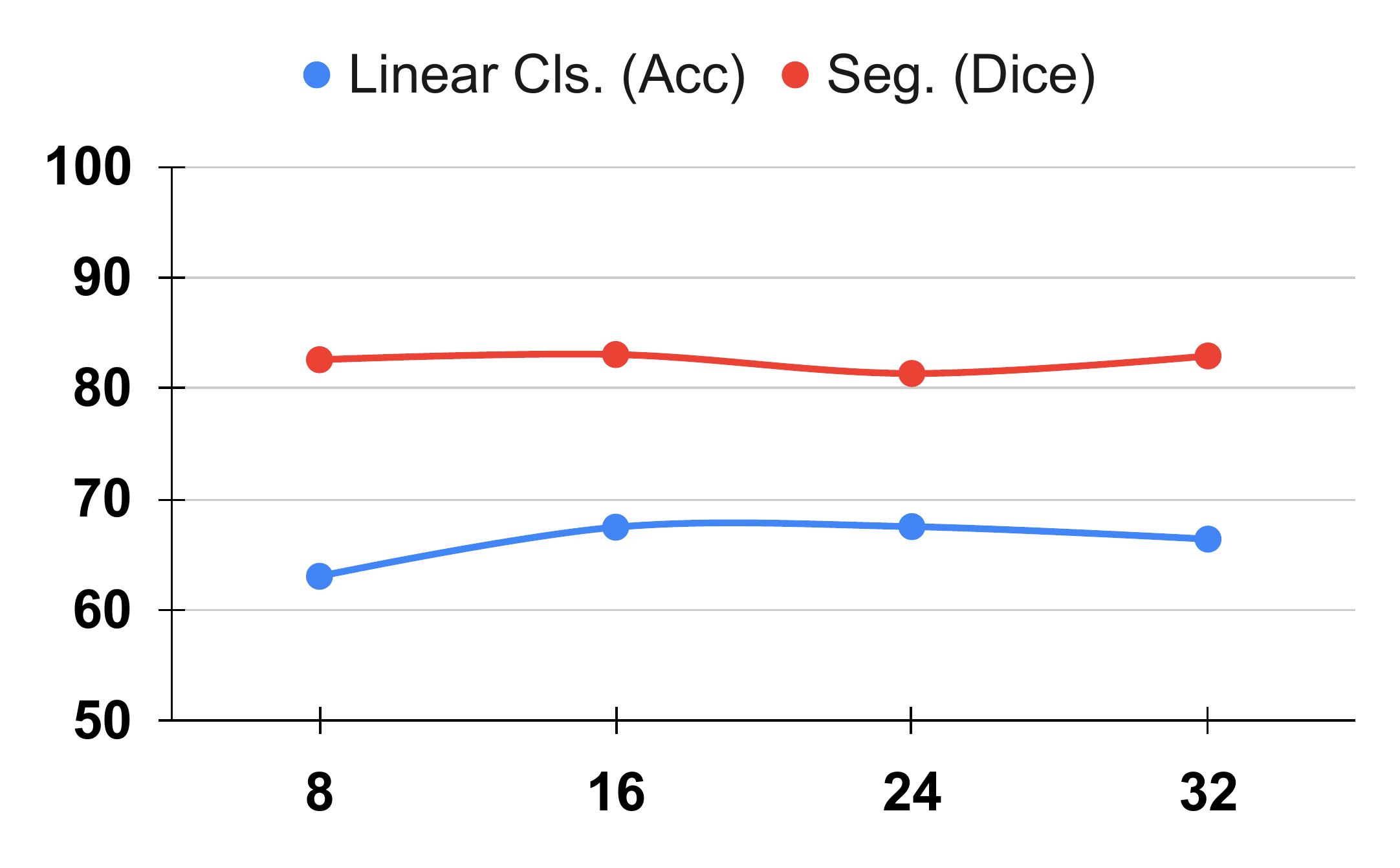}
\caption{\small{LVM-Med performance when varying the number of nodes in graph construction.}}
\label{fig:graph-numbers}
\end{figure}
  \end{minipage}
  \hfill
  \begin{minipage}[t]{0.48\textwidth}
    \begin{figure}[H]
\centering
\includegraphics[width=\textwidth]{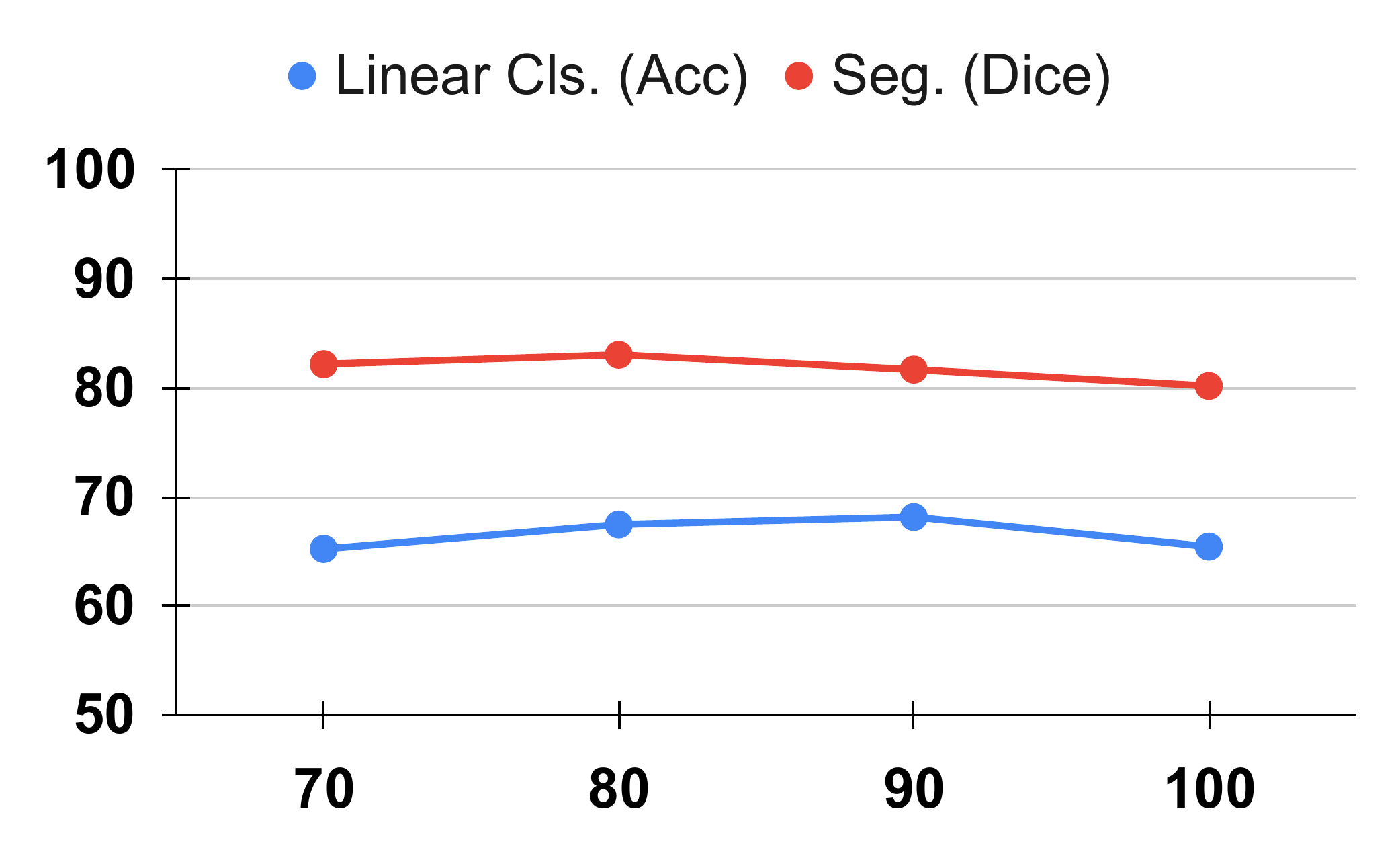}
\caption{\small{LVM-Med performance when varying the $\lambda$ in backpropagation step.}}
\label{fig:lambda_solver}
\end{figure}
  \end{minipage}
\end{minipage}

\subsection{Performance on weighting global and local similarities}
We test with different $\alpha = \{0.7, 0.8, 0.9\}$ which used to fuse global- and local-based similarities $c^{v}_{ij}$. Table \ref{tab:ablation-study-2} demonstrates that $\alpha = 0.8$ is generally the best value in average across segmentation, classification, and object detection tasks.

\begin{table}[H]
\centering
\caption{{LVM-Med ablation studies trained with full data, small-scale, and different hyper-parameter $\alpha$ fusing global- and local-based similarities. Results are reported on an average of five 2D segmentation, two linear classifications, and two object detection tasks. The most impacted factors are highlighted.}}
\label{tab:ablation-study-2}
\vspace{2mm}
\resizebox{0.8\columnwidth}{!}{
\begin{tabular}{ @{\hspace{-0pt}}l@{\hspace{8pt}}c@{\hspace{8pt}}c@{\hspace{10pt}}c}
\toprule
Method         & Cls.(Acc)  & Seg. (Dice) & Detect. (mAP50)\\
\midrule
LVM-Med (full, $\alpha = 0.8$)                    & \textbf{67.47}     & \textbf{83.05}  & {57.1} \\
\rowcolor{cyan!50}
LVM-Med (small-scale, $\alpha = 0.8$)                   &  63.83  & 81.97  & 56.03  \\
LVM-Med (full, $\alpha = 0.7$)         & 65.89    & 82.20 & 56.49  \\
\rowcolor{cyan!50}
LVM-Med (full, $\alpha = 0.9$)            &  65.03   & 81.09 & \textbf{57.14} \\
\bottomrule
\end{tabular}}
\end{table}

\subsection{Computational complexity}
We present a parameter comparison of LVM-Med with other foundation models in Table \ref{tab:num-params}. Our LVM-Med model, based on ResNet-50, has significantly fewer parameters, approximately 3-4 times smaller than models such as Flava or SAM, while still maintaining competitive performance. When utilizing the ViT encoder pre-trained by the SAM method, LVM-Med's parameters are comparable to the Flava model and slightly higher than Clip and Align by $1.03$ and $1.43$ times, respectively. However, it is important to note that both LVM-Med and SAM outperform these models by a significant margin in various settings.

\begin{table}[H]
\centering
\caption{{Computational complexity of our approaches and other foundation models.}}
\label{tab:num-params}
\vspace{2mm}
\resizebox{0.95\columnwidth}{!}{
\begin{tabular}{ @{\hspace{-0pt}}l@{\hspace{10pt}}c@{\hspace{10pt}}c@{\hspace{10pt}}c@{\hspace{10pt}}c@{\hspace{10pt}}c@{\hspace{10pt}}c}
\toprule
Method         & LVM-Med (R50)  & LVM-Med (ViT)  & Clip \citep{radford2021learning} & Flava \citep{singh2022flava} & Align \citep{jia2021scaling} & SAM (Encoder) \citep{kirillov2023segment}\\
\midrule
\textbf{\#Param}   
& 25.55\,M     & 88.88\,M   & 85.80\,M  & 86.39\,M  & 62.14\,M  & 88.88\,M \\
\bottomrule
\end{tabular}}
\end{table}

\section{Prompt-based segmentation on 3D datasets and classification tasks}
\label{sec:linear-seg}
We provide additional results for LVM-Med on 3D-based prompt segmentation and image classification tasks with several fully connected layers. 
\subsection{Promt-based Segmentation on 3D datasets}
We perform experiments on three 3D datasets in Table \ref{tab:prompt-3Dsegmentation}, including BraTS, MMWHS-MRI, and MMWHS-CT. The setup for box prompts follows 2D segmentation cases. We discover that the LMV-Med in 3D cases consistently improves the performance of fine-tuned SAM \cite{ma2023segment} as in 2D settings and attains a large margin compared with SAM without training \cite{kirillov2023segment}. This evidence thus confirms that LVM-Med is also effective under prompt-based scenarios.

\begin{table}[H]
\centering
\caption{Prompt-based segmentation on 3D datasets.}
\vspace{0.1in}
\label{tab:prompt-3Dsegmentation}
\resizebox{1.0\columnwidth}{!}{
\begin{tabular}{l|l|ccc}
\toprule
 & \textbf{Method}              & \textbf{BraTS} & \textbf{MMWHS-MRI} & \textbf{MMWHS-CT} \\ \midrule
\multirow{3}{*}{\textbf{Prompt-based Seg.}} & SAM (fixed encoder) \citep{ma2023segment}& 85.37 $\pm$ 0.07 & 77.64 $\pm$ 1.14 & 76.61 $\pm$ 1.91 \\
 & SAM with Prompt (no-train) \citep{kirillov2023segment}  & 38.97 $\pm$ 0.21  & 59.74 $\pm$ 0.76      & 50.25 $\pm$ 0.33     \\
 & \textbf{LVM-Med (SAM's ViT)} &    \textbf{85.76 $\pm$0.07}            & \textbf{78.91 $\pm$ 0.80}      & \textbf{78.03 $\pm$ 0.93}     \\ \bottomrule
\end{tabular}}
\end{table}

\subsection{Image classification}
We aim to inspect whether foundation models improve their performance given more fully connected layers for image classification tasks with both frozen encoders or fully fine-tuning.
For each method in this category and our LVM-Med (ResNet-50 and ViT), we configure two fully connected layers with sizes $512-256$ and $512-128$ for the Brain and FGADR respectively that map from the output dimension of each network to a number of desired classes. Table \ref{tab:linear-classification} presents obtained results where new settings are highlighted in color. We notice the following points. (i) Firstly, using more fully connected layers tends to improve the performance of foundation models, especially on linear evaluation. For e.g., the Clip increases from $4.79\%-9.98\%$ on FGADR and Brain Tumor classification tasks, respectively. Similarly, our LVM-Med with SAM's ViT also achieves better results by approximately  $1.37\%$ and $4.82\%$ on those tasks. (ii) Secondly, LVM-Med overall  attains the best results in four settings using linear or several fully connected layers with ResNet-50. LVM-Med with ViT architecture also delivers the best records on three of four test cases compared with foundation models. 

\begin{table}[!t]
\begin{center}
\caption{\small{Comparing SSL approaches and Foundation models on classification tasks with two evaluation protocols, Linear evaluation and full Fine-tuning. Settings used with several fully connected layers are in cyan. The best results in 2D-SSL and foundation models (two fully connected layers) are in bold; the best results overall are in bold and underlined.}}
\vspace{0.1in}
\label{tab:linear-classification}
\resizebox{\columnwidth}{!}{
\begin{tabular}{l|l|cccc}
\toprule
\textit{} &
  \textbf{Method} &
  \multicolumn{2}{c}{\textbf{Linear Evaluation (Frozen)}} &
  \multicolumn{2}{c}{\textbf{Fine-tuning}} \\ \midrule
\textit{} &
   &
  \textbf{FGADR (DR Grading)} &
  \textbf{Brain Tumor Class.} &
  \textbf{FGADR (DR Grading)} &
  \textbf{Brain Tumor Class.} \\ \midrule
\multirow{8}{*}{\textbf{2D-SSL on medical}} &
  Twin-Barlon \cite{zbontar2021barlow}&
  {66.86 $\pm$ 0.41} &
  63.03 $\pm$ 0.32 &
  {66.37 $\pm$ 0.77} &
  74.20 $\pm$ 1.38 \\
 &
  Dino \cite{caron2021emerging}&
  65.98 $\pm$ 1.91 &
  62.27 $\pm$ 0.32 &
  67.35 $\pm$ 1.36 &
  71.91 $\pm$ 1.55 \\
 &
  SimCLR \cite{chen2020simple}&
  65.30 $\pm$ 1.70 &
  62.52 $\pm$ 1.67 &
  67.55 $\pm$ 0.28 &
  73.52 $\pm$ 3.56 \\
 &
  Moco-v2 \citep{chen2020improved}&
  65.98 $\pm$ 1.04 &
  62.35 $\pm$ 1.92 &
  67.55 $\pm$ 1.79 &
  74.53 $\pm$ 0.43 \\
 &
  Deepcluster \citep{caron2018deep}&
  65.34 $\pm$ 1.93 &
  {64.47 $\pm$ 0.55} &
  67.94 $\pm$ 1.78 &
  73.10 $\pm$ 0.55 \\
 &
  VicRegl \citep{bardes2022vicregl}&
  64.71 $\pm$ 0.60 &
  59.64 $\pm$ 1.36 &
  65.69 $\pm$ 1.46 &
  73.18 $\pm$ 2.03 \\ 
 &
  \multirow{2}{*}{\textbf{LVM-Med (R50)}} &
  \textbf{\underline{68.33} $\pm$ 0.48} 
  & 66.33 $\pm$ 0.31
  & 68.32 $\pm$ 0.48
  & 76.82 $\pm$ 2.23  \\ 
  & & 
  \cellcolor{cyan!50}
  {{66.67} $\pm$ 0.84} &
  \cellcolor{cyan!50} \textbf{\underline{74.70} $\pm$ 0.84} &
  \cellcolor{cyan!50} \textbf{\underline{70.58} $\pm$ 0.36} &
  \cellcolor{cyan!50} \textbf{\underline{78.77} $\pm$ 0.78} \\ \midrule
\multirow{10}{*}{\textbf{Foundation Model}} &
  \multirow{2}{*}{Clip \citep{radford2021learning}} &
  57.87 $\pm$ 0.50  
  & 57.87 $\pm$ 0.71
  & 57.48 $\pm$ 0.86
  & 34.86 $\pm$ 2.27 \\
  & &
  \cellcolor{cyan!50}62.66 $\pm$ 0.36 &
  \cellcolor{cyan!50}\textbf{67.85 $\pm$ 0.23} &
  \cellcolor{cyan!50}56.21 $\pm$ 1.86 &
  \cellcolor{cyan!50}21.74 $\pm$ 1.14 \\ \cmidrule{2-6}
 &
  \multirow{2}{*}{Flava \citep{singh2022flava}} &
  31.87 $\pm$ 0.69 
  & 35.19 $\pm$ 0.43
  & 57.18 $\pm$ 0.96
  & 34.01 $\pm$ 5.97 \\
  & &
  \cellcolor{cyan!50}32.84 $\pm$ 0.12  &
 \cellcolor{cyan!50} 24.45 $\pm$ 4.30 &
  \cellcolor{cyan!50}{56.01 $\pm$ 0.86} &
  \cellcolor{cyan!50} 33.67 $\pm$ 8.11  \\ \cmidrule{2-6}
 &
  \multirow{2}{*}{Algin \citep{jia2021scaling}} &
  36.95 $\pm$ 1.04 
  & 30.71 $\pm$ 2.35
  & 57.28 $\pm$ 0.97
  & 63.96 $\pm$ 0.04 \\ 
  & &
   \cellcolor{cyan!50}38.12 $\pm$ 1.45 &
 \cellcolor{cyan!50} 30.34 $\pm$ 1.35 &
  \cellcolor{cyan!50}57.87 $\pm$ 0.90 &
  \cellcolor{cyan!50}61.42 $\pm$ 0.25 \\ \cmidrule{2-6}
 &
  \multirow{2}{*}{SAM \citep{kirillov2023segment}} &
  55.13 $\pm$ 0.41 
  & 31.81 $\pm$ 4.26
  & 58.75 $\pm$ 1.32
  & 60.66 $\pm$ 1.36 \\
  & &
  \cellcolor{cyan!50}57.48 $\pm$ 0.24 &
  \cellcolor{cyan!50} 36.89 $\pm$ 1.61 &
  \cellcolor{cyan!50} 58.75 $\pm$ 0.99 &
  \cellcolor{cyan!50} 60.07 $\pm$ 0.31 \\ \cmidrule{2-6}
 &
  \multirow{2}{*}{\textbf{LVM-Med (SAM's ViT)}} &
  62.46 $\pm$ 0.86 
  & 59.31 $\pm$ 0.48
  & 63.44 $\pm$ 0.73
  & 67.34 $\pm$ 2.08\\ 
  & &
  \cellcolor{cyan!50}\textbf{63.83 $\pm$ 1.36} &
  \cellcolor{cyan!50}64.13 $\pm$ 1.14 &
  \cellcolor{cyan!50}\textbf{59.04 $\pm$ 0.14} &
  \cellcolor{cyan!50}\textbf{64.97 $\pm$ 2.71} \\ \bottomrule
\end{tabular}}
\end{center}
\end{table}

\section{Visualizing results}
\begin{figure}[H]
    \centering
    \fontsize{26pt}{26pt}\selectfont
    \resizebox{1.0\columnwidth}{!}{
    \begin{tabular}{cccc}
    \includegraphics[width=10cm]{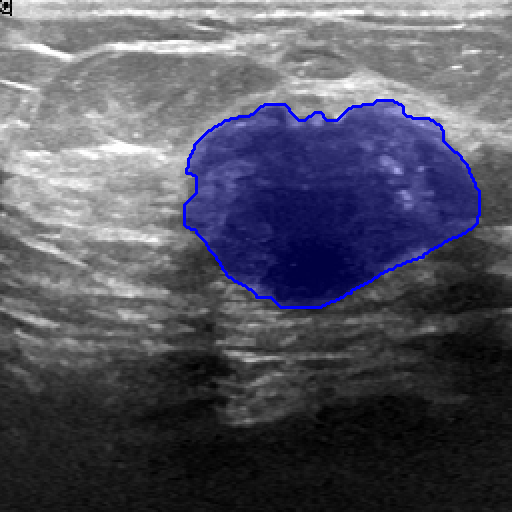} &
      \includegraphics[width=10cm]{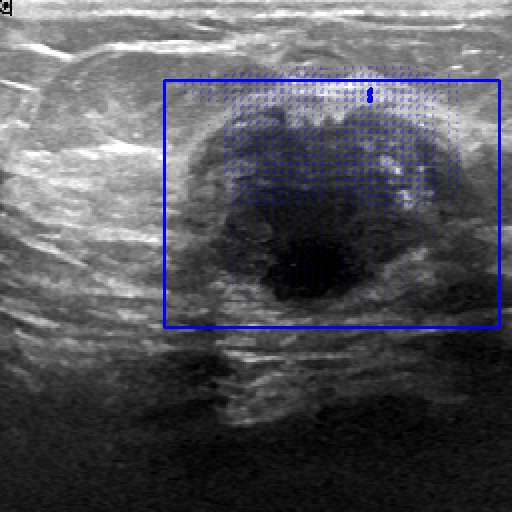} &
      \includegraphics[width=10cm]{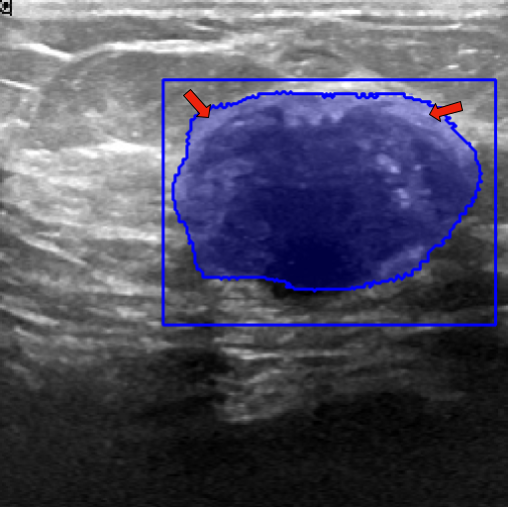}  &
      \includegraphics[width=10cm]{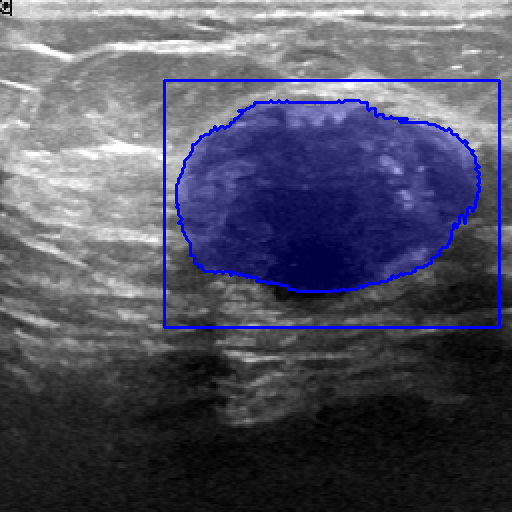}  \\
      
      \includegraphics[width=10cm]{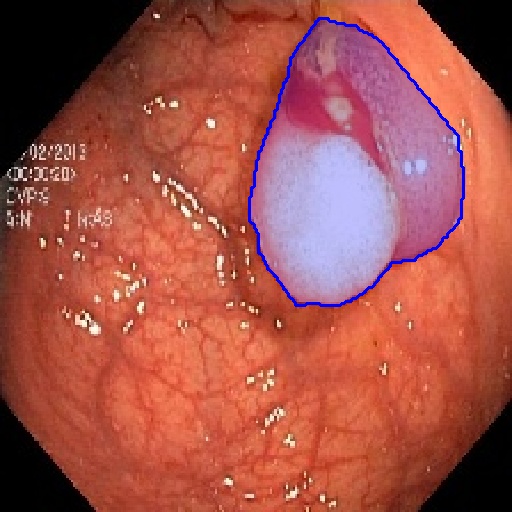} &
      \includegraphics[width=10cm]{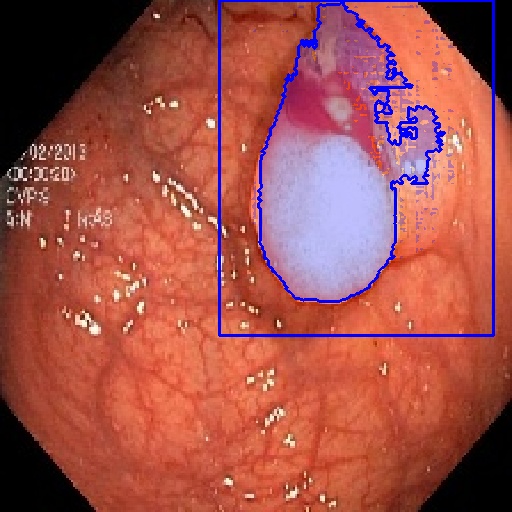} &
      \includegraphics[width=10cm]{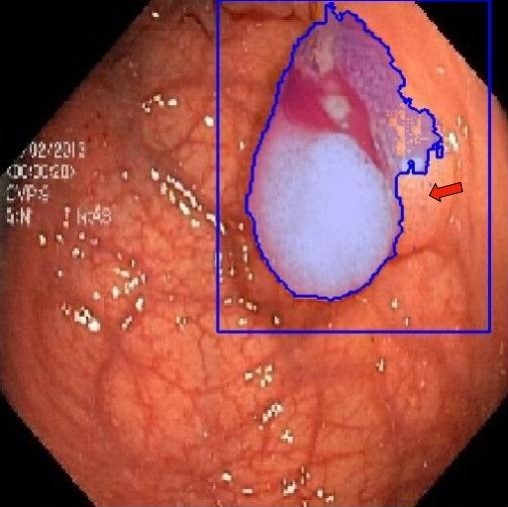}  &
      \includegraphics[width=10cm]{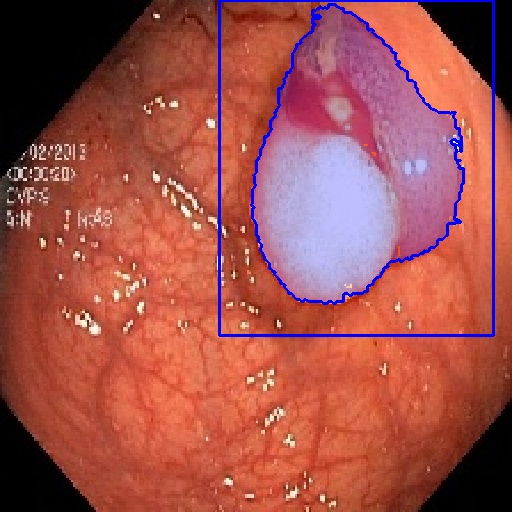}  \\

    \includegraphics[width=10cm]{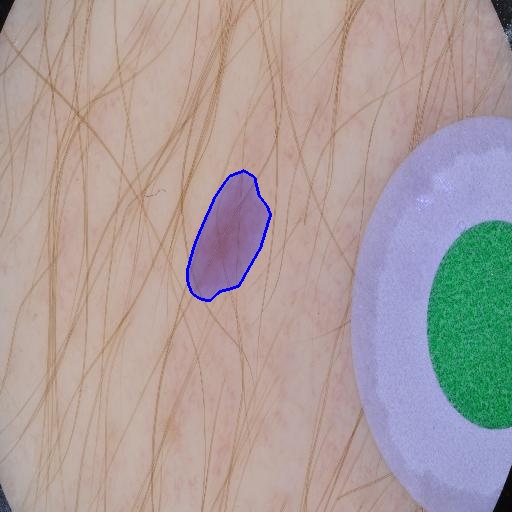} &
      \includegraphics[width=10cm]{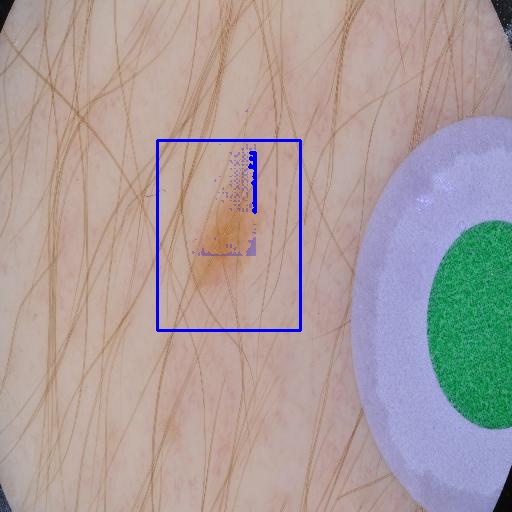} &
      \includegraphics[width=10cm]{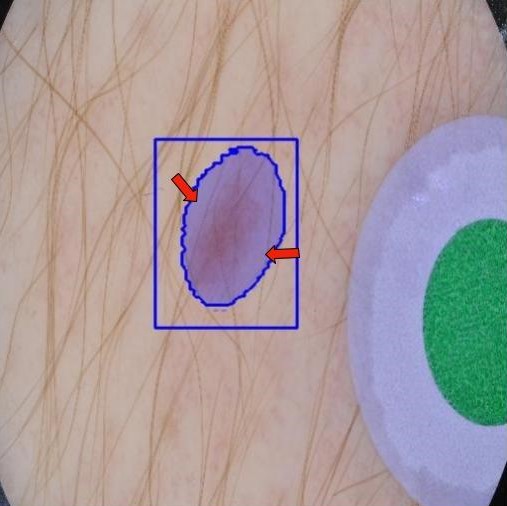}  &
      \includegraphics[width=10cm]{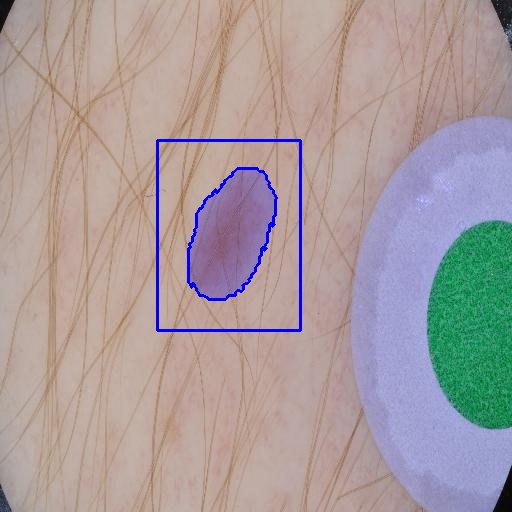}  \\
       
      Ground Truth &
      SAM (no fine-tuning) \cite{kirillov2023segment} &
      SAM (fine-tuning) \cite{ma2023segment}&
      LVM-Med (SAM's ViT)  \\      
    \end{tabular}}    
    \caption{Visualizing prompt-based predictions on three datasets: BUID, Kvasir, and ISIC. Red arrows show differences between SAM (fine-tuning) and LVM-Med using SAM's ViT architecture. Best viewed in color with \textbf{zoom}.}
    \label{fig:SAM_demo}
\end{figure}

We provide qualitative results for prompt-based segmentation in Figure \ref{fig:SAM_demo}. We compare three approaches, including (i) the standard SAM without fine-tuning \cite{kirillov2023segment} (second column), (ii) SAM with encoders and prompt networks are frozen, and only decoder layers are trained as  \cite{mazurowski2023segment} (third column), and (iii) a similar setting as (ii) but encoders taken from LVM-Med version with SAM's ViT architecture (fourth column). For all methods, we simulate box-based prompts using the ground-truth masks and define boxes covering those target regions perturbed by offset values. 

Figure \ref{fig:SAM_demo} demonstrates that the original SAM is prone to generate useless predictions (top and bottom rows) or less precise boundaries. In contrast, updated SAM and LVM-Med produce more accurate results, confirming the importance of fine-tuning to achieve adequate results. Figures in the third and fourth columns also illustrate that SAM tends to over-segment or lacks structures on an object's edges in several cases, while LVM-Med is more stable in those situations (red arrows). 
\label{sec:visualize_images}

\section{Dataset overviews}
\label{sec:dataset_overview}
Table \ref{tab:data1} overviews the dataset used in our study. For each dataset, we provide its modality, data dimension, and the total of samples. If the training/testing rate is available (column \textbf{Train/Test Rate}), we utilize all training data; otherwise, we sample $20\%$ total samples to avoid potential test data leaking for downstream tasks used in the pre-training  step. For datasets whose data dimensions are  3D volumes, we sample 2D slices from those formats. Some datasets, such as MSD or ADNI, comprise different sub-datasets inside; we consider these sub-sets as independent ones to avoid confusion during the training steps. In summary, a total of $55$ datasets are used with approximately $40\%$ in 3D datasets and $60\%$ in 2D images as presented in Figure \ref{fig:3d-2d-ration}. Moreover, we also outline ratios between distinct data modalities such as MRI, CT, X-ray, grayscale types such as Ultrasound, OCT, and finally, color images depicted in Figure \ref{fig:modularity-ratio}.
\vspace{-0.3in}
\begin{minipage}{\textwidth}
  \begin{minipage}[t]{0.48\textwidth}
    \begin{figure}[H]
\centering
\includegraphics[width=\textwidth]{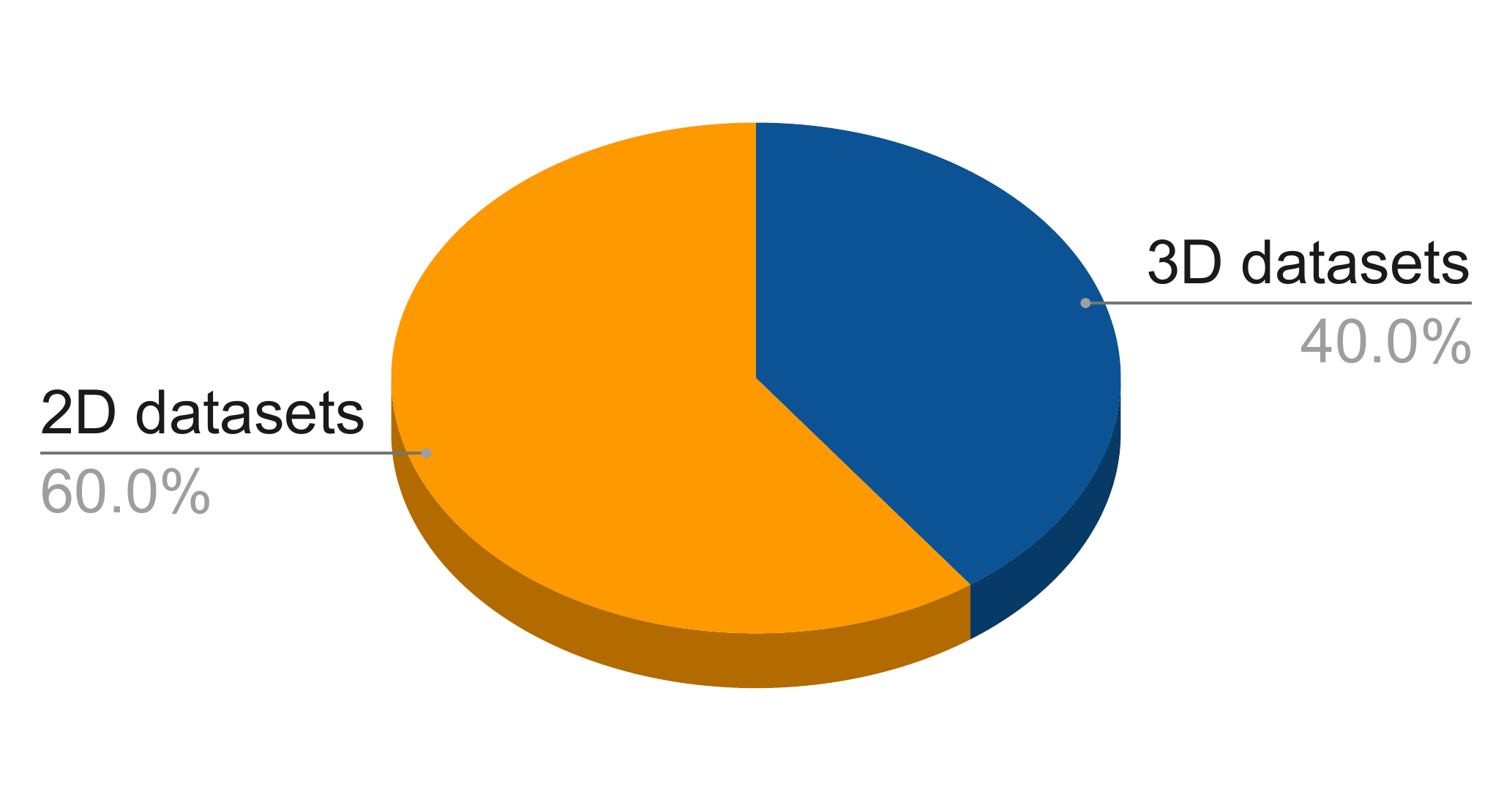}
\caption{\small{Pie chart illustrating the ratio of the number of datasets with the 3D  and 2D dimension.}}
\label{fig:3d-2d-ration}
\end{figure}
  \end{minipage}
  \hfill
  \begin{minipage}[t]{0.48\textwidth}
    \begin{figure}[H]
\centering
\includegraphics[width=\textwidth]{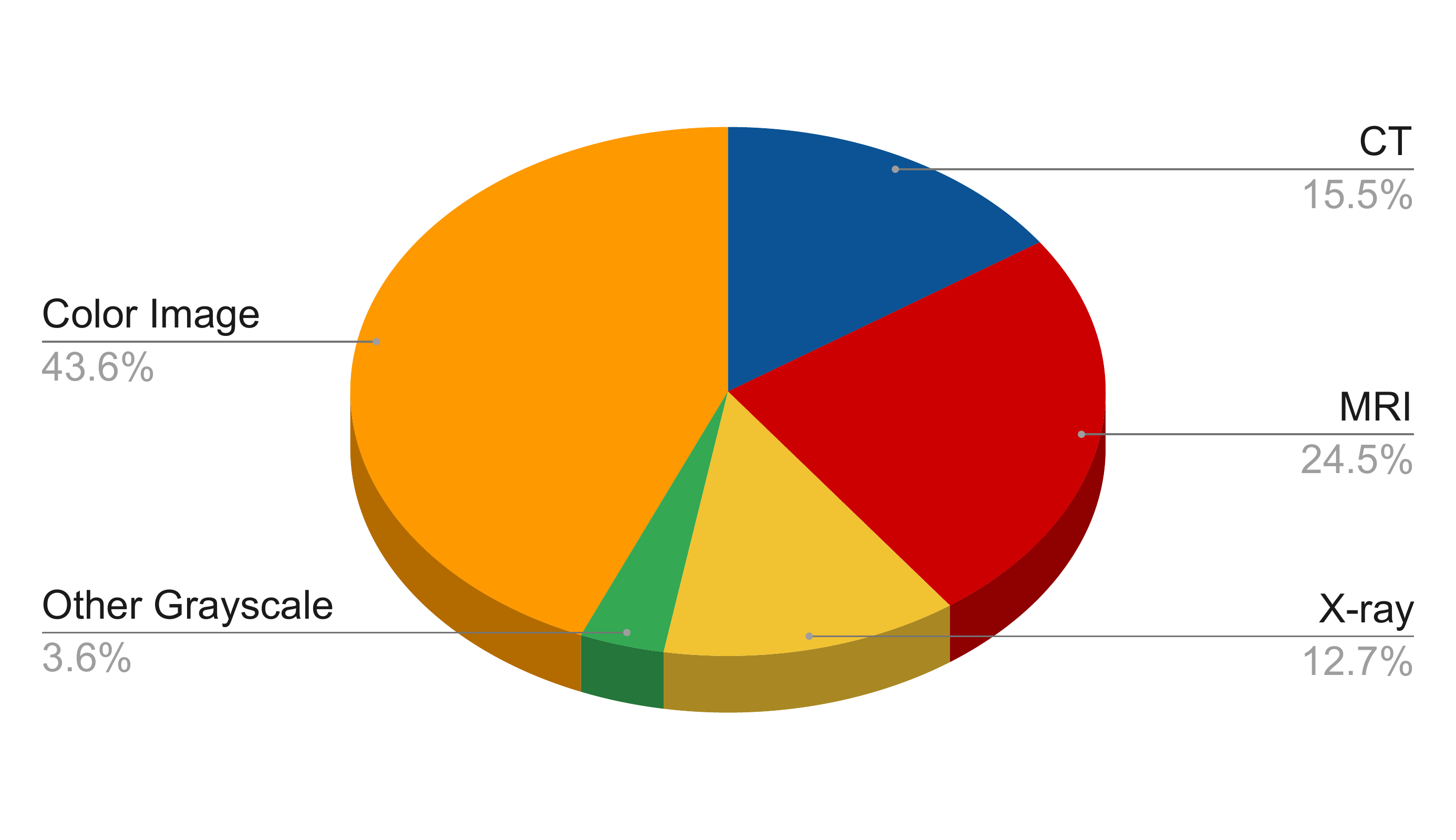}
\caption{\small{Pie chart illustrating the ratio of different data modalities in our collected dataset.}}
\label{fig:modularity-ratio}
\end{figure}
  \end{minipage}
\end{minipage}

\begin{table}[H]
\caption{Overview of our collected medical dataset}
\vspace{2mm}
\centering
\resizebox{1.0\columnwidth}{!}{
\begin{tabular}{p{1cm}p{2cm}p{2cm}p{3cm}p{1.2cm}p{1.5cm}p{1.23cm}p{1cm}}
\toprule
\textbf{No}&\textbf{Data Name}&\textbf{Topic}&\textbf{Disease}&\textbf{Modality}&\textbf{Dimension}&\textbf{Train/Test Rate?}&\textbf{Total}\\ \midrule
1&HyperKvasir \cite{borgli2020hyperkvasir}&Polyp&Pathological classification&Color images&2D&Yes&110079\\\midrule
2&PatchCamelyon \cite{Veeling2018-qh, bejnordi2017diagnostic}&Cells&Histopathologic scans of lymph node sections.&Color images&2D&Yes&327680\\\midrule
3&BraTS2018 \cite{6975210, lloyd2017high, bakas2018identifying}&Brain&Tumor Segmentation&MRI&3D&No&760\\\midrule
4&HNSCC \cite{grossberg2018imaging} &Head Neck&No Label&CT&3D&No&155\\\midrule
5&LiTS2017 \cite{bilic2023liver}&Liver&Segmentation of Liver and Tumor Lesions&CT&3D&No&200\\\midrule
6&MSD-Heart \cite{simpson2019large}&Heart&Heart Segmentation&MRI&3D&No&30\\\midrule
7&MSD-Liver \cite{simpson2019large}&Liver&Liver Segmentation&MRI&3D&No&201\\\midrule
8&MSD-Lung \cite{simpson2019large}&Lung&Lung Segmentation&MRI&3D&No&96\\\midrule
9&MSD-Pancreas \cite{simpson2019large}&Pancreas&Pancrea Segmentation&MRI&3D&No&420\\\midrule
10&MSD-HepaticVessel \cite{simpson2019large}&Hepatic Vessel &Hepatic Vessel Segmentation&MRI&3D&No& 443\\\midrule
11&MSD-Spleen \cite{simpson2019large}&Spleen&Spleen Segmentation&MRI&3D&No& 61\\\midrule
12&MSD-Colon \cite{simpson2019large}&Colon&Colon Segmentation&MRI&3D&No&190\\\midrule
13&OPC-Radiomics \cite{kwan2018radiomic,clark2013cancer, kwandata}&Oropharynx&No Label&CT&3D&No&120\\\midrule
14&Osteosarcoma-UT \cite{leavey2019osteosarcoma,yorke2019pelvic, clark2013cancer}&Osteosarcoma&No Label&Color images&2D&No&547\\ \midrule
15&Pancreas-CT \cite{roth2016data, roth2015deeporgan, clark2013cancer}&Pancreas&No Label&CT&3D&No&16\\\bottomrule
\end{tabular}}
\vspace{1mm}
\label{tab:data1}
\end{table} 

\begin{table}[H]
\centering
\resizebox{1.0\columnwidth}{!}{
\begin{tabular}{p{1cm}p{2cm}p{2cm}p{3cm}p{1.2cm}p{1cm}p{1.23cm}p{1cm}}
\toprule
\textbf{No}&\textbf{Data Name}&\textbf{Topic}&\textbf{Disease}&\textbf{Modality}&\textbf{Format}&\textbf{Default Train/Test Rate}&\textbf{Total}\\ \toprule
16&Pelvic-Reference-Data \cite{yorke2019pelvic, clark2013cancer}&Pelvic&No Label&CT&3D&No&12\\\midrule
17&ProstateX \cite{litjens2014computer, Litjens2017, clark2013cancer}&Prostate&The clinical significance of prostate lesions prediction&MRI&3D&No&40\\\midrule
18&TCGA-CESC \cite{lucchesi2016radiology, clark2013cancer}&Cervical&No Label&Color images&2D&No&3977\\\midrule
19&TCGA-COAD \cite{kirk2016radiology, clark2013cancer}&Colon&No Label&Color images&2D&No&1644\\\midrule
20&TCGA-ESCA \cite{Lucchesi20164, clark2013cancer}&Cuticle&No Label&Color images&2D&No&4427\\\midrule
21&TCGA-KICH \cite{linehan2016radiology, clark2013cancer}&Kidney&No Label&Color images&2D&No&2192\\\midrule
22&TCGA-KIRC \cite{akin2016radiology, clark2013cancer}&Kidney&No Label&Color images&2D&No&34108\\\midrule
23&TCGA-READ \cite{kirk2016radiology, clark2013cancer}&Rectum&No Label&Color images&2D&No&248\\\midrule
24&TCGA-SARC \cite{roche2016radiology, clark2013cancer}&Sarcoma&No Label&Color images&2D&No&624\\\midrule
25&TCGA-THCA \cite{kirk2016radiology, clark2013cancer}&Thyroid&No Label&Color images&2D&No&665\\\midrule
26&VinDr \citep{nguyen2022vindr}&Lung&Abnormal Disease Classification&X-ray&2D&No&18000\\\midrule
27&LUNA2016 \cite{setio2015automatic}&Lung&Nodule Detection and False Positive Reduction&CT&3D&No&49386\\\midrule
28&BCCD \cite{BCCD_Dataset}&Cells&Blood cell detection&Color images&2D&No&364\\ \midrule
29&C-NMC\_Leukemia \cite{gehlot2020sdct, gupta2019all}&Cells&Leukemia detection&Color images&2D&Yes&12529\\ \midrule
30&CBIS-DDSM \cite{lee2017curated, sawyer2016curated}&Breast&Breast Cancer Classification&X-ray&2D&No&6774\\\midrule
31&COVIDx \cite{Wang2020}&Lung&Covid-19 Detection&X-ray&2D&Yes&194922\\\midrule
32&Heidelberg OCT \cite{kermany2018labeled}&Eye&OCT Imaging Classification&OCT&2D&Yes&84495\\\midrule
33 &m2caiSeg \cite{maqbool2020m2caiseg}&Laparoscopic&Semantic Segmentation Laparoscopic&Color images&2D&Yes&614\\\midrule
34&NuCLS \cite{amgad2102nucls}&Nucleus&Nucleus Segmentation Detection / Classification&Color images&2D&Yes&1744\\\midrule
35&SARAS-MESAD \cite{cuzzolin2021saras}\cite{saras-mesad2}\cite{saras-mesad3}&Prostatectomy Procedures&Action classification in Prostatectomy Surgey&Color images&2D&Yes&29454\\\midrule
36&Shoulder X-ray images from Sun Yat-sen Memorial Hospital \cite{shoulder}&Shoulder&Shoulder X-ray Classification&X-ray&2D&Yes&1049\\\bottomrule
\end{tabular}}
\vspace{1mm}
\label{tab:data2}
\end{table}

\begin{table}[H]
\centering
\resizebox{1.0\columnwidth}{!}{
\begin{tabular}{p{1cm}p{2cm}p{2cm}p{3cm}p{1.2cm}p{1cm}p{1.23cm}p{1cm}}
\toprule
\textbf{No}&\textbf{Data Name}&\textbf{Topic}&\textbf{Disease}&\textbf{Modality}&\textbf{Format}&\textbf{Default Train/Test Rate}&\textbf{Total}\\ \toprule
37&Shenzhen Hospital X-ray Set \cite{ShenzhenHospitalX-raySet}&Lung&Lung segmentation&X-ray&2D&No&566\\\midrule
38&ADNI 1.5T  \cite{mueller2005alzheimer,petersen2010alzheimer}&Brain&Alzheimer's Disease Classification&MRI&3D&No&639\\\midrule
39&ADNI 3T \cite{mueller2005alzheimer,petersen2010alzheimer}&Brain&Alzheimer's Disease Classification&MRI&3D&No&119\\\midrule
40&AML-Cytomorphology \cite{matek2019single, matek2019human,clark2013cancer}&Cell&Peripheral blood smears&Color images&2D&No&18365\\\midrule
41&APTOS 2019 \cite{karthick2019aptos}&Eye&Severity of diabetic retinopathy Classification&Color images&2D&No&3662\\ \midrule
42&BCSS \cite{amgad2019structured}&Cells&Breast cancer semantic segmentation&Color images&2D&No&151\\\midrule
43&Dental Panoramic \cite{abdi2020panoramic}&Tooth&Mandible segmentation&X-ray&2D&No&116\\\midrule
44&HC18 \cite{van2018automated}&Fetal&Fetal head circumference (HC)&Ultrasound&2D&No&999\\\midrule
45&Hippseg 2011 \cite{Hippseg2011}&Brain&Hippocampus Segmentation&MRI&3D&No&3050\\\midrule
46&ISIC Challenge 2019 \cite{isic}&Skin&Skin Cancer Classification&Color images&2D&No&25331\\\midrule
47&KiTS19-21 \cite{taha2018kid}&Kidney&Kidney Segmentation&CT&3D&No&45424\\\midrule
48&Kvasir v2 \cite{kvarsir}&Gastrointestinal&Gastrointestinal cancer image classification&Color images&2D&No&6000\\\midrule
49&LHNCBC Malaria \cite{Malaria}&Cells&Malaria Classification&Color images&2D&No&27560\\\midrule
50&MitoEM \cite{wei2020mitoem}&Cells&Mitochondria Instance Segmentation&MRI/CT&3D&No&1000\\\midrule
51&MLL Bone Marrow \cite{matek2021highly}&Cells&Blood cell classification&Color images&2D&No&171374\\\midrule
52&MMWHS-CT\cite{zhuang2016multi}&Heart&Sub-structure Heart segmentation&CT&3D&Yes& 40\\\midrule
53&MMWHS-MRI \cite{zhuang2016multi}&Heart&Sub-structure Heart segmentation&MRI&3D&Yes& 40\\\midrule
54&RSNA Bone Age \cite{halabi2019rsna}&Bone&Bone age prediction&X-ray&2D&No&12611\\\midrule
55&EyePACS \cite{eyePACS} &Eye& Diabetic Retinopathy Detection&Color Images&2D&Yes&88702\\\bottomrule
\end{tabular}}
\vspace{2mm}
\label{tab:data3}
\end{table} 

\end{document}